\documentclass[lettersize,journal]{IEEEtran}
\usepackage{amsmath,amsfonts}
\usepackage{algorithmic}
\usepackage{algorithm}
\usepackage{array}
\usepackage[caption=false,font=normalsize,labelfont=scriptsize,textfont=scriptsize]{subfig}
\usepackage{textcomp}
\usepackage{stfloats}
\usepackage{url}
\usepackage{verbatim}
\usepackage{graphicx}
\usepackage{cite}
\usepackage{threeparttable}
\usepackage{multirow}
\usepackage{multicol}
\usepackage{booktabs}
\usepackage{makecell}
\usepackage{pifont}
\usepackage[colorlinks,bookmarksopen,bookmarksnumbered,citecolor=green,linkcolor=red,urlcolor=blue]{hyperref}
\usepackage[table]{xcolor}

\usepackage{ragged2e}

\begin{document}

\title{Vision-Language Modeling Meets Remote Sensing: Models, Datasets and Perspectives}

\author{Xingxing Weng, Chao Pang, and Gui-Song Xia 
\thanks{X. Weng, C. Pang, and G.-S. Xia are with the School of Computer Science, Wuhan University, Wuhan {\rm 430072}, China. C. Pang, and G.-S. Xia are also with the School of Artificial Intelligence, Wuhan University, Wuhan {\rm 430072}, China.}
\thanks{\textit{Corresponding Author: Gui-Song Xia (guisong.xia@whu.edu.cn)}}}



\maketitle

\begin{abstract}
Vision-language modeling (VLM) aims to bridge the information gap between images and natural language. Under the new paradigm of first pre-training on massive image-text pairs and then fine-tuning on task-specific data, VLM in the remote sensing domain has made significant progress. The resulting models benefit from the absorption of extensive general knowledge and demonstrate strong performance across a variety of remote sensing data analysis tasks. Moreover, they are capable of interacting with users in a conversational manner. In this paper, we aim to provide the remote sensing community with a timely and comprehensive review of the developments in VLM using the two-stage paradigm. Specifically, we first cover a taxonomy of VLM in remote sensing: contrastive learning, visual instruction tuning, and text-conditioned image generation. For each category, we detail the commonly used network architecture and pre-training objectives. Second, we conduct a thorough review of existing works, examining foundation models and task-specific adaptation methods in contrastive-based VLM, architectural upgrades, training strategies and model capabilities in instruction-based VLM, as well as generative foundation models with their representative downstream applications. Third, we summarize datasets used for VLM pre-training, fine-tuning, and evaluation, with an analysis of their construction methodologies (including image sources and caption generation) and key properties, such as scale and task adaptability. Finally, we conclude this survey with insights and discussions on future research directions: cross-modal representation alignment, vague requirement comprehension, explanation-driven model reliability, continually scalable model capabilities, and large-scale datasets featuring richer modalities and greater challenges.
\end{abstract}

\begin{IEEEkeywords}

Remote Sensing, Vision-Language Modeling, Contrastive Learning, Visual Instruction Tuning, Diffusion Model
\end{IEEEkeywords}

\section{Introduction}
\IEEEPARstart{V}{ision}-language modeling (VLM) in remote sensing, aiming to bridge the information gap between remote sensing images and natural language, facilitates a deeper understanding of remote sensing scene semantics like the attributes of ground objects and their relationship, and enables more natural human interaction with intelligent remote sensing data analysis models or methods~\cite{bashmal2023language,li2024vision}. Since the introduction of remote sensing tasks such as image captioning~\cite{qu2016deep}, visual question answering~\cite{lobry2020rsvqa}, text-image (or image-text) retrieval~\cite{abdullah2020textrs}, and text-based image generation~\cite{bejiga2019retro}, VLM in remote sensing has achieved significant success, driven by advancements in deep learning. 

\begin{figure}[ht]
\centering
\includegraphics[width=\linewidth]{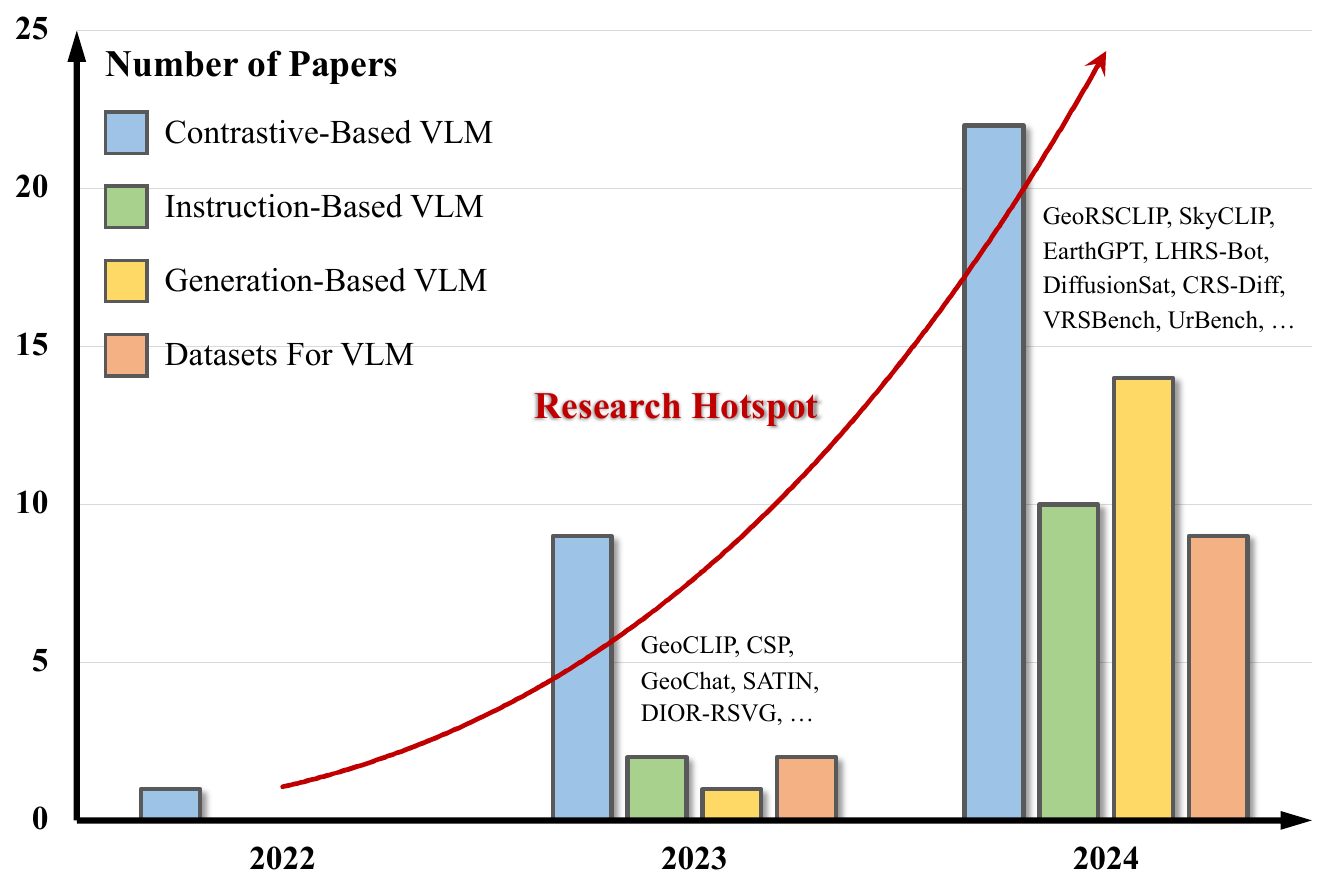}
\caption{The number of publications on visual-language modeling in remote sensing using the pre-training and fine-tuning paradigm.}
\label{fig:papernumber}
\end{figure}

Early works on VLM primarily emphasize the careful design of model architectures, followed by supervised training from scratch on small-scale datasets. For example, in image captioning research, many efforts~\cite{zhang2019vaa,li2020multi,zhao2021high,zia2022transforming} have been made to effectively combine convolutional neural networks (\emph{e.g.} VGG~\cite{simonyan2014very} and ResNet~\cite{he2016deep}) with sequential models (\emph{e.g.} LSTM~\cite{schmidhuber1997long} and Transformer~\cite{vaswani2017attention}) before training on UCM-captions~\cite{qu2016deep} and Sydney-captions~\cite{qu2016deep} datasets. Under this classical construction paradigm, deep models often excel on test datasets but struggle to perform satisfactorily in large-scale deployments. Moreover, although these models are capable of describing image content, they fall short when tasked with answering questions about the images. In other words, they struggle to accomplish related tasks, such as visual question answering. The task-specific nature of these models seriously limits their applicability across diverse scenarios.

Recently, a new paradigm of pre-training followed by fine-tuning provides a promising solution to address the challenges mentioned above. The core idea is to first pre-train a model on massive image-text data, enabling it to capture general knowledge that covers a wide range of visual and textual concepts, along with their underlying correspondence. The pre-trained model is then fine-tuned on task-specific training data. The integration of general knowledge has been shown to enhance the model's generalization ability in a single task~\cite{zhang2024rs5m,wang2024skyscript}, while also making it more versatile and capable of handling a variety of downstream tasks~\cite{hu2023rsgpt,kuckreja2024geochat}. Consequently, vision-language modeling with this new paradigm has emerged as a prominent research focus in the field of remote sensing. 

To date, significant progress has been achieved, as illustrated in Fig.~\ref{fig:papernumber}. This includes works based on 1) contrastive learning~\cite{hadsell2006dimensionality}, such as GeoRSCLIP~\cite{zhang2024rs5m}, SkyCLIP~\cite{wang2024skyscript} and RemoteCLIP~\cite{liu2024remoteclip}, which have driven substantial advancements in various cross-modal tasks and zero-shot image understanding tasks. 2) learning an implicit joint distribution between text and images, like RS-SD~\cite{zhang2024rs5m}, DiffusionSat~\cite{khanna2023diffusionsat} and CRS-Diff~\cite{tang2024crs}, which allow for image generation from text prompts. 3) visual instruction tuning~\cite{liu2024visual}, such as GeoChat~\cite{kuckreja2024geochat}, LHRS-Bot~\cite{muhtar2024lhrs}, and SkySenseGPT~\cite{luo2024skysensegpt}, which have demonstrated improved performance, diverse capabilities, and conversational interactions in remote sensing data analysis.

\vspace{-0.1em}

Despite these remarkable achievements, it is widely acknowledged that VLM remains an open challenge. Indeed, existing works have not yet achieved the level of remote sensing experts in processing remote sensing data. To provide clarity and motivation for further advances in the research community, several surveys have reviewed vision-language modeling in remote sensing. For instance, Li et al.~\cite{li2024vision} summarize vision-language models from an application perspective and suggest potential research opportunities. However, due to time constraints, they primarily concentrate on vision-only foundation models and early works. Zhou et al.~\cite{zhou2024towards} review recent developments but lack an in-depth analysis of key designs, which is significant for inspiring future research. Moreover, datasets, as a prerequisite of visual-language modeling research, have not been given adequate attention in existing surveys.

\begin{figure*}[t]
\centering
\includegraphics[width=\textwidth]{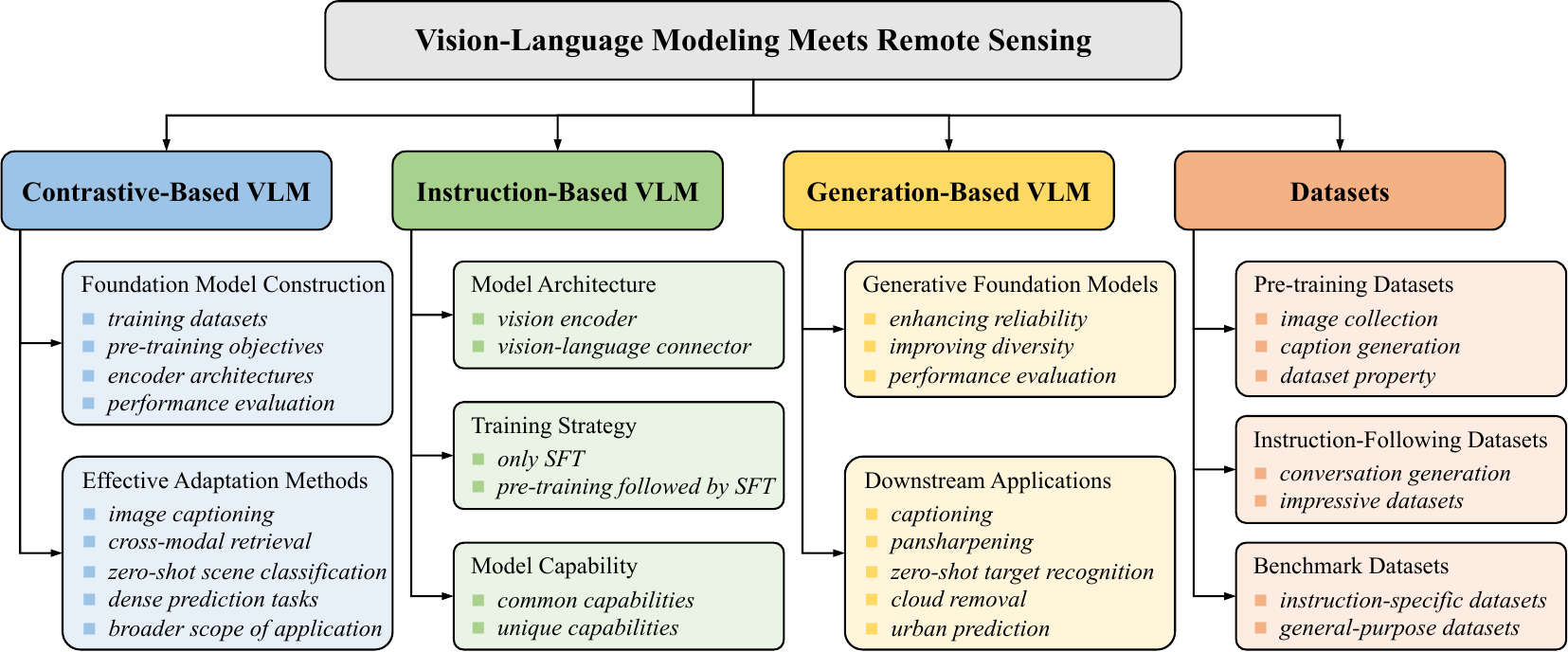}
\caption{Overview of this survey.}
\label{fig:overview}
\end{figure*}

In this work, we aim to provide a timely and comprehensive review of the literature, with a focus on vision-language modeling based on the \textit{pre-training and fine-tuning} paradigm in the field of remote sensing. Specifically, we cover: 1) a taxonomy of VLM in remote sensing, detailing commonly used network architectures and pre-training objectives for each category; 2) the latest advancements in contrastive-based, instruction-based, and generation-based vision-language modeling in remote sensing, highlighting key designs and downstream applications; 3) progress in datasets for VLM pre-training, fine-tuning, and evaluation; 4) several challenges and potential research directions. Fig.~\ref{fig:overview} presents an overview of this paper.

\begin{figure*}[ht]
\centering
\subfloat[Contrastive-based VLM]{\includegraphics[width=0.3\textwidth]{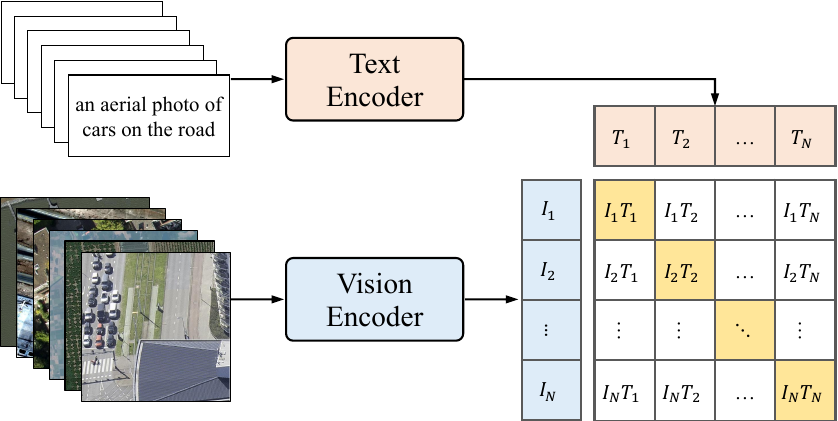} \label{constrastive-vlm}} \hfill
\subfloat[Instruction-based VLM]{\includegraphics[width=0.3\textwidth]{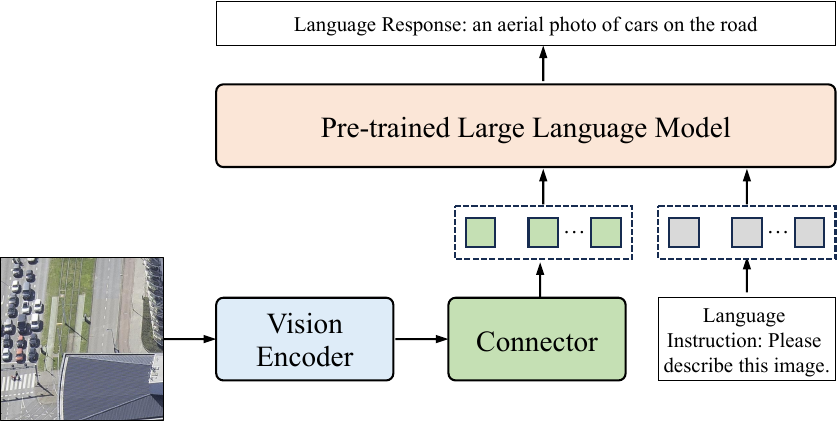} \label{instruction-vlm}} \hfill
\subfloat[ Generation-based VLM]{\includegraphics[width=0.3\textwidth]{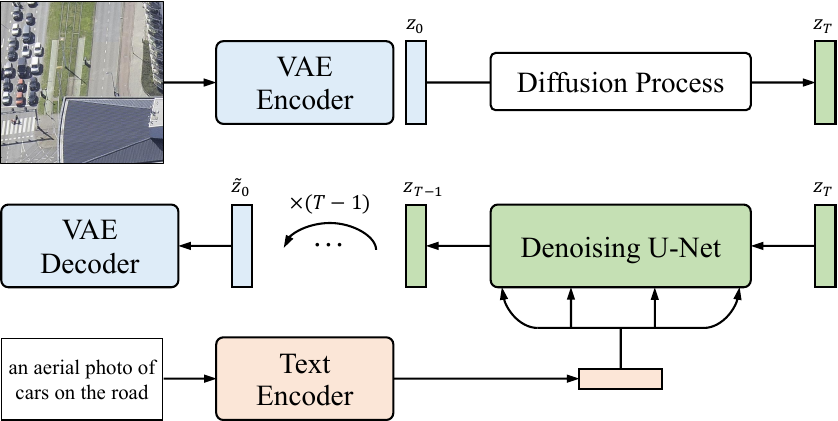} \label{generation-vlm}}
\caption{Based on the strategies employed to bridge remote sensing images and natural language during the pre-training phase, vision-language modeling (VLM) in remote sensing can be categorized into three groups: (a) contrastive learning, (b) visual instruction tuning, and (c) text-conditioned image generation. This image is recreated by us based on \cite{radford2021learning,liu2024visual,rombach2022high}.}
\label{fig:taxonomy-VLM}
\end{figure*}

\section{The Taxonomy of Visual-Language Modeling in Remote Sensing}
Under the \textit{pre-training and fine-tuning} paradigm, vision-language modeling in remote sensing can be divided into three distinct groups based on their strategies for bridging the two modalities in the pre-training phase: contrastive learning, visual instruction tuning, and text-conditioned image generation. In this section, we present commonly used network architectures and pre-training objectives within each group.

\paragraph*{\textbf{Contrastive Learning}} The motivation behind applying contrastive learning to vision-language modeling is training a model to map vision and language into a shared representation space, where an image and its corresponding text share similar representations while differing from other texts, which was first implemented by the pioneering work CLIP~\cite{radford2021learning} in the field of computer vision. As illustrated in Fig.\ref{fig:taxonomy-VLM} (a), CLIP utilizes two independent encoders responsible for encoding visual and textual information. Given a batch of $N$ image-text pairs $\{(x_i, y_i)\}_{i=1}^{N}$, the embeddings extracted by the two encoders followed by normalization are $\{\mathbf{e}_i^I, \mathbf{e}_i^T\}_{i=1}^N$. To achieve the similarity between visual and textual representations, CLIP is trained to maximize the cosine similarity of the embeddings of the $i_{th}$ image and the corresponding $i_{th}$ text ($i \in [N]$ with $[N]=\{1, 2, \dots, N\}$), while minimizing the cosine similarity of the embeddings of the $i_{th}$ image and $j_{th}$ text ($i, j \in [N]$, $i \neq j$). The loss is computed using InfoNCE~\cite{oord2018representation} as follows:
\begin{equation}
\resizebox{0.9\linewidth}{!}{$
\mathcal{L}_{\mathrm{InfoNCE}}=-\frac{1}{2N}\sum_{i=1}^N \left(\log \sigma_\tau(s_{ii}^I, \{s_{ij}^I\}_{j=1}^N) + \log \sigma_\tau(s_{ii}^T, \{s_{ij}^T\}_{j=1}^N)  \right)$},
\end{equation}
where $s_{ij}^I=\mathcal{S}(\mathbf{e}_i^I,\mathbf{e}_j^T)$ and $s_{ij}^T=\mathcal{S}(\mathbf{e}_i^T,\mathbf{e}_j^I)$ are the similarity scores between the image and text embeddings, as computed by the function $\mathcal{S}(\mathbf{e}, \mathbf{e}^{'})=\frac{\mathbf{e} \cdot \mathbf{e}^{'}}{|\mathbf{e}||\mathbf{e}^{'}|}$. $\sigma_\tau(s_i,\{ s_j\}_{j=1}^N)= \mathrm{exp}(s_i/\tau)/(\sum_{j=1}^N \mathrm{exp}(s_j/\tau))$ is the softmax function, which normalizes the similarity score $s_{ii}^I$ or $s_{ii}^T$ over the corresponding sets $\{s_{ij}^I\}_{j=1}^N$ or $\{s_{ij}^T\}_{j=1}^N$. $\tau$ is the temperature.

The original CLIP model was trained on 400 million image-text pairs collected from the Internet and has demonstrated impressive results across various computer vision tasks~\cite{li2022comprehending,wang2023clipn,zhou2023zegclip}. These advancements spark interest in extending its capability to advance vision-language modeling in remote sensing. Two primary lines of research have been actively explored. The first, following the CLIP learning way, focuses on pre-training foundation models that are task-agnostic but specifically adapted for remote sensing domain. This includes efforts such as constructing large-scale image-text datasets~\cite{zhang2024rs5m,wang2024skyscript} and developing novel pre-training objectives~\cite{mall2024remote,song2024set}. The second line explores effective adaptation of pre-trained CLIP models toward diverse downstream tasks, including image captioning~\cite{li2023rs,wang2024multi}, zero-shot scene classification~\cite{al2023vision,xu2023deep}, image-text retrieval~\cite{zavras2024mind,zhao2024luojiahog}, etc.

\paragraph*{\textbf{Visual Instruction Tuning}} Optimizing a model from scratch for image-text alignment is extremely resource-intensive due to the need for vast amounts of data and computational power. Fortunately, many pre-trained vision encoders and language models have been released. Pre-trained vision encoders can provide high-quality visual representations, while pre-trained language models, particularly large language models (LLMs), demonstrate advanced language understanding capabilities. As a result, recent works increasingly leverage these models to achieve image-text alignment through visual instruction tuning, as introduced by LLaVA~\cite{liu2024visual} and MiniGPT-4~\cite{zhu2023minigpt}. Fig.\ref{fig:taxonomy-VLM} (b) illustrates a network architecture for this type of work, consisting of three key components: a pre-trained vision encoder, a connector, and a large language model. Specifically, the vision encoder compresses remote sensing images into compact visual representations, while the connector maps image embeddings into the word embedding space of the LLM. The LLM then receives both visual information and language instructions to perform reasoning tasks. Different from CLIP, which directly takes images and corresponding texts as input, this type of work preprocesses image-text pairs to instruction-following data. In this setup, each image is accompanied by a question, or language instruction, requiring the LLM to describe the image, while the corresponding text serves as the ground truth for the LLM's predictions. Denote a batch of instruction-following data as $\{(x_i, q_i, y_i)\}_{i=1}^N$, where $q_i$ is the question associated with the $i_{th}$ image. The pre-training objective is defined as:
\begin{equation}
\mathcal{L}_{\mathrm{VIT}}=-\frac{1}{N}\sum_{i=1}^N \frac{1}{L_i} \sum_{j=1}^{L_i} \log P(w_j|x_i,q_i,y_{i,<j}),    
\end{equation}
where $L_i$ is the length of the caption $y_i=\{w_1, w_2, \dots, w_{L_i} \}$, $y_{i,<j}$ denotes the masked sequence that contain only the first $j-1$ words, and $P(w_j|x_i,q_i,y_{i,<j})$ is the conditional probability of generating the caption word $w_j$ given the image $x_i$, question $q_i$ and the previous words $y_{i,<j}$ in the caption. During pre-training, the vision encoder and large language model are typically kept frozen, with only the parameters of the connector being trainable.

In \cite{liu2024visual,zhu2023minigpt}, the authors demonstrated that pre-training with visual instruction tuning can align vision and language representations while preserving extensive knowledge. Since then, advances have been made by modifying network architectures and creating high-quality pre-training datasets~\cite{muhtar2024lhrs,pang2024vhm}. In addition to improving alignment, another line of research focuses on supervised fine-tuning, aiming to enable the model to perform a variety of remote sensing image analysis tasks and interact with users in a conversational manner. This includes efforts to generate task-specific instruction-following data~\cite{kuckreja2024geochat,luo2024skysensegpt,zhan2024skyeyegpt}, design novel training strategies~\cite{pang2024vhm,zhang2024earthmarker}, and incorporate cutting-edge vision encoders~\cite{zhang2024earthmarker}.

\paragraph*{\textbf{Text-Conditioned Image Generation}} Taking advantage of the advances in conditional image generation, a group of works~\cite{zhang2024rs5m,khanna2023diffusionsat,tang2024crs} uses off-the-shelf generative models, primarily Stable Diffusion~\cite{rombach2022high}, to generate remote sensing images given text prompts, which essentially learns an implicit joint distribution between images and texts. As illustrated in Fig.\ref{fig:taxonomy-VLM} (c), their network architecture comprises three main components: a text encoder, a variational autoencoder (VAE)~\cite{kingma2013auto}, and a denoising U-Net. During training, the VAE encoder first transforms an image into a latent representation $z_0$. Gaussian noise $\epsilon$ is then added to this latent representation at different timesteps $t$, resulting in $z_t=\sqrt{\bar{\alpha}_t}z_0+\sqrt{1-\bar{\alpha}_t}\epsilon$, where $\bar{\alpha}_t$ is a time-dependent scaling factor. Next, the conditioning representation $c$, extracted from the text encoder, is provided as input alongside $z_t$ to the denoising U-Net, which predicts the noise added at timestep $t$. Finally, the VAE decoder upsamples the denoised latent representation to reconstruct the input image. Based on image-text pairs, the training objective is defined as:
\begin{equation}
\mathcal{L}_{\mathrm{LDM}}=\mathbb{E}_{z_0,c,\epsilon,t}\left[\Vert \epsilon-\epsilon_\theta(z_t,t,c)\Vert_2^2\right],
\end{equation}
where $\epsilon_\theta(z_t,t,c)$ is the model's predicted noise, parameterized by $\theta$. The application of diffusion models in remote sensing has shown rapid development, encompassing areas such as remote sensing image generation, enhancement, and interpretation~\cite{liu2024diffusion}. As this paper focuses on vision-language modeling using diffusion models, we do not attempt to cover every instance of their application in remote sensing tasks. Instead, we specifically highlight works that integrate text conditions with diffusion models. For a more comprehensive overview of the advancements in diffusion models for remote sensing, please refer to \cite{liu2024diffusion}.

Based on the fundamental principles outlined above, two major research groups have emerged. The first group aims to develop generative foundation models for various remote sensing images, including satellite~\cite{khanna2023diffusionsat}, aerial~\cite{arrabi2024cross}, hyperspectral~\cite{pang2024hsigene}, and multi-resolution images~\cite{yu2024metaearth}. The second group, in contrast, extends text-conditioned diffusion models to specific remote sensing tasks, such as image or change captioning~\cite{cheng2024vcc,yu2024diffusion}, pansharpening~\cite{xing2024empower} and zero-shot target recognition~\cite{wang2024leveraging}.

\begin{table*}[ht]
\centering
\caption{Summary of contrastive-based vision-language foundation models in remote sensing.}
\label{tab:contrastive-VLFM}
\resizebox{\textwidth}{!}{
\rowcolors{2}{gray!6}{white}
\begin{tabular}{lcccclc}
\toprule
\textbf{Model} & \textbf{Vision Encoder} & \textbf{Text Encoder} & \textbf{Training Dataset} & \textbf{Pre-training Objective} & \textbf{Evaluation Task} & Public \\ \midrule
GeoRSCLIP~\cite{zhang2024rs5m} \href{https://huggingface.co/Zilun/GeoRSCLIP}{[link]} & ViT-B/32, ViT-H/14~\cite{dosovitskiy2020image} & Transformer~\cite{vaswani2017attention} & RS5M~\cite{zhang2024rs5m} & InfoNCE loss~\cite{oord2018representation}  &  \begin{tabular}[l]{@{}l@{}}Zero-Shot Scene Classification \\ Image-Text Retrieval \\ Semantic Localization \end{tabular} & \ding{51} \\ 

RemoteCLIP~\cite{liu2024remoteclip} \href{https://github.com/ChenDelong1999/RemoteCLIP}{[link]} & \begin{tabular}[c]{@{}c@{}} ResNet-50~\cite{he2016deep}, \\ ViT-B/32, ViT-L/14~\cite{dosovitskiy2020image} \end{tabular} & Transformer~\cite{vaswani2017attention} & RET-3+SEG-4+DET-10~\cite{liu2024remoteclip} &  InfoNCE loss~\cite{oord2018representation} & \begin{tabular}[l]{@{}l@{}} Zero/Few-Shot Scene Classification \\ Image-Text Retrieval \\ Object Counting \\ Linear/$k$-NN Classification \end{tabular} & \ding{51} \\ 

SkyCLIP~\cite{wang2024skyscript} \href{https://github.com/wangzhecheng/SkyScript}{[link]} & ViT-B/32, ViT-L/14~\cite{dosovitskiy2020image} & Transformer~\cite{vaswani2017attention} & SkyScript~\cite{wang2024skyscript} & InfoNCE loss~\cite{oord2018representation} & \begin{tabular}[l]{@{}l@{}} Zero-Shot Scene Classification \\ Image-Text Retrieval \\ Zero-Shot Fine-Grained Classification \end{tabular} & \ding{51} \\

S-CLIP~\cite{mo2023s} \href{https://github.com/alinlab/s-clip}{[link]} & \begin{tabular}[c]{@{}c@{}} ResNet-50~\cite{he2016deep}, \\ ViT-B/32, ViT-B/16~\cite{dosovitskiy2020image} \end{tabular} & Transformer~\cite{vaswani2017attention} & \begin{tabular}[c]{@{}c@{}}RS-ALL~\cite{mo2023s} \\ NWPU-RESISC45~\cite{cheng2017remote} \end{tabular}& \begin{tabular}[c]{@{}c@{}} InfoNCE loss~\cite{oord2018representation} \\ Pseudo-label losses~\cite{mo2023s} \end{tabular} & \begin{tabular}[l]{@{}l@{}} Zero-Shot Scene Classification \\ Image-Text Retrieval \end{tabular} & \ding{51} \\

Set-CLIP~\cite{song2024set} & ResNet~\cite{he2016deep}& Transformer~\cite{vaswani2017attention} & RS-ALL~\cite{mo2023s} & \begin{tabular}[c]{@{}c@{}} InfoNCE loss~\cite{oord2018representation} \\ SSL loss~\cite{chen2020simple} \\ MK-MMD, SDD losses~\cite{song2024set} \end{tabular} & \begin{tabular}[l]{@{}l@{}} Zero-Shot Scene Classification \\ Image-Text Retrieval\end{tabular} & \ding{55} \\

GRAFT~\cite{mall2024remote} \href{https://graft.cs.cornell.edu/}{[link]} & ViT-B/16~\cite{dosovitskiy2020image} & - & \begin{tabular}[c]{@{}c@{}} Internet-NAIP Dataset~\cite{mall2024remote} \\ Internet-Sentinel-2 Dataset~\cite{mall2024remote} \end{tabular} & \begin{tabular}[c]{@{}c@{}} Image/Pixel-level \\ contrastive losses~\cite{mall2024remote} \end{tabular}  & \begin{tabular}[l]{@{}l@{}} Zero-Shot Scene Classification \\ Zero-Shot Text-to-Image Retrieval \\ Zero-Shot Semantic Segmentation \\ Zero-Shot Visual Question Answering\end{tabular} & \ding{51} \\

CSP~\cite{mai2023csp} \href{https://gengchenmai.github.io/csp-website/}{[link]} & ResNet-50~\cite{he2016deep} & Sinusoidal transform+FcNet~\cite{mai2020multi} & fMoW~\cite{christie2018functional} & InfoNCE loss~\cite{oord2018representation} & Few-Shot Geo-Aware Scene Classification & \ding{51} \\

GeoCLIP~\cite{cepeda2023geoclip} \href{https://github.com/VicenteVivan/geo-clip}{[link]} & ViT-L/14~\cite{dosovitskiy2020image} & Random Fourier features+MLP~\cite{tancik2020fourier} & MP-16~\cite{larson2017benchmarking} & \begin{tabular}[c]{@{}c@{}}InfoNCE loss~\cite{oord2018representation} \\ SSL loss~\cite{chen2020simple} \end{tabular}  & \begin{tabular}[l]{@{}l@{}} Image Geo-Localization \\ Text-Query Geo-Localization \\ Geo-Aware Image Classification \end{tabular} & \ding{51} \\ 

SatCLIP~\cite{klemmer2023satclip} \href{https://github.com/microsoft/satclip}{[link]} & \begin{tabular}[c]{@{}c@{}} ResNet-18, ResNet-50~\cite{he2016deep}, \\ ViT-S/16~\cite{dosovitskiy2020image} \end{tabular}& Spherical harmonics+Siren~\cite{russwurm2024geographic} & S2-100K~\cite{klemmer2023satclip} & InfoNCE loss~\cite{oord2018representation} & \begin{tabular}[l]{@{}l@{}} Air Temperature Prediction \\ Elevation Prediction \\ Median Income Estimation \\ California Housing Price Estimation \\ Population Density Estimation \\ Biome Classification \\ Ecoregion Classification \\ Country Code Classification \\
Species Recognition\end{tabular} & \ding{51} \\

PIR-CLIP~\cite{pan2024pir} \href{https://github.com/jaychempan/PIR-CLIP}{[link]} & ViT-B/32~\cite{dosovitskiy2020image}+ResNet-50~\cite{he2016deep} & Transformer~\cite{vaswani2017attention} & RS5M~\cite{zhang2024rs5m} & \begin{tabular}[c]{@{}c@{}} InfoNCE loss~\cite{oord2018representation} \\ Affiliation loss~\cite{pan2024pir} \end{tabular} & Image-Text Retrieval& \ding{51} \\

\bottomrule
\end{tabular}
}

\vspace{3pt}
\begin{minipage}{\textwidth}
    \footnotesize \textcolor{blue}{[link]} directs to model websites. Image-text retrieval involves image-to-text and text-to-image retrieval tasks. In \textit{Vision Encoder} column, we separate different architectures of vision encoders with commas, whereas in \cite{pan2024pir}, the vision encoder is a combination of two architectures. \textit{Public} refers to the availability of both code and model weights. S-CLIP has only open-sourced its code.
\end{minipage}
\end{table*}

\section{Contrastive-based Vision-Language Modeling}
Most existing works on VLM fall into the group of employing contrastive learning. As mentioned previously, there are two main research directions being actively investigated, namely \textit{foundation model construction} and \textit{effective adaptation}. Specifically, foundation model construction concerns the large domain gap between natural and remote sensing images, aiming to learn visual representations with rich remote sensing scene semantics and well-aligned with textual representations. On the other hand, effective adaptation answers the question of how to leverage pre-trained CLIP models for specific remote sensing tasks. In the following sections, we analyze the existing works from these two directions. 

\subsection{Foundation Model Construction}
Table~\ref{tab:contrastive-VLFM} summarizes contrastive-based vision-language models in remote sensing. To build foundation models, three key components need to be carefully designed: training datasets, pre-training objectives, and encoder architectures.

\paragraph*{\textbf{Training Datasets}} Large-scale image-text datasets form the basis for constructing foundation models. Ready-made image-text datasets in remote sensing, \emph{e.g.} UCM-captions and Sydney-captions, suffer from limited data volume and insufficient image diversity, rendering them inadequate for pre-training models to capture general knowledge of the domain. Recognizing the availability of numerous remote sensing image datasets, \cite{zhang2024rs5m} and \cite{liu2024remoteclip} use open-source datasets as the image source and develop image captioning methods to generate corresponding textual descriptions. Notably, \cite{zhang2024rs5m} filter 11 commonly-used image-text datasets using RS-related keywords and caption 3 large-scale RS image datasets (Million-AID~\cite{long2021creating}, fMoW~\cite{christie2018functional}, and BigEarthNet~\cite{sumbul2019bigearthnet}) with the aid of the tuned BLIP2 model~\cite{li2023blip}, resulting in the RS5M dataset, which contains over 5 million image-text pairs. Off-the-shelf vision-language models are indeed powerful tools for building large-scale image-text datasets due to their availability and ease of use, but ensuring captioning accuracy remains a significant challenge. To address this, \cite{liu2024remoteclip} proposes a rule-based method called mask-to-box and box-to-caption, which converts pixel-wise or bounding box annotations into natural language captions. Another concern is that the semantic diversity of the generated captions is constrained by the limited number of predefined classes in open-source remote sensing image datasets. Given this, Wang et al.~\cite{wang2024skyscript} attempt to leverage rich semantic information contained in OpenStreetMap, allowing the textual descriptions to encompass not only a wide variety of object categories but also fine-grained subcategories and object attributes. Similar to \cite{liu2024remoteclip}, captions are assembled from object tags following predefined rules.

\paragraph*{\textbf{Pre-training Objectives}} Instead of creating large-scale datasets, several works~\cite{mo2023s,mall2024remote,song2024set} explore new training objectives to facilitate model pre-training with few remote sensing image-text pairs. For instance, S-CLIP~\cite{mo2023s} introduces caption-level and keyword-level pseudo labels to fine-tune the original CLIP model on massive unpaired remote sensing images alongside a few image-text pairs. Denote a large number of unpaired images as $\{u_i \}_{i=1}^{M}$ ($M \gg N$). The caption-level pseudo-label $q_i^c \in \mathbb{R}^N$ is based on the assumption that the semantics of an unpaired image $u_i$ can be represented as a combination of those of paired images. Thus, $q_i^c$ represents a probability distribution over the captions of $N$ paired images, derived from the relationships between unpaired and paired images, which are formulated as an optimal transport problem. The keyword-level pseudo label $q_i^k \in \mathbb{R}^K$ relies on the assumption that $u_i$ shares keywords with visually similar images, representing the similarity between the embeddings of $u_i$ and the keywords (drawn from the nearest paired image) $\{k_j\}_{j=1}^K$. The loss functions for two pseudo-labels are defined as:
\begin{align}
    & \mathcal{L}_{\mathrm{caption}}=-\frac{1}{M} \sum_{i=1}^M \sum_{j=1}^N q_{ij}^c \log \sigma_\tau(s_{ij}^U,\{s_{il}^U\}_{l=1}^N), \label{eq:caption-loss} \\
    & \mathcal{L}_{\mathrm{keyword}}=-\frac{1}{M} \sum_{i=1}^M \sum_{j=1}^K q_{ij}^k \log \sigma_\tau(s_{ij}^U,\{s_{il}^U\}_{l=1}^K). \label{eq:keyword-loss} 
\end{align}
In Eq.\eqref{eq:caption-loss}, $s_{il}^U=\mathcal{S}(\mathbf{e}_i^U, \mathbf{e}_l^T)$ refers to the similarity score of the unpaired image embedding $\mathbf{e}_i^U$ and the caption embedding $\mathbf{e}_l^T$ of the paired image. Meanwhile, in Eq.\eqref{eq:keyword-loss},  $s_{il}^U=\mathcal{S}(\mathbf{e}_i^U, \mathbf{e}_l^k)$ denotes the similarity score between $e_i^U$ and the keyword embedding $\mathbf{e}_l^k$. Since few image-text pairs are involved in the fine-tuning process, the overall training objective also includes the InfoNCE loss, as shown in Eq.\eqref{eq:sclip-loss}.
\begin{equation}
    \mathcal{L}_{\mathrm{S-CLIP}} = \mathcal{L}_{\mathrm{InfoNCE}}+\frac{1}{2}(\mathcal{L}_{\mathrm{caption}} + \mathcal{L}_{\mathrm{keyword}}). \label{eq:sclip-loss}
\end{equation}

With a similar training data setup consisting of massive unpaired images and texts $\{u_i, v_i\}_{i=1}^M$ and limited image-text pairs $\{(x_i,y_i)\}_{i=1}^N$, Set-CLIP~\cite{song2024set} transforms the representation alignment between images and texts into a manifold matching problem, developing a multi-kernel maximum mean discrepancy loss $\mathcal{L}_{\mathrm{MK-MMD}}$ and a semantic density distribution loss $\mathcal{L}_{\mathrm{SDD}}$. The loss $\mathcal{L}_{\mathrm{MK-MMD}}$ constrains the consistency of whole representation distributions of images and texts, thereby achieving macro-level alignment, as formulated in Eq.\eqref{eq:mk-mmd-loss}.
\begin{equation}
\resizebox{0.9\linewidth}{!}{ $
    \mathcal{L}_{\mathrm{MK-MMD}}=\left\lVert \frac{1}{M+N} \sum_{i=1}^{M+N} \phi(\mathbf{e}_i^I) - \frac{1}{M+N} \sum_{i=1}^{M+N} \phi(\mathbf{e}_i^T) \right\rVert_{\mathcal{H}_{\mathrm{RKHS}}}^2, $} \label{eq:mk-mmd-loss}
\end{equation}
where $\phi(\cdot)$ is a linear combination of Gaussian and Polynomial kernel functions, used to map image and text embeddings $\mathbf{e}_i^I$, $\mathbf{e}_i^T$ into Reproducing Kernel Hilbert Space (RKHS) $\mathcal{H}_{\mathrm{RKHS}}$. Meanwhile, the loss $\mathcal{L}_{\mathrm{SDD}}$ refines the alignment between the two modalities by ensuring that their probability density distributions remain similar in the representation space. It can be formulated as:
\begin{equation}
    \mathcal{L}_{\mathrm{SDD}}=\frac{1}{2}(\Gamma(I,T)+\Gamma(T,I)),
\end{equation}
where $I=\{\mathbf{e}_i^I\}_{i=1}^{M+N}$ and $T=\{\mathbf{e}_i^T\}_{i=1}^{M+N}$ refer to embedding distributions of images and texts, respectively. $\Gamma(\cdot,\cdot)$ represents the Kullback-Leibler divergence, which measures the dissimilarity between two distributions. The format of $\Gamma(I,T)$ is given by:
\begin{equation}
\resizebox{0.9\linewidth}{!}{ $
    \Gamma(I,T)=\sum_{i=1}^{M+N}\frac{\kappa(\mathbf{e}_i^I,I)}{\sum_{j=1}^{M+N}\kappa(\mathbf{e}_j^I,I)} \log \frac{\kappa(\mathbf{e}_i^I,I)/\sum_{j=1}^{M+N}\kappa(\mathbf{e}_j^I,I)}{\kappa(\mathbf{e}_i^I,T)/\sum_{j=1}^{M+N}\kappa(\mathbf{e}_j^I,T)}, $ }
\end{equation}
with $\Gamma(T,I)$ defined similarly by swapping the roles of $I$ and $T$. Here, $\kappa(\cdot,\cdot)$ is an exponential probability density function that estimates the density value for each embedding in the corresponding distribution. Additionally, the self-supervised contrastive loss $\mathcal{L}_{\mathrm{SSL}}$ is introduced to obtain robust feature representations for each modality independently, defined as:
\begin{equation}
    \mathcal{L}_{\mathrm{SSL}}=-\frac{1}{M+N} \sum_{i=1}^{M+N} \log \frac{\mathrm{exp}(\mathcal{S}(\mathbf{e}_i, \mathbf{e}_i^+)/\tau)}{\sum_{j=1}^{M+N} \mathrm{exp}(\mathcal{S}(\mathbf{e}_i, \mathbf{e}_j)/\tau)},
\end{equation}
where $\mathbf{e}_i$ is an embedding, and $\mathbf{e}_i^+$ is the representation of the positive sample generated by augmentation techniques. Ultimately, the overall training objective is formulated as:
\begin{equation}
\resizebox{0.9\linewidth}{!}{ $
    \mathcal{L}_{\mathrm{Set-CLIP}}=\alpha \mathcal{L}_{\mathrm{InfoNCE}} + \mu \mathcal{L}_{\mathrm{SSL}} + \delta \mathcal{L}_{\mathrm{MK-MMD}} + \eta \mathcal{L}_{\mathrm{SDD}},$}
\end{equation}
where $\alpha$, $\mu$, $\delta$ and $\eta$ are tunable hyperparameters to balance the loss terms.

Even a small set of image-text pairs requires expert knowledge to craft textual descriptions. To avoid textual annotations entirely, GRAFT~\cite{mall2024remote} proposes utilizing co-located ground images as the bridge between satellite images and language. In doing so, a dataset of two million ground-satellite image pairs, denoted as $\{x_i,\{g_{ij}\}_{j=1}^{N_i}\}_{i=1}^N$, is collected to support model training. Building on this, a feature extractor is designed to map satellite images to the representation space of the CLIP model, which was trained on internet image-text pairs. Since a satellite image $x_i$ can cover a large ground area and thus be associated with multiple ground images $\{g_{ij}\}_{j=1}^{N_i}$, the extractor is optimized using the following loss:
\begin{equation}
\resizebox{0.9\linewidth}{!}{ $
    \mathcal{L}_{\mathrm{GRAFT}}=-\frac{1}{N} \sum_{i=1}^N \frac{1}{N_i} \sum_{j=1}^{N_i} \log \frac{\mathrm{exp}(\mathcal{S}(\mathbf{e}_i^I, \mathbf{e}_{ij}^G)/\tau)}{\sum_{a=1}^N\sum_{b=1}^{N_a}\mathrm{exp}(\mathcal{S}(\mathbf{e}_i^I, \mathbf{e}_{ab}^G)/\tau)}, $}
\end{equation}
where $\mathbf{e}_{ij}^G$ is the embedding of the $j_{th}$ ground image corresponding to the $i_{th}$ satellite image. This loss focuses solely on aligning image-level representations between the two types of images while ignoring the fact that a ground image $g_{ij}$ can be mapped to a specific pixel location $p_{ij}$ in the corresponding satellite image $x_i$. Thus, GRAFT trains the feature extractor with an additional pixel-level contrastive loss as follows:
\begin{equation}
\resizebox{0.9\linewidth}{!}{ $
    \mathcal{L}_{\mathrm{GRAFT}}=-\frac{1}{N} \sum_{i=1}^N \frac{1}{N_i} \sum_{j=1}^{N_i} \log \frac{\mathrm{exp}(\mathcal{S}(\mathbf{e}_{ij}^P, \mathbf{e}_{ij}^G)/\tau)}{\sum_{a=1}^N\sum_{b=1}^{N_a}\mathrm{exp}(\mathcal{S}(\mathbf{e}_{ij}^P, \mathbf{e}_{ab}^G)/\tau)}, $}
\end{equation}
where $\mathbf{e}_{ij}^P$ denotes the feature vector for pixel $p_{ij}$.

\paragraph*{\textbf{Encoder Architectures}} Unlike natural images, where corresponding textual information typically describes the image content, the textual information for remote sensing images can be represented by their geographic coordinates (longitude and latitude), which are beneficial for tasks such as scene classification and object recognition~\cite{christie2018functional}. Therefore, some works propose aligning representations from remote sensing images and their geographic coordinates, with particular attention given to the choice of location encoders~\cite{mai2023csp,cepeda2023geoclip,klemmer2023satclip}. A location encoder generally consists of a nonparametric functional positional encoding combined with a small neural network. In CSP~\cite{mai2023csp}, an existing 2D location encoder, namely Space2Vec's grid proposed in \cite{mai2020multi}, is utilized. This encoder maps image coordinates into high-dimensional representations using sinusoid transforms for position encoding, followed by fully connected ReLU layers. GeoCLIP~\cite{cepeda2023geoclip}, on the other hand, first applies equal earth projection to the image coordinates to reduce distortions inherent in standard geographic coordinate systems. It then adopts random Fourier features~\cite{tancik2020fourier} to capture high-frequency details, varying the frequency to construct hierarchical representations. These hierarchical representations are processed through separate multi-layer perceptions (MLPs), followed by element-wise addition, resulting in a joint representation. This design enables the location encoder to effectively capture features of a specific location across multiple scales. In SatCLIP~\cite{klemmer2023satclip}, the authors use a location encoder that combines spherical harmonics basis functions with sinusoidal representation networks to map geographic coordinates into latent representations~\cite{russwurm2024geographic}.

In addition to adapting encoders for different types of textual information, another purpose of adjusting encoders is to improve the representations of visual concepts. Remote sensing images typically cover a large field of view and include a variety of objects. However, the corresponding textual descriptions often center on specific objects of interest and their relationships. Semantic noise, such as irrelevant objects and background, can interfere with the representation of key content in images, thereby obstructing the alignment between image and text representations. In \cite{pan2024pir}, this problem is addressed by using prior knowledge of remote sensing scenes to instruct the pre-trained model in filtering out semantic noise before calculating the similarity between image and text embeddings. This is practically achieved by adding an instruction encoder and a transformer encoder layer on top of the vision encoder. The instruction encoder, pre-trained on the scene classification dataset AID~\cite{xia2017aid}, generates instruction embeddings, which are used to filter image embeddings via a soft belief strategy. The filtered embeddings are then activated using instruction information through the transformer encoder layer, producing relevant image embeddings for subsequent alignment.

\begin{table*}[ht]
\centering
\caption{Performance of contrastive-based vision-language foundation models.}
\label{tab:performance-contrastive-VLFM}
\resizebox{\textwidth}{!}{
\rowcolors{4}{gray!6}{white}
\begin{tabular}{lccccccccccccccccc}
\toprule
\multicolumn{18}{c}{\textbf{Zero-Shot Scene Classification: Top-1 Accuracy}} \\ \cmidrule{1-18}
\textbf{Model} & \makecell{\textbf{Vision} \\ \textbf{Encoder}} & \makecell{AID \\ \cite{xia2017aid}} & \makecell{EuroSAT \\ \cite{helber2019eurosat}} & \makecell{fMoW \\ \cite{christie2018functional}} & \makecell{Million-AID \\ \cite{long2021creating}} & \makecell{MLRSNet \\ \cite{qi2020mlrsnet}} & \makecell{RESISC \\ \cite{cheng2017remote}} & \makecell{Optimal-31 \\ \cite{wang2018scene}} & \makecell{PatternNet \\ \cite{zhou2018patternnet}} & \makecell{RSSCN7 \\ \cite{zou2015deep}} & \makecell{RSC11 \\ \cite{zhao2016feature}} & \makecell{RSI-CB128 \\ \cite{li2020rsi}} & \makecell{RSI-CB256 \\ \cite{li2020rsi}} & \makecell{RSICD \\ \cite{lu2017exploring}} & \makecell{SIRI-WHU \\ \cite{zhao2015dirichlet}} & \makecell{UCM \\ \cite{yang2010bag}} & \makecell{WHU-RS19 \\ \cite{xia2010structural} }\\ \midrule
SkyCLIP~\cite{wang2024skyscript} & ViT-L/14~\cite{dosovitskiy2020image}    & 71.70 & 51.33  & \textbf{27.12} &  \textbf{67.45} & - & 70.94 & - & \textbf{80.88} & - & - & - & 50.09 & - & - & - & - \\
Set-CLIP~\cite{song2024set} & ResNet-50~\cite{he2016deep}        & 76.20 & - & - & - & - & - & - & - & 66.20 & - & - & - & 69.20 & - & 67.50 & 89.00 \\
GRAFT~\cite{mall2024remote} & ViT-B/16~\cite{dosovitskiy2020image} & - & 63.76 & - & - & - & - & - & - & - & - & - & - & - & - & - & - \\ 
GeoRSCLIP~\cite{zhang2024rs5m} & ViT-H/14~\cite{dosovitskiy2020image}      & 76.33 & \textbf{67.47} & - & - & - & 73.83 & - & - & - & - & - & - & - & - & -& - \\
S-CLIP~\cite{mo2023s} & ViT-B/16~\cite{dosovitskiy2020image}               & 85.20 & - & - & - & - & - & - & - & \textbf{76.30} & - & - & - & \textbf{79.50} & - & \textbf{82.30} & 93.90 \\
RemoteCLIP~\cite{liu2024remoteclip} & ViT-L/14~\cite{dosovitskiy2020image} & \textbf{87.90} & 59.94 & - & - & \textbf{66.32} & \textbf{79.84} & \textbf{90.05} & 68.75 & 72.32 & \textbf{74.90} &  \textbf{37.22} & \textbf{52.82} & - & \textbf{70.83} & - & \textbf{94.66} \\ \midrule

\multicolumn{18}{c}{\cellcolor{white} \textbf{Image-Text Retrieval: Recall@1,  Recall@5, and Recall@10} } \\ \cmidrule{1-18}
\multirow{2}{*}{\textbf{Model}} & \textbf{Vision} & \multicolumn{4}{c}{RSICD~\cite{lu2017exploring}} & \multicolumn{4}{c}{RSITMD~\cite{yuan2022exploring}} & \multicolumn{4}{c}{UCM-captions~\cite{qu2016deep}} & \multicolumn{4}{c}{Sydney-captions~\cite{qu2016deep}} \\ \cmidrule{3-18}
  & \textbf{Encoder} & R@1 & R@5 & R@10 & Mean & R@1 & R@5 & R@10 & Mean & R@1 & R@5 & R@10 & Mean & R@1 & R@5 & R@10 & Mean \\ \cmidrule{1-18}
\multicolumn{18}{l}{\cellcolor{white} Image-to-Text Retrieval} \\ \midrule
S-CLIP~\cite{mo2023s} & ResNet-50~\cite{he2016deep} & 4.20 & 18.40 & - & - & - & - & - & - & 11.60 & 45.70 & - & - & \textbf{14.90 }& 50.00 & - & - \\
Set-CLIP~\cite{song2024set} & ResNet-50~\cite{he2016deep} & - & 19.60 & - & - & - & - & - & - & - & 46.30 & - & - & - & \textbf{51.10} & - & -  \\ 
SkyCLIP~\cite{wang2024skyscript}$\dag$ & ViT-L/14~\cite{dosovitskiy2020image} & - & - & - & 23.70 & - & - & - & 30.75 & - & - & - & \textbf{72.22} & - & - &- & - \\
RemoteCLIP~\cite{liu2024remoteclip} & ViT-L/14~\cite{dosovitskiy2020image} & 18.39 & 37.42 & 51.05 & 35.62 & 28.76 & 52.43 & 63.94 & 48.38 & \textbf{19.05} & \textbf{54.29} & \textbf{80.95} & 51.43 & - & - &- & - \\
GeoRSCLIP~\cite{zhang2024rs5m} & ViT-B/32~\cite{dosovitskiy2020image} & 21.13 & 41.72 & \textbf{55.63} & 39.49 & 32.30 & 53.32 & 67.92 & 51.18 & - & - &- & - & - & - &- & - \\ 
PIR-CLIP~\cite{pan2024pir} & \begin{tabular}[c]{@{}c@{}} ViT-B/32~\cite{dosovitskiy2020image} \\ +ResNet-50~\cite{he2016deep} \end{tabular}  & \textbf{27.63} & \textbf{45.38} & 55.26 & \textbf{42.76} & \textbf{45.58} & \textbf{65.49} & \textbf{75.00} & \textbf{62.02} & - & - &- & - & - & - &- & - \\ \midrule

\multicolumn{18}{l}{\cellcolor{white} Text-to-Image Retrieval} \\ \midrule
S-CLIP~\cite{mo2023s} & ResNet-50~\cite{he2016deep} & 4.20 & 16.80 & - & - & - & - & - & - & 11.1 & 43.50 & - & - & \textbf{17.8} & 55.10 & - & - \\
Set-CLIP~\cite{song2024set} & ResNet-50~\cite{he2016deep} & - & 17.40  & - & - & - & - & - & - & - & 44.10 & - & - & - & \textbf{55.20} & - & - \\
SkyCLIP~\cite{wang2024skyscript} $\dag $ & ViT-L/14~\cite{dosovitskiy2020image} & - & - & - & 19.97 & - & - & - & 30.58 & - & - & - & \textbf{59.33} & - & - &- & - \\
RemoteCLIP~\cite{liu2024remoteclip} & ViT-L/14~\cite{dosovitskiy2020image} & 14.73 & 39.93 & 56.58 & 37.08 & 23.76 & \textbf{59.51} & \textbf{74.73} & \textbf{52.67} & \textbf{17.71} & \textbf{62.19} & \textbf{93.90} & 57.93 & - & - &- & - \\
GeoRSCLIP~\cite{zhang2024rs5m} & ViT-B/32~\cite{dosovitskiy2020image} & 15.59 & 41.19 & \textbf{57.99} & 38.26 & 25.04 & 57.88 & 74.38 & 52.43 & - & - &- & - & - & - &- & -\\ 
PIR-CLIP~\cite{pan2024pir} & \begin{tabular}[c]{@{}c@{}} ViT-B/32~\cite{dosovitskiy2020image} \\ +ResNet-50~\cite{he2016deep} \end{tabular} & \textbf{21.10} & \textbf{44.87} & 56.12 & \textbf{40.70} & \textbf{30.13} & 55.44 & 68.54 & 51.37 & - & - &- & - & - & - &- & - \\
\bottomrule
\end{tabular}
}

\vspace{3pt}
\begin{minipage}{\textwidth}
    \footnotesize \textit{RESISC} is short for NWPU-RESISC45~\cite{cheng2017remote}. \textit{Mean} refers to the mean recall averaged of recall@1 (\textit{R@1}), recall@5 (\textit{R@5}) and recall@10 (\textit{R@10}). $\dag$ indicates that the training data corresponding to the test dataset was not used to pre-train the model.
\end{minipage}
\end{table*}

\paragraph*{\textbf{Performance Evaluation}} Zero-shot scene classification and image-text retrieval are commonly used tasks to assess foundation models' capabilities to capture a wide range of visual and textual concepts, along with their correspondence. Table~\ref{tab:performance-contrastive-VLFM} lists the performance of contrastive-based vision-language foundation models on these two tasks. Note that it shows the best model performance in the original papers. For zero-shot scene classification, three conclusions can be drawn: 1) AID~\cite{xia2017aid} is the most widely used dataset for evaluation, with RemoteCLIP achieving the best accuracy of 87.9\%; 2) Using the same vision encoder (ViT-L/14), SkyCLIP, which is pre-trained on images collected from Google Earth, outperforms RemoteCLIP only on PatternNet~\cite{zhou2018patternnet}, a dataset sourced from Google Earth; 3) Comparing models pre-trained on limited image-text pairs (\emph{i.e.} S-CLIP, SetCLIP, and GRAFT) with RemoteCLIP, their performance is somewhat comparable, showing the importance of exploring data-effective pre-training. For cross-modal retrieval, PIR-CLIP achieves state-of-the-art performance on RSICD~\cite{lu2017exploring} and RSITMD~\cite{yuan2022exploring}. In comparison, S-CLIP and SetCLIP perform significantly worse, with a performance gap exceeding 20\%. This suggests that existing foundation models pre-trained on limited image-text pairs are capable of acquiring general visual and textual concepts but may face challenges in learning the relationships between them.

\begin{table*}[ht]
\centering
\caption{Summary of contrastive-based vision-language foundation models applied to remote sensing tasks.}
\label{tab:application-contrastive-VLFM}
\resizebox{\textwidth}{!}{
\rowcolors{2}{gray!6}{white}
\begin{tabular}{llll}
\toprule
\textbf{Work} & \textbf{Task} & \textbf{Vision-Language Foundation Model} & \textbf{Adaptation} \\ \midrule
VLCA~\cite{wei2023vlca} & Image Captioning & CLIP ResNet-50, ViT~\cite{radford2021learning} &  \makecell[{{p{9cm}}}]{Utilize external attention~\cite{guo2022beyond} to enhance visual representations extracted by CLIP's vision encoder.} \\
MVP~\cite{wang2024multi} &  Image Captioning & CLIP ViT-B/16~\cite{radford2021learning} & \makecell[{{p{9cm}}}]{Fuse visual representations from CLIP and pre-trained vision models.} \\
BITA~\cite{yang2024bootstrapping} & Image Captioning & CLIP ViT-L/14~\cite{radford2021learning}  & \makecell[{{p{9cm}}}]{Introduce the Fourier-based transformer to align images and texts in the remote sensing domain.} \\ 
MGIMM~\cite{yang2024mgimm} & Image Captioning & CLIP ViT-L/14~\cite{radford2021learning} & \makecell[{{p{9cm}}}]{Region-level and image-level instruction tuning for generating captions with detailed geographical information.}  \\
CFD-C~\cite{ricci2024machine} & Image Captioning & CLIP ViT-B/16~\cite{radford2021learning} & \makecell[{{p{9cm}}}]{Define image captioning as the aggregation of information from multi-turn dialogues} \\
Prompt-CC~\cite{liu2023decoupling} & Change Captioning & CLIP ViT-B/32~\cite{radford2021learning} & \makecell[{{p{9cm}}}]{Develop an image-level classifier, a feature-level encoder on top of pre-trained vision encoder for detecting the presence of changes and extracting discriminative features.}   \\ \midrule
CISEN~\cite{zhao2024luojiahog} & Image-Text Retrieval & \makecell[l]{\gape CLIP ResNet-50, ViT-B/32+Transformer~\cite{radford2021learning} \\ GeoRSCLIP ViT-B/32+Transformer~\cite{zhang2024rs5m}} & \makecell[{{p{9cm}}}]{Two-stage training to progressively fuse fine-grained semantic features to enrich visual representation.}  \\
KTIR~\cite{mi2024knowledge} & Image-Text Retrieval & BLIP ViT-B/16+BERT~\cite{li2022blip} & \makecell[{{p{9cm}}}]{Incorporate knowledge into text features for better image-text alignment.} \\
Zavras et al.~\cite{zavras2024mind} & \begin{tabular}[l]{@{}l@{}} Zero-Shot RGB-Multispectral Retrieval \\ Zero-Shot Scene Classification \end{tabular} & \begin{tabular}[l]{@{}l@{}} CLIP ViT-B/32, ViT-B/16, \\ ViT-L/14, ViT-L/14@336~\cite{radford2021learning} \end{tabular}
& \makecell[{{p{9cm}}}]{ Two-stage fine-tuning to achieve alignment between distinct image modalities in the CLIP representation space.}  \\
WEICOM~\cite{psomas2024composed} & Composed-to-Image Retrieval & \makecell[l]{\gape CLIP ViT-L/14+Transformer~\cite{radford2021learning}\\ RemoteCLIP ViT-L/14+Transformer~\cite{liu2024remoteclip} }& \makecell[{{p{9cm}}}]{A weighting parameter for combining image-query and text-query retrieval results, achieving composed query.}\\ \midrule
Rahhal et al.~\cite{al2023vision} & Zero-Shot Scene Classification & Thirteen CLIP/Open-CLIP Models~\cite{radford2021learning} & \makecell[{{p{9cm}}}]{Tuning prompts by adding task-relevant context words, including ``remote sensing image" and ``scene".} \\
Lan et al.~\cite{lan2024efficient} & Few-Shot Fine-Grained Ship Classification & CLIP ResNet-50, ViT-B/16+Transformer~\cite{radford2021learning} & \makecell[{{p{9cm}}}]{ Inject domain priors to adapt frozen vision encoder to remote sensing scenes, \\ Design hierarchical, learnable prompts to capture rich task-specific knowledge. } \\
RS-CLIP~\cite{li2023rs}  & Zero-Shot Scene classification & CLIP ViT-L/14+Transformer~\cite{tay2017learning} & \makecell[{{p{9cm}}}]{Use pseudo labeling and curriculum learning to fine-tune the pre-trained model in multiple rounds.}  \\
DSVA~\cite{xu2023deep} & Zero-Shot Scene Classification & CLIP ViT-B/32+Transformer~\cite{radford2021learning} & \makecell[{{p{9cm}}}]{Employ the pre-trained model to annotate attributes for each class.} \\ \midrule
Text2Seg~\cite{zhang2023text2seg} & Zero-Shot Semantic Segmentation & CLIP ViT-B/16+Transformer~\cite{radford2021learning}  & \makecell[{{p{9cm}}}]{Employ CLIP to endow segmentation masks with semantic categories.}  \\
Lin et al.~\cite{lin2024practical} & Semantic Segmentation & CLIP ViT-B/16+Transformer~\cite{radford2021learning} & \makecell[{{p{9cm}}}]{Modify CLIPSeg decoder~\cite{luddecke2022image} to receive CLIP joint image-text embeddings as input and generate segmentation masks.} \\ 
SegEarth-OV~\cite{li2024segearth} & Open-Vocabulary Semantic Segmentation & CLIP ViT-B/16+Transformer~\cite{radford2021learning} & \makecell[{{p{9cm}}}]{Develop SimFeatUp to robustly upsample low-resolution CLIP embeddings, \\ Execute subtraction operations between patch embeddings and the class token embedding to alleviate global bias.} \\
SCM~\cite{tan2024segment} & Unsupervised Change Detection & CLIP~\cite{radford2021learning} & \makecell[{{p{9cm}}}]{Employ CLIP to identify objects of interest in bi-temporal images, helping filter out pseudo changes.}\\
ChangeCLIP~\cite{dong2024changeclip} & Change Detection & CLIP ResNet-50, ViT-B/16+Transformer~\cite{radford2021learning} & \makecell[{{p{9cm}}}]{Employ CLIP to construct, encode multi-modal input data for change detection, \\ Design transformer-based decoder to combine vision-language features with image features, predicting change maps.} \\
BAN~\cite{li2024new} & Change Detection & \makecell[l]{\gape CLIP ViT-B/16, ViT-L/14~\cite{radford2021learning} \\ RemoteCLIP ViT-B/32, ViT-L/14~\cite{liu2024remoteclip} \\ GeoRSCLIP ViT-B/32, ViT-L/14~\cite{zhang2024rs5m} }  & \makecell[{{p{9cm}}}]{Design bridging modules to inject general knowledge from foundation models to existing change detection models.} \\ \midrule
Bazi et al.~\cite{bazi2022bi} & Visual Question Answering & CLIP ViT+Transformer~\cite{radford2021learning} & \makecell[{{p{9cm}}}]{Employ CLIP to encode images and questions.}   \\
Czerkawski et al.~\cite{czerkawski2023detecting}& Cloud Presence Detection & CLIP~\cite{radford2021learning} & \makecell[{{p{9cm}}}]{Apply CLIP to this task via tuning prompts or adding linear classifier.} \\
TGN~\cite{tang2024text} & Text-based Image Generation & CLIP~\cite{radford2021learning} & \makecell[{{p{9cm}}}]{Utilize CLIP to classify generated images, ensuring semantic class consistency.}  \\
AddressCLIP~\cite{xu2024addressclip} & Image Address Localization & CLIP ViT-B/16+Transformer~\cite{radford2021learning} & \makecell[{{p{9cm}}}]{ Align image with scene captions and address texts by contrastive learning, \\ Introduce image-geography matching to constrain image features with the spatial distance.}  \\
\bottomrule
\end{tabular}
}
\end{table*}

\subsection{Effective Adaptation Methods}
\label{contrastive-application}
Inspired by the outstanding performance of foundation models, numerous methods have been developed to effectively adapt pre-trained models for specific remote sensing tasks, including image captioning, cross-modal retrieval, zero-shot classification, dense prediction, etc. This section presents the development of adaptation methods within the context of different downstream tasks.

\paragraph*{\textbf{Image Captioning}} aims to describe the content of a given image using natural language. Vision-language foundation models are well-suited for this task, as they align image and text representations. One can use the vision encoder of these models to extract representations of image content, which prepares the input for language models to generate captions~\cite{wei2023vlca,wang2024multi,yang2024bootstrapping}. Commonly used vision encoders for remote sensing image captioning are from CLIP, while language models for this task include GPT-2~\cite{radford2019language}, BERT~\cite{kenton2019bert}, and OPT~\cite{zhang2022opt}. To improve caption accuracy, a few works are devoted to enhancing models' visual representation capabilities or further aligning images and texts. 

For example, regarding representation enhancement, VLCA~\cite{wei2023vlca} introduces external attention~\cite{guo2022beyond} to capture potential correlations between different images, improving visual representations. MVP~\cite{wang2024multi} fuses visual representations from pre-trained vision-language models and pre-trained vision models through stacked transformer encoders. It also leverages CLIP's vision encoder, followed by adaptive average pooling layers, to generate visual prefixes, which are concatenated with token embeddings. A BERT-based caption generator is subsequently developed to combine the fused visual representations and concatenated embeddings, enabling the generation of accurate captions. In contrast, BITA~\cite{yang2024bootstrapping} focuses on improving the alignment of images and texts in the remote sensing domain by introducing an interactive Fourier transformer. In this design, learnable visual prompts are fed to the Fourier layer and interact with image embeddings from a pre-trained vision encoder, capturing the most relevant visual representations. Through contrastive learning, these visual representations are aligned with textual representations, also extracted by the Fourier-based transformer. The interactive Fourier transformer then connects the frozen vision encoder with the frozen language model, leveraging the language model's generation and reasoning capabilities.

Beyond improving caption accuracy, generating detailed captions has also been explored~\cite{yang2024mgimm,ricci2024machine}. In~\cite{yang2024mgimm}, a two-stage instruction fine-tuning is proposed for the vision-to-language mapping layer to generate geographically detailed captions. The first stage, guided by the instruction ``[region] Based on the provided region of the remote sensing image, describe the basic attributes of the main objects in that region.", aligns geographic object regions with their attribute descriptions. The second stage, guided by the instruction ``[image] Please provide a detailed description of this image.", focuses on understanding the spatial distribution of geographic objects within images. In~\cite{ricci2024machine}, image captioning is defined as the aggregation of information from multi-turn dialogues, where each turn serves to query the image content. In each turn, CLIP's vision encoder extracts image features, which are then input into an auto-regressive language model~\cite{radford2019language} along with previous questions, answers, and the current question to generate the response. After several dialogue turns, GPT-3~\cite{brown2020language} summarizes the dialogue information to produce an enriched textual image description.

Unlike previous works that caption a single image, Prompt-CC~\cite{liu2023decoupling} utilizes pre-trained models to describe differences between bi-temporal images, a task known as change captioning. Based on bit-temporal visual representations from CLIP, Prompt-CC introduces an image-level classifier to detect the presence of changes, and a feature-level encoder to extract discriminative features that identify the specific changes.

\paragraph*{\textbf{Cross-Modal Retrieval}} is the task of retrieving data from one modality by using queries from another modality. Based on vision-language foundation models, there is ongoing research that explores retrieval between images and text~\cite{zhao2024luojiahog,mi2024knowledge,psomas2024composed}, as well as between images captured with different imaging parameters~\cite{zavras2024mind}. Image-text retrieval involves the challenges of finding textual descriptions that correspond to given image queries and vice versa. Taking advantage of advances in vision-language foundation models, existing works use pre-trained encoders to encode images and text separately, mapping them into a shared representation space for similarity measurement. Typically, CISEN~\cite{zhao2024luojiahog} adopts encoders from CLIP and GeoRSCLIP as the backbone for its retrieval network. Given the abundance of objects in remote sensing images, CISEN is trained in two stages to enhance visual representation. In the first stage, an image adapter~\cite{gao2024clip} is trained to map global visual features into textual-like features. In the second stage, a feature pyramid network is used to integrate the textual-like features into local visual features, which are then utilized to enrich global visual features. Rather than focusing on effective representation, KTIR~\cite{mi2024knowledge} aims to improve alignment by explicitly incorporating knowledge into text features. The knowledge is derived from textual information using off-the-shelf knowledge sources like RSKG~\cite{li2021robust} and ConceptNet~\cite{speer2017conceptnet}. Once converted to knowledge sentences, the knowledge is processed by the text encoder to extract knowledge features, which are then fused with text features via a single cross-attention layer.

Image-image retrieval involves searching for relevant images across different imaging parameters, such as matching RGB images with multispectral images, as explored in \cite{zavras2024mind}. The authors use a pre-trained CLIP text encoder as the classification head, and fine-tune the CLIP vision encoder alongside a newly added multispectral-specific encoder. The training is conducted in stages: first, the CLIP vision encoder is fine-tuned on RGB images composited from multispectral images, followed by fine-tuning the new encoder on multispectral images. This encoder is guided to produce discriminative representations that can achieve accurate image classification, while ensuring its representations are similar to those of RGB images extracted by the CLIP vision encoder.

\begin{figure}[t]
\centering
\includegraphics[width=\linewidth]{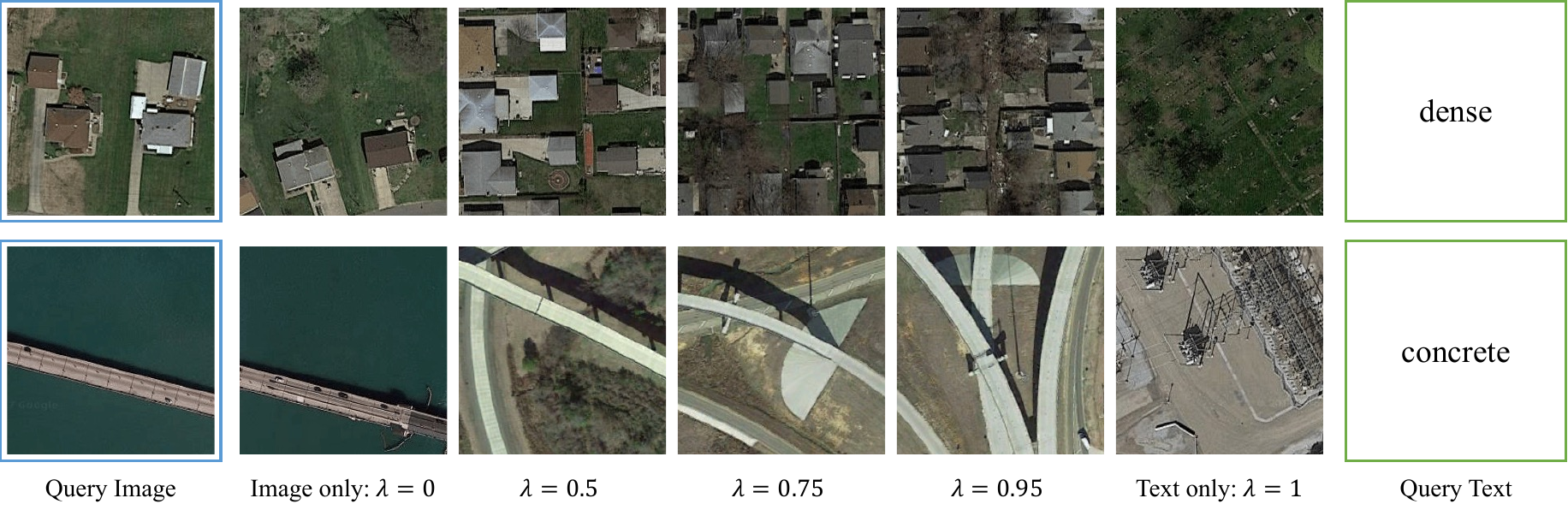}
\caption{Illustration of composed image retrieval for remote sensing~\cite{psomas2024composed}.}
\label{fig:composed-to-image}
\end{figure}

Considering the complexity of Earth's surface, single-modality queries, such as text queries, require users to fully articulate their needs to pinpoint relevant images. This raises the barrier to using retrieval models. To address this issue, Psomas et al.~\cite{psomas2024composed} introduce the composed-to-image retrieval task, which searches for remote sensing images based on a composed query of image and text. The retrieved images share the same scene or object category as the image query and reflect the attribute defined by the text query. To achieve this, the authors in \cite{psomas2024composed} propose calculating similarity scores for the image and text query separately, then normalizing and combining these scores using a convex combination controlled by a weighting parameter $\lambda$, which adjusts the contribution of each modality. As shown in Fig.~\ref{fig:composed-to-image}, when $\lambda=0$, composed-to-image retrieval simplifies to image-to-image retrieval, while at $\lambda=1$, it simplifies to text-to-image retrieval.

\paragraph*{\textbf{Zero-shot Scene Classification}} challenges models to identify remote sensing images from scene categories that were not seen during training. The idea of applying vision-language foundation models to this task is straightforward: since these models align images and texts, zero-shot scene classification can be achieved by comparing image embeddings with text embeddings extracted by the text encoder, which takes as input textual descriptions specifying the unseen classes. By default, these descriptions follow the format ``a photo of a [CLASS].", where the class token is replaced by the specific class name, such as ``farmland", ``forest" or ``playground". The input text, also known as prompt, plays a crucial role in the performance of foundation models on downstream tasks. As a result, several works~\cite{al2023vision,li2023rs} pay attention to prompt tuning and suggest adding task-relevant context words to improve performance. An example is provided by \cite{al2023vision}, where the authors replace ``photo" with terms such as ``remote sensing image", ``top view image", ``satellite image" and ``scene". This slight modification results in more than a 5\% increase in the accuracy of the CLIP model with the ViT-B/32 vision encoder on UCM dataset~\cite{yang2010bag}. However, as noted in other experiments from \cite{al2023vision}, prompts subjected to extensive tuning are not guaranteed to be optimal for the task and may not be suitable for other test datasets. To avoid manual prompt tuning, Lan et al.~\cite{lan2024efficient} model context words in a prompt as learnable vectors, which are combined with the class token embeddings before being input to the text encoder. Moreover, the class token embeddings for each category are organized across multiple levels of granularity, taking the form: ``a photo of a ship, the primary type is [CLASS1], secondary type is [CLASS2], final type is [CLASS3].", aimed at the task of few-shot fine-grained ship classification. Compared to hand-crafted prompts, these hierarchical, learnable prompts incorporate richer task-specific knowledge.

Most works employ and freeze pre-trained CLIP models. However, due to the significant domain gap between web images and remote sensing images, their performance tends to be limited. To mitigate this, one can inject remote sensing domain priors into the vision encoder~\cite{lan2024efficient}, or fine-tune the entire model using pseudo labeling techniques~\cite{li2023rs}. Typically, Lan et al.~\cite{lan2024efficient} introduce a lightweight network that is trained on data from seen classes to capture the domain prior. This prior is then combined with image embeddings output by the vision encoder, allowing the pre-trained model to adapt better. On the other hand, RS-CLIP~\cite{li2023rs} leverages the strong transferability of the pre-trained model to automatically generate pseudo labels from unlabeled images, which are used to fine-tune the model. Additionally, a curriculum learning strategy is developed to gradually select more pseudo-labeled images for model training in multiple rounds, further boosting the model's performance in zero-shot scene classification.

In addition to comparing the similarity between embeddings of images and unseen-class texts, DSVA~\cite{xu2023deep} introduces a solution that uses pre-trained models to annotate attributes for each scene class, and predict scene categories by evaluating the similarity between attribute values derived from image embeddings and those associated with scene classes. For automatic attribute annotation, textual descriptions have the form of ``This photo contains [ATTRIBUTE]", where the attribute token is replaced by specific attribute names, such as ``red", ``cement" or ``rectangle". Meanwhile, the attribute value for each class is calculated by measuring the similarity between embeddings of attribute text and images belonging to that class.

\paragraph*{\textbf{Dense Prediction Tasks}} such as semantic segmentation and change detection, which produce pixel-level predictions for input images, have recently benefited from the application of vision-language foundation models. Typically, Text2Seg~\cite{zhang2023text2seg} uses a CLIP model to classify category-agnostic segmentation masks generated by the segment anything model (SAM)~\cite{kirillov2023segment}, enabling zero-shot semantic segmentation of remote sensing images. Lan et al.~\cite{lin2024practical} instead modify the CLIPSeg decoder~\cite{luddecke2022image} to receive joint image-text embeddings from the CLIP model as input, producing a binary segmentation mask. The decoder is composed of transformer blocks, convolution layers, and linear interpolation layers. 

Without the need for additional decoders or segmentation models, one can perform an upsampling operation on image embeddings from CLIP, and compare the similarity between image patch embeddings and text embeddings to produce segmentation results. However, empirical findings in \cite{li2024segearth} suggest that for CLIP with a vision encoder based on ViT-B/16, the image embeddings are downsampled to 1/16 of the input image size, This downsampling leads to distorted object shapes and poorly fitting boundaries in segmentation masks. Furthermore, CLIP's self-attention causes global information from the class token embedding to be attached to the patch embeddings, which significantly degrades performance in semantic segmentation. To handle these issues, SegEarth-OV~\cite{li2024segearth} incorporates SimFeatUp on top of the CLIP vision encoder to restore lost spatial information in image embeddings. Subsequently, subtraction operations are executed between patch embeddings and the class token embeddings before similarity measurement, alleviating global bias in patch embeddings. In particular, SimFeatUp is a learnable upsampler consisting of a single parameterized JBU operator. It is trained with a frozen CLIP model, a learnable downsampler, and a lightweight content retention network. The training objective is to ensure that the image embeddings, after the up-down-sampling process, remain similar to those from CLIP. Meanwhile, the image generated by the content retention network, which takes the upsampled image embeddings as input, should closely resemble the input image to CLIP.

\begin{figure}[t]
    \centering
    \includegraphics[width=\linewidth]{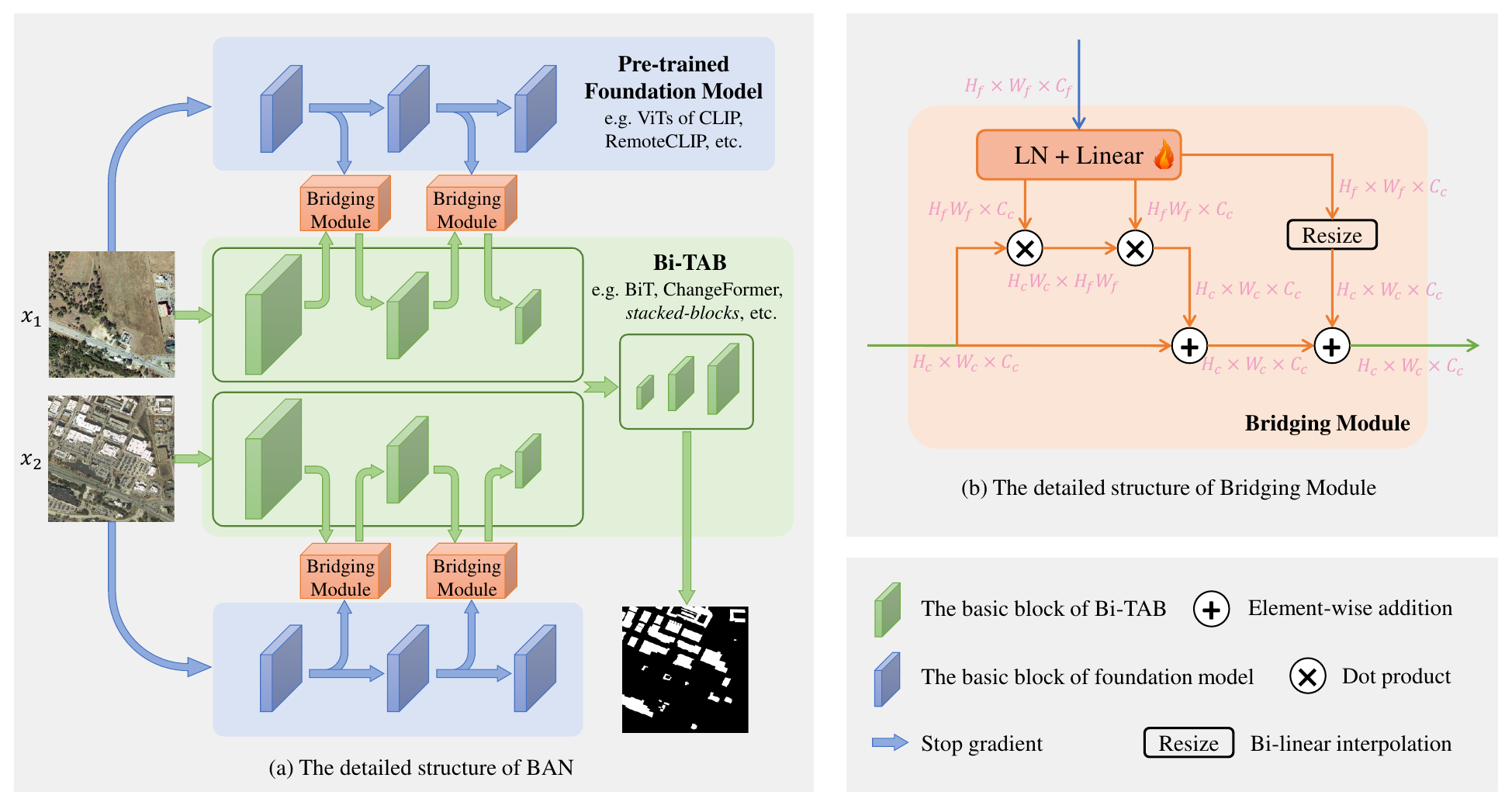}
    \caption{Illustration of BAN architecture~\cite{li2024new}.}
    \label{fig:ban}
\end{figure}

When it comes to change detection, similar adaptation strategies used in segmentation tasks, such as combining SAM with CLIP and designing decoders, are applied. For instance, in SCM~\cite{tan2024segment}, CLIP is applied after SAM to identify objects of interest in bi-temporal images, helping to filter out pseudo changes. In ChangeCLIP~\cite{dong2024changeclip}, CLIP is used to construct and encode multi-modal input data for change detection tasks, while a transformer-based decoder combines bi-temporal vision-language features with image features to produce change maps. To be specific, the authors pre-define 56 common land cover categories in remote sensing images and use the CLIP model to identify bi-temporal images from these categories. Each image is paired with textual descriptions of the foreground and background, formatted as ``remote sensing image foreground objects, [Predict Classes]" and ``remote sensing image background objects", respectively. This way allows bi-temporal texts to highlight the changing object, providing additional information for change detection.

Alternatively, only the vision encoders of foundation models are adopted, as most existing datasets provide only bi-temporal images for identifying changes. In BAN~\cite{li2024new}, the authors introduce bridging modules to inject general knowledge extracted from the vision encoders of foundation models into existing change detection models like BiT~\cite{chen2021remote} and ChangeFormer~\cite{bandara2022transformer}. Since the input image sizes for foundation models and change detection models may differ, and not all general knowledge contributes to predicting changes, the bridging modules are responsible for selecting, aligning, and injecting this knowledge. As shown in Fig.\ref{fig:ban}, these modules consist of three main components: layer normalization and a linear layer to mitigate the distribution inconsistency between the image features from the two models, cross-attention to obtain valuable information from general knowledge, and bi-linear interpolation to solve the misalignment problem. The bridging modules are placed between the two encoders, executing multi-level knowledge injection.

\paragraph*{\textbf{Broader Scope of Application}}
In addition to remote sensing tasks previously discussed, there are several emerging applications of contrastive-based vision-language foundation models, including:

(1) \textit{Visual Question Answering} attempts to provide answers to questions related to the content of images. In \cite{bazi2022bi}, CLIP is used to extract both visual and textual representations from images and questions. Two decoders, followed by two classifiers, are then developed to capture the intradependencies and interdependencies within and between these representations. The final answer is determined by averaging the predictions from both classifiers.

(2) \textit{Cloud Presence Detection} involves identifying satellite images that are affected by clouds. The authors in \cite{czerkawski2023detecting} explore several strategies to adapt CLIP for this task: One approach, similar to zero-shot scene classification~\cite{al2023vision}, involves selects prompts such as ``This is a satellite image with clouds" and ``This is a satellite image with clear sky", and then classifying satellite images via similarity measurement. Another strategy follows CoOp~\cite{zhou2022learning}, combining learnable context with the class token embeddings. Alternatively, the vision encoder can be employed by itself, with a linear classifier added on top. Unlike the first strategy, the other approaches necessitate training for prompts or classifiers.

(3) \textit{Text-based Image Generation} refers to the task of creating images from textual descriptions, which can help mitigate the issue of class imbalance commonly found in remote sensing data. Typically, TGN~\cite{tang2024text} utilizes CLIP to classify generated images, ensuring semantic class consistency between input images and generated ones.

\begin{figure}[t]
    \centering
    \includegraphics[width=\linewidth]{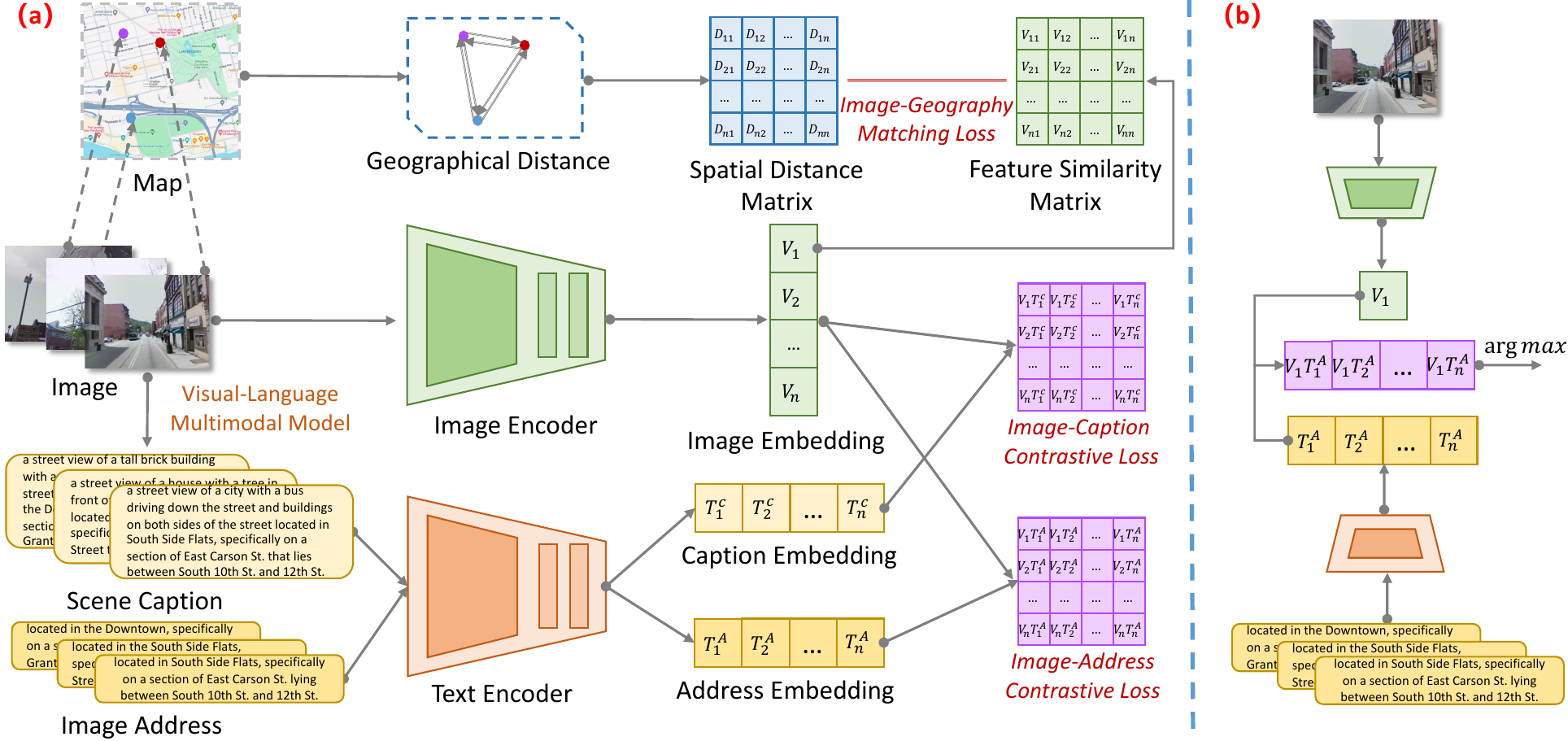}
    \caption{Illustration of AddressCLIP framework~\cite{xu2024addressclip}.}
    \label{fig:addressclip}
\end{figure}

(4) \textit{Image Address Localization} aims to predict the readable textual address where an image was taken~\cite{xu2024addressclip}. Different from image geo-localization in \cite{cepeda2023geoclip}, which predicts GPS coordinates (\emph{i.e.}, latitude and longitude) from images, this task outputs semantic address text such as ``Grant Street, Downtown". Utilizing existing foundation models like CSP~\cite{mai2023csp} and GeoCLP~\cite{cepeda2023geoclip}, a viable solution is predicting GPS coordinates with a foundation model and then converting them into readable addresses. However, this mapping from coordinates to addresses often presents ambiguities. A recent work, AddressCLIP~\cite{xu2024addressclip}, introduces a novel idea for performing image address localization in an end-to-end manner. In particular, AddressCLIP applies contrastive learning to align images with scene captions and address texts. Furthermore, image-geography matching is developed to bring features of geographically proximate images closer together while distancing features of images that are far apart geographically. Fig.\ref{fig:addressclip} shows the framework of AddressCLIP.

\section{Instruction-based Vision-Language Modeling}
Today, instruction-based vision-language modeling is advancing rapidly. Since 2023, over ten impressive vision-language models have emerged, as shown in Table~\ref{tab:instruction-VLM}. They are not only versatile, capable of performing a range of remote sensing image analysis tasks, but also able to interact with users in a conversational manner. This broadens the accessibility of intelligent models beyond experts in remote sensing, facilitating their widespread deployment and application. This section presents critical developments in terms of \textit{model architecture}, and \textit{training strategy}, \textit{model capability}.

\begin{table*}[ht]
\centering
\caption{Summary of instruction-based vision-language models in remote sensing.}
\label{tab:instruction-VLM}
\resizebox{\textwidth}{!}{
\rowcolors{2}{gray!6}{white}
\begin{tabular}{lcccclc}
\toprule
 \textbf{Model} & \textbf{Vision Encoder} & \textbf{Connector} & \textbf{Large Language Model} & \textbf{Training Dataset} & \textbf{Training Strategy}  & \textbf{Public} \\\midrule
 
RSGPT~\cite{hu2023rsgpt} & EVA CLIP-G~\cite{fang2023eva} & \makecell{\gape Q-Former~\cite{dai2023instructblip} \\ +Linear layer} & Vicuna-7B/13B~\cite{vicuna2023} & RSICap~\cite{hu2023rsgpt}& \makecell[{{p{6cm}}}]{ \raisebox{0.5ex}{\scalebox{0.7}{$\bullet$}} Load pre-trained weights of InstructBLIP~\cite{dai2023instructblip} \\ \raisebox{0.5ex}{\scalebox{0.7}{$\bullet$}} Fine-tune Q-Former and linear layer } &  \ding{55}\\

SkyEyeGPT~\cite{zhan2024skyeyegpt} & EVA CLIP-G~\cite{fang2023eva} & Linear layer & LLaMA 2-Chat~\cite{touvron2023llama2} & SkyEye-968k~\cite{zhan2024skyeyegpt} & \makecell[{{p{6cm}}}]{ \raisebox{0.5ex}{\scalebox{0.7}{$\bullet$}} Load pre-trained weights of \cite{fang2023eva,chen2023minigpt} \\ \raisebox{0.5ex}{\scalebox{0.7}{$\bullet$}} Employ LoRA~\cite{hu2021lora} for linear layer and LLM in two stages } & \ding{55} \\

EarthGPT~\cite{zhang2024earthgpt} & \makecell{DINOv2 ViT-L/14~\cite{oquab2023dinov2} \\ + CLIP ConvNeXt-L~\cite{radford2021learning}} & Linear layer & LLaMA 2-13B~\cite{touvron2023llama2} & \begin{tabular}[l]{@{}l@{}} LAION-400M~\cite{schuhmann2021laion} \\ COCO Caption~\cite{chen2015microsoft} \\ MMRS-1M~\cite{zhang2024earthgpt} \end{tabular} & \makecell[{{p{6cm}}}]{ \raisebox{0.5ex}{\scalebox{0.7}{$\bullet$}} Load pre-trained weights of \cite{oquab2023dinov2,radford2021learning,touvron2023llama2} \\ \raisebox{0.5ex}{\scalebox{0.7}{$\bullet$}} Train linear layer and LLM on LAION-400M and COCO Caption \\ \raisebox{0.5ex}{\scalebox{0.7}{$\bullet$}} Employ bias tuning~\cite{gao2023llama} for LLM on MMRS-1M } & \ding{55} \\

EarthMarker~\cite{zhang2024earthmarker} & \begin{tabular}[c]{@{}c@{}} DINOv2 ViT-L/14~\cite{oquab2023dinov2} \\ +CLIP ConvNeXt~\cite{radford2021learning} 
\end{tabular} & Projection layer & LLaMA 2-13B~\cite{touvron2023llama2} & 
\begin{tabular}[c]{@{}c@{}} COCO Caption~\cite{chen2015microsoft} \\ RSVP-3M~\cite{zhang2024earthmarker} \\ RefCOCO~\cite{kazemzadeh2014referitgame} \\ RefCOCO+~\cite{yu2016modeling} \end{tabular} & 
\makecell[{{p{6cm}}}]{\raisebox{0.5ex}{\scalebox{0.7}{$\bullet$}} Load pre-trained weights of \cite{oquab2023dinov2,radford2021learning,touvron2023llama2} \\ \raisebox{0.5ex}{\scalebox{0.7}{$\bullet$}} Train projection layer on COCO Caption and RSVP subset \\ \raisebox{0.5ex}{\scalebox{0.7}{$\bullet$}} Fine-tune attention layers of LLM on RefCOCO and RefCOCO+ \\ \raisebox{0.5ex}{\scalebox{0.7}{$\bullet$}} Employ LoRA~\cite{hu2021lora} for LLM on RSVP-3M } & \ding{55} \\

Popeye~\cite{zhang2024popeye} & \makecell{DINOv2 ViT-L/14~\cite{oquab2023dinov2} \\ +CLIP ViT-L/14~\cite{radford2021learning}} & Projection layer &  LLaMA-7B~\cite{touvron2023llama1} & \begin{tabular}[l]{@{}l@{}} COCO Caption~\cite{chen2015microsoft} \\ MMShip~\cite{zhang2024popeye} \end{tabular}  & \makecell[{{p{6cm}}}]{\raisebox{0.5ex}{\scalebox{0.7}{$\bullet$}} Load pre-trained weights of \cite{oquab2023dinov2,radford2021learning,touvron2023llama1} \\ \raisebox{0.5ex}{\scalebox{0.7}{$\bullet$}} Train projection layer and employ LoRA~\cite{hu2021lora} for LLM on COCO Caption \\ \raisebox{0.5ex}{\scalebox{0.7}{$\bullet$}} Fine-tuning projection layer and employ LoRA and instruction adapters for LLM on MMShip }  & \ding{55} \\

LHRS-Bot~\cite{muhtar2024lhrs} \href{https://github.com/NJU-LHRS/LHRS-Bot}{[link]} & CLIP ViT-L/14~\cite{radford2021learning} & Vision perceiver & 
LLaMA 2-7B~\cite{touvron2023llama2} & \begin{tabular}[c]{@{}c@{}} LHRS-Align~\cite{muhtar2024lhrs} \\ LHRS-Instruct~\cite{muhtar2024lhrs} \\ 
Multi-task Dataset~\cite{muhtar2024lhrs} \\ LLaVA-Instruct-158K~\cite{liu2024visual} \end{tabular} & \makecell[{{p{6cm}}}]{\raisebox{0.5ex}{\scalebox{0.7}{$\bullet$}} Load pre-trained weights of \cite{radford2021learning,touvron2023llama2} \\ \raisebox{0.5ex}{\scalebox{0.7}{$\bullet$}} Train vision perceiver on LHRS-Align \\ \raisebox{0.5ex}{\scalebox{0.7}{$\bullet$}} Fine-tune vision perceiver and Employ LoRA~\cite{hu2021lora} for LLM on LHRS-Instruct, multi-task dataset \\ \raisebox{0.5ex}{\scalebox{0.7}{$\bullet$}} Fine-tune LLM on LHRS-Instruct, multi-task dataset and LLaVA-Instruct-158K subset} & 
\ding{51} \\

VHM~\cite{pang2024vhm} \href{https://github.com/opendatalab/VHM}{[link]} & CLIP ViT-L/14~\cite{radford2021learning} & MLP & Vicuna-v1.5-7B~\cite{vicuna2023} & \begin{tabular}[l]{@{}l@{}} VersaD~\cite{pang2024vhm} \\ VersaD-Instruct~\cite{pang2024vhm} \\ HnstD~\cite{pang2024vhm} \\ VariousRS-Instruct~\cite{pang2024vhm} \end{tabular} & \makecell[{{p{6cm}}}]{ \raisebox{0.5ex}{\scalebox{0.7}{$\bullet$}} Load pre-trained weights of \cite{radford2021learning,vicuna2023} \\ \raisebox{0.5ex}{\scalebox{0.7}{$\bullet$}} Train vision encoder, MLP and LLM on VersaD \\ \raisebox{0.5ex}{\scalebox{0.7}{$\bullet$}} Fine-tune MLP and LLM on VersaD-Instruct, HnstD and VariousRS-Instruct } & \ding{51} \\

GeoChat~\cite{kuckreja2024geochat} \href{https://github.com/mbzuai-oryx/geochat}{[link]} & CLIP ViT-L/14~\cite{radford2021learning} & 
MLP & Vicuna-v1.5-7B~\cite{vicuna2023} & Multimodal Instruction Dataset~\cite{kuckreja2024geochat} & \makecell[{{p{6cm}}}]{\raisebox{0.5ex}{\scalebox{0.7}{$\bullet$}} Load pre-trained weights of \cite{tay2017learning,liu2024improved,vicuna2023} \\ 
\raisebox{0.5ex}{\scalebox{0.7}{$\bullet$}} Employ LoRA~\cite{hu2021lora} for LLM } & \ding{51} \\

RS-LLaVA~\cite{bazi2024rs} \href{https://github.com/BigData-KSU/RS-LLaVA}{[link]} & CLIP ViT-L@336~\cite{radford2021learning} & MLP & Vicuna-v1.5-7B/13B~\cite{vicuna2023} & RS-Instructions~\cite{bazi2024rs}&\makecell[{{p{6cm}}}]{\raisebox{0.5ex}{\scalebox{0.7}{$\bullet$}} Load pre-trained weights of \cite{radford2021learning,vicuna2023}\\ \raisebox{0.5ex}{\scalebox{0.7}{$\bullet$}} Train MLP on a general image-language dataset \\ \raisebox{0.5ex}{\scalebox{0.7}{$\bullet$}} Employ LoRA~\cite{hu2021lora} for LLM }  & \ding{51} \\

SkySenseGPT~\cite{luo2024skysensegpt} \href{https://github.com/Luo-Z13/SkySenseGPT?tab=readme-ov-file}{[link]} & CLIP ViT-L/14~\cite{radford2021learning} & MLP & Vicuna-v1.5~\cite{vicuna2023} & \begin{tabular}[c]{@{}c@{}} FIT-RS~\cite{luo2024skysensegpt} \\ Additional Instruction Dataset~\cite{luo2024skysensegpt} \end{tabular} & \makecell[{{p{6cm}}}]{ \raisebox{0.5ex}{\scalebox{0.7}{$\bullet$}} Load pre-trained weights of \cite{radford2021learning,liu2024improved,vicuna2023} \\ \raisebox{0.5ex}{\scalebox{0.7}{$\bullet$}} Fine-tune MLP and employ LoRA~\cite{hu2021lora} for LLM } & \ding{51} \\

IFShip~\cite{guo2024ifship} & CLIP ViT-L/14~\cite{radford2021learning} & MLP & Vicuna-13B~\cite{vicuna2023} & TITANIC-FGS~\cite{guo2024ifship} & \makecell[{{p{6cm}}}]{\raisebox{0.5ex}{\scalebox{0.7}{$\bullet$}} Load pre-trained weights of \cite{radford2021learning,vicuna2023} \\ \raisebox{0.5ex}{\scalebox{0.7}{$\bullet$}} Employ LoRA~\cite{hu2021lora} for MLP and LLM } & \ding{55} \\

TEOChat~\cite{irvin2024teochat}\href{https://github.com/ermongroup/TEOChat?tab=readme-ov-file}{[link]} & CLIP ViT-L/14~\cite{radford2021learning} & 
MLP & LLaMA 2~\cite{touvron2023llama2} & TEOChatlas~\cite{irvin2024teochat} & \makecell[{{p{6cm}}}]{\raisebox{0.5ex}{\scalebox{0.7}{$\bullet$}}  Load pre-trained weights of \cite{radford2021learning,lin2023video} \\ \raisebox{0.5ex}{\scalebox{0.7}{$\bullet$}} Employ LoRA~\cite{hu2021lora} for LLM } & \ding{51} \\
\bottomrule
\end{tabular}
}

\vspace{3pt}
\begin{minipage}{\textwidth}
    \footnotesize \textcolor{blue}{[link]} directs to model websites. Detailed information about training datasets can be found in Tables~\ref{tab:pretraining-datasets} and \ref{tab:fine-tuning-dataset}. \textit{Public} refers to the availability of both code and model weights.
\end{minipage}
\end{table*}

\subsection{Model Architecture}
As CLIP has been trained to align image and text representations, most works~\cite{kuckreja2024geochat,muhtar2024lhrs,bazi2024rs,luo2024skysensegpt,pang2024vhm} directly employ its vision encoder. The commonly used large language models include the LLaMA family~\cite{touvron2023llama1,touvron2023llama2} and its derivative Vicuna family~\cite{vicuna2023}. Developed by Meta, the first-generation LLaMA~\cite{touvron2023llama1} adopts transformer architectures ranging from 7B to 65B parameters, pre-trained on approximately 1.4 trillion tokens of publicly available text corpora. Despite its smaller scale, LLaMA outperforms proprietary models like GPT-3~\cite{brown2020language}. Its successor, LLaMA 2~\cite{touvron2023llama2}, introduces significant improvements, including expanded pre-training data (2 trillion tokens), architectural upgrades such as grouped-query attention~\cite{ainslie2023gqa}, and a scaled-up 70B-parameters variant. Through supervised fine-tuning and reinforcement learning with human feedback, LLaMA 2 was further optimized for dialogue, yielding the LLaMA 2-Chat series~\cite{touvron2023llama2}. Parallel to this development, the research community developed Vicuna~\cite{vicuna2023} by fine-tuning LLaMA models on 70K conversations from ShareGPT. These models achieve more than 90\% of ChatGPT's quality.

Regarding model architecture, existing works are devoted to improving visual encoders to enhance visual perception, and designing connectors to promote the alignment between the two modalities. Specifically,

\paragraph*{\textbf{Vision Encoder}} Through a mask image modeling pretext task, EVA~\cite{fang2023eva} incorporates geometry and structure information into CLIP's visual representations, leading to improved performances across a wide range of visual perception tasks. Consequently, RSGPT~\cite{hu2023rsgpt} and SkyEyeGPT~\cite{zhan2024skyeyegpt} adopt EVA as their vision encoder. An alternative to complementing CLIP is utilizing the strengths of diverse vision encoders. CLIP learns visual representations through language supervision, which limits the information retained about the image. Captions only approximate the main content of images, often failing to present complex pixel-level details. Targeting this limitation, some works~\cite{zhang2024earthgpt,zhang2024earthmarker,zhang2024popeye} propose combining vision encoders from CLIP and DINOv2~\cite{oquab2023dinov2}. DINOv2 learns visual representation from images alone via self-supervised learning, enabling it to capture both image-level and pixel-level information. Moreover, given the varied object sizes in remote sensing images, these works refine visual representations by incorporating multi-scale information. In \cite{zhang2024earthgpt}, a vision encoder with the convolutional neural network architecture is used to extract multi-scale visual features. In \cite{zhang2024earthmarker,zhang2024popeye}, the input image is downsampled to different resolutions and then respectively fed into two vision encoders. The encoded visual features are transformed to the same dimension and concatenated channel-wise.

\paragraph*{\textbf{Vision-Language Connector}} The linear layer and MLP are widely used as vision-language connectors, serving as key components in most models~\cite{zhan2024skyeyegpt,zhang2024earthgpt,pang2024vhm,kuckreja2024geochat,bazi2024rs,luo2024skysensegpt,guo2024ifship,irvin2024teochat}. Differently, RSGPT~\cite{hu2023rsgpt} and LHRS-Bot~\cite{muhtar2024lhrs} explore alternative connector architectures. Following InstructBLIP~\cite{dai2023instructblip}, RSGPT includes an instruction-aware query transformer (Q-Former) as an intermediate module between the vision encoder and LLM. The Q-Former is designed to extract task-relevant visual representations by interacting additional query embeddings with instruction and image embeddings. It achieves this through the attention mechanism: first, self-attention is applied to implement interaction between instruction and query embeddings, and then cross-attention is employed between query and image embeddings. The resulting output from the Q-Former, after passing through a linear layer, is then input into the LLM along with the instruction embeddings to generate responses. LHRS-Bot proposes incorporating multi-level image embeddings to sufficiently capture the semantic content of images. To mitigate the computational burden and the risk of overwhelming language information with excessive visual information, it introduces a set of learnable queries for each level of visual representation. These queries are used to summarize the semantic information of each level through stacked cross-attention and MLP layers. As a result, a dedicated visual perceiver is developed, with experimental results demonstrating that, compared to single-level visual representations and a two-layer MLP, multi-level representations paired with the visual perceiver improve the model's performance in scene classification, visual question answering, and visual grounding tasks.

\subsection{Training Strategy}
Training instruction-based vision-language models typically involves two stages: pre-training for modality alignment and supervised fine-tuning (SFT) for following task-specific instructions. 

\paragraph*{\textbf{Only SFT}} Due to the lack of large-scale image-text datasets specifically designed for the remote sensing domain, most works target the SFT stage using carefully crafted instruction-following datasets~\cite{hu2023rsgpt,zhan2024skyeyegpt,kuckreja2024geochat,luo2024skysensegpt,guo2024ifship,irvin2024teochat}. To preserve the general knowledge embedded in pre-trained vision encoders, the vision encoder is typically kept frozen during training, with the connector or the LLM undergoing fine-tuning. For instance, RSGPT~\cite{hu2023rsgpt} fine-tunes the connector, while GeoChat~\cite{kuckreja2024geochat} and TEOChat~\cite{irvin2024teochat} fine-tune the LLM. SkyEyeGPT~\cite{zhan2024skyeyegpt}, SkySenseGPT~\cite{luo2024skysensegpt} and IFShip~\cite{guo2024ifship} fine-tune both the connector and LLM. To avoid the expense of full-parameter tuning, LoRA (Low-Rank Adaptation)~\cite{hu2021lora} is often adopted, which introduces low-rank learnable matrices into the layers of the connector~\cite{zhan2024skyeyegpt,guo2024ifship} or LLM~\cite{zhan2024skyeyegpt,kuckreja2024geochat,luo2024skysensegpt,guo2024ifship,irvin2024teochat}. 

\paragraph*{\textbf{Pre-training followed by SFT}} A couple of recent works have investigated how to implement the pre-training stage to boost model performance. Based on the choice of training data, they can be categorized into two groups: those that combine available image-text pairs from multiple domains for pre-training and those that direct attention toward creating large-scale RS image-text datasets. For combining available data, EarthGPT~\cite{zhang2024earthgpt}, Popeye~\cite{zhang2024popeye} and RS-LLaVA~\cite{bazi2024rs} utilize natural image-text datasets, while EarthMarker~\cite{zhang2024earthmarker} integrates data from both computer vision and remote sensing domains. COCO Caption~\cite{chen2015microsoft} is a commonly used pre-training dataset. Unlike RS-LLaVA, which limits pre-training to the connector, the other three models perform pre-training for the connector and the LLM.

\begin{table*}[ht]
\centering
\caption{Capability comparisons of instruction-based vision-language models in remote sensing.}
\label{tab:capability-VLM}
\rowcolors{3}{gray!6}{white}
\begin{tabular}{lccccccccl}
\toprule
\multirow{2}{*}{\textbf{Model}} & \textbf{SIT}  & \textbf{Cap.} & \textbf{CLS} & \multirow{2}{*}{\textbf{VQA}} & \multirow{2}{*}{\textbf{VG}} & \multirow{2}{*}{\textbf{OD}} & \multirow{2}{*}{\textbf{ORR}} &  \multirow{2}{*}{\textbf{MTC}} & \multirow{2}{*}{\textbf{Others}}  \\
 & Opt/SAR/IR & Img/Reg/Pt & Img/Reg/Pt & & & & & & \\ \midrule
RSGPT~\cite{hu2023rsgpt} & \ding{51} / \ding{55} / \ding{55} & \ding{51} / \ding{55} / \ding{55} & \ding{55} / \ding{55} / \ding{55} & \ding{51} & \ding{55} & \ding{55} & \ding{55} & \ding{55} & \ding{55} \\
SkyEyeGPT~\cite{zhan2024skyeyegpt}  & \ding{51} / \ding{55} / \ding{55} & \ding{51} / \ding{55} / \ding{55} & \ding{51} / \ding{55} / \ding{55} & \ding{51} & \ding{51} & \ding{51} & \ding{55} & \ding{51} & \begin{tabular}[l]{@{}l@{}} Video Captioning \\ Referring Expression Generation \end{tabular}  \\
EarthGPT~\cite{zhang2024earthgpt}  & \ding{51} / \ding{51} / \ding{51}  & \ding{51} / \ding{51} / \ding{55} & \ding{51} / \ding{55} / \ding{55} & \ding{51} & \ding{51} & \ding{51} & \ding{55} & \ding{51} & \ding{55} \\
EarthMarker~\cite{zhang2024earthmarker}  & \ding{51} / \ding{55} / \ding{55} & \ding{51} / \ding{51} / \ding{51} & \ding{51} / \ding{51} / \ding{51} & \ding{55} & \ding{55} & \ding{51} & \ding{51} & \ding{51}  & \ding{55}  \\
Popeye~\cite{zhang2024popeye}$\dag$ & \ding{51} / \ding{51} / \ding{55} & \ding{51} / \ding{55} / \ding{55} & \ding{55} / \ding{55} / \ding{55} & \ding{55} & \ding{55} & \ding{51} & \ding{55} & \ding{51} & \ding{55} \\
LHRS-Bot~\cite{muhtar2024lhrs} & \ding{51} / \ding{55} / \ding{55} & \ding{51} / \ding{55} / \ding{55} & \ding{51} / \ding{55} / \ding{55} & \ding{51} & \ding{51} & \ding{55} & \ding{51} & \ding{51} &  \begin{tabular}[l]{@{}l@{}} Object Counting \\ Object Attribute Recognition \\  Image Property Recognition \end{tabular}  \\
VHM~\cite{pang2024vhm} & \ding{51} / \ding{55} / \ding{55} & \ding{51} / \ding{55} / \ding{55} & \ding{51} / \ding{55} / \ding{55} & \ding{51} & 
\ding{51} & \ding{55} & \ding{51} & \ding{51} & \begin{tabular}[l]{@{}l@{}} Object Counting \\ Geometric Measurement \\ Building Vectorizing \\ Image Property Recognition \\ Multi-Label Classification \\ Honest Question Answering \end{tabular} \\
GeoChat~\cite{kuckreja2024geochat}  & \ding{51} / \ding{55} / \ding{55} & \ding{51} / \ding{51}/ \ding{55} & \ding{51} / \ding{55} / \ding{55} & \ding{51} & \ding{51} & \ding{55} & \ding{55} & \ding{51} &  \ding{55} \\
RS-LLaVA~\cite{bazi2024rs} & \ding{51} / \ding{55} / \ding{55} & \ding{51} / \ding{55} / \ding{55} & \ding{55} / \ding{55} / \ding{55} & \ding{51} & \ding{55} & \ding{55} & \ding{55} & \ding{55} & \ding{55}  \\
SkySenseGPT~\cite{luo2024skysensegpt} & \ding{51} / \ding{55} / \ding{55}  & \ding{51} / \ding{51} / \ding{55} & \ding{51} / \ding{55} / \ding{55} & \ding{51} & \ding{51} & \ding{51} & \ding{51} & \ding{51} & \begin{tabular}[l]{@{}l@{}} Multi-Label Classification \\ Image/Region Scene Graph Generation \end{tabular} \\
IFShip~\cite{guo2024ifship}$\dag$ & \ding{51} / \ding{55} / \ding{55} & \ding{51} / \ding{55} / \ding{55} & \ding{51} / \ding{55} / \ding{55} & \ding{51} & \ding{55} & \ding{55} & \ding{55} & \ding{51} & \ding{55}  \\
TEOChat~\cite{irvin2024teochat} & \ding{51} / \ding{55} / \ding{55} & \ding{51} / \ding{51} / \ding{55} & \ding{51} / \ding{55} / \ding{55} & \ding{51} & \ding{51} & \ding{55} & \ding{55} & \ding{51} & \begin{tabular}[l]{@{}l@{}} Change Detection \\ Temporal Scene Classification \\ Temporal Referring Expression \\ Spatial Change Referring Expression \\ Image/Region Change Question Answering \end{tabular} \\
\bottomrule
\end{tabular}

\vspace{3pt}
\begin{minipage}{\textwidth}
    \footnotesize $\dag$ indicates the model's capability tailored for remote sensing images of ships. \textit{SIT}, \textit{Cap.}, \textit{CLS}, \textit{VQA}, \textit{VG}, \textit{OD}, \textit{ORR}, and \textit{MTC} are short for \textit{Supported Image Type}, \textit{Captioning}, \textit{Classification}, \textit{Visual Question Answering}, \textit{Visual Grounding}, \textit{Object Detection}, \textit{Object Relationship Reasoning}, and \textit{Multi-Turn Conversation}, respectively. We present the \textit{Opt/SAR/IR} column to indicate whether the model supports input of optical (Opt), synthetic aperture radar (SAR), or infrared (IR) images, and the \textit{Img/Reg/Pt} column to denote the granularity of the analyzed information, which includes image level (Img), region level (Reg), or point level (Pt).
\end{minipage}
\end{table*}

To address the domain gap between remote sensing images and natural images, researchers have developed large-scale RS image-text datasets, such as LHRS-Align~\cite{muhtar2024lhrs} and VersaD~\cite{pang2024vhm}, both of which contain over one million training samples. Leveraging the LHRS-Align dataset, Muhtar et al.~\cite{muhtar2024lhrs} design a three-stage curriculum learning strategy for LHRS-Bot. First, the vision perceiver is pre-trained on LHRS-Align. Next, the vision perceiver and LLM are fine-tuned on LHRS-Instruct subset and multi-task dataset, equipping the model to handle multimodal tasks. Finally, the LLM undergoes additional fine-tuning on all instruction data from LHRS-Instruct, the multi-task dataset, and LLaVA-Instruct-158K subset to fully unlock LHRS-Bot's capabilities. Note that these models still keep the vision encoder frozen. In contrast, Pang et al.~\cite{pang2024vhm} unfreeze the vision encoder, MLP and LLM for pre-training on VersaD. They subsequently fine-tune the MLP and LLM on customized datasets, including VersaD-Instruct, HnstD and VariousRS-Instruct. Their experimental results show that the model pre-trained with RS data significantly outperforms the one only fine-tuned with RS data across multiple tasks, confirming the importance of incorporating extensive RS visual knowledge by pre-training.

\subsection{Model Capability} 
Table~\ref{tab:capability-VLM} lists the capabilities of existing instruction-based vision-language models in remote sensing. Key observations include: 1) Most models are developed for general-purpose remote sensing data analysis~\cite{zhan2024skyeyegpt,zhang2024earthgpt,zhang2024earthmarker,muhtar2024lhrs,pang2024vhm,kuckreja2024geochat,luo2024skysensegpt,irvin2024teochat}, with only Popeye~\cite{zhang2024popeye} and IFShip~\cite{guo2024ifship} specifically tailored for remote sensing images of ships. These models primarily process optical images but have expanded to include SAR (Synthetic Aperture Radar)~\cite{zhang2024earthgpt,zhang2024popeye} and infrared~\cite{zhang2024earthgpt} images. 2) Most models support conversational interaction with users and can perform a variety of tasks~\cite{zhan2024skyeyegpt,zhang2024earthgpt,zhang2024earthmarker,muhtar2024lhrs,pang2024vhm,kuckreja2024geochat,luo2024skysensegpt,irvin2024teochat}, ranging from single image analysis to temporal series analysis~\cite{irvin2024teochat}. 3) The granularity of analyzed information has progressed from the image level to region level~\cite{luo2024skysensegpt,zhang2024earthmarker}, and even to the point level~\cite{zhang2024earthmarker}. In this section, we examine the models' performance on common capabilities and offer an in-depth exploration of unique capabilities.

\begin{figure}[t]
\centering
\subfloat[SkyEyeGPT~\cite{zhan2024skyeyegpt}]{\includegraphics[width=\linewidth]{Figures/SkyEyeGPT.pdf}} \\
\subfloat[LHRS-Bot~\cite{muhtar2024lhrs}]{\includegraphics[width=\linewidth]{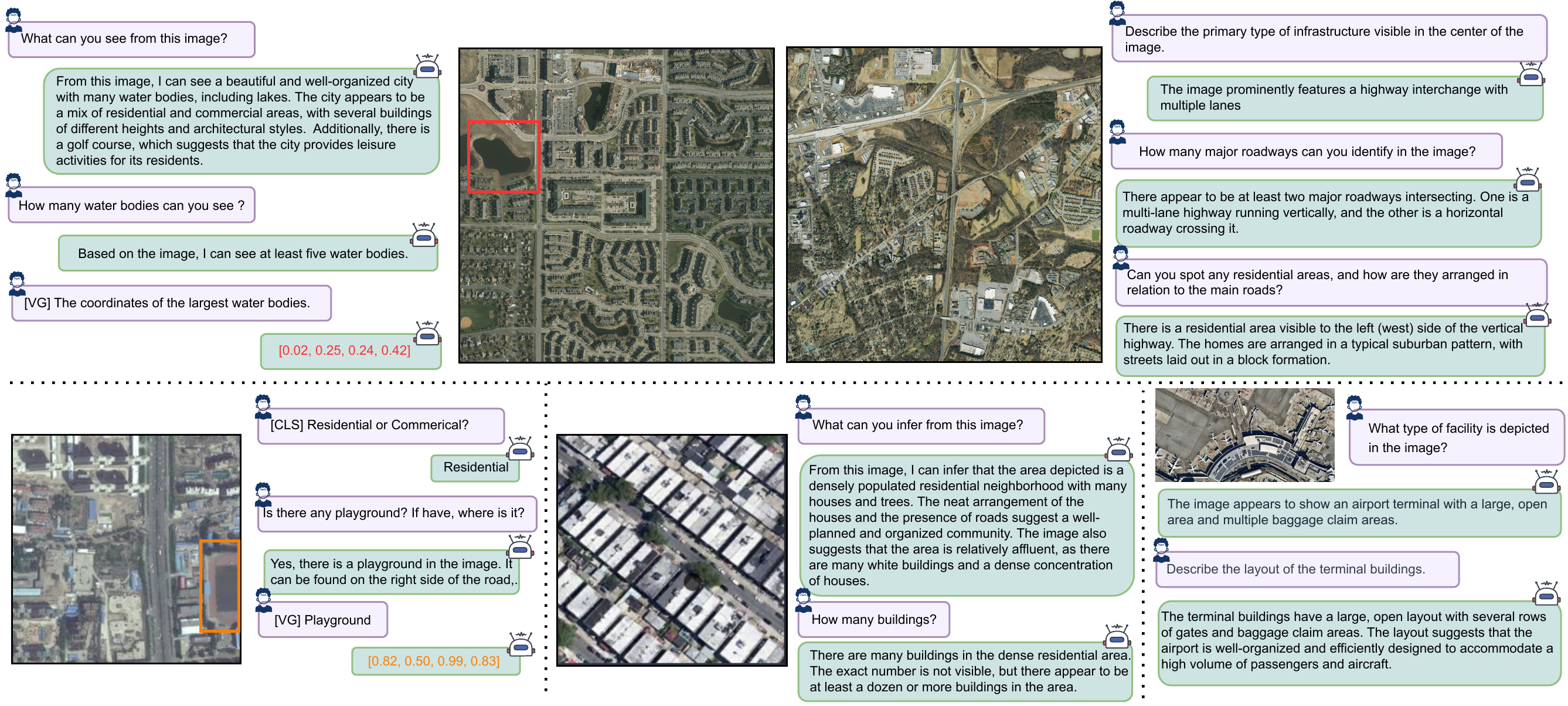}}
\caption{Examples of conversations between vision-language models and users.}
\label{fig:multi-turn-conversation}
\end{figure}

\paragraph*{\textbf{Common Capabilities}}  As shown in Table~\ref{tab:capability-VLM}, most models are capable of performing remote sensing tasks such as image captioning, scene classification, visual question answering, and visual grounding. Accordingly, we present model performance on these tasks in Tables~\ref{tab:performance-VLM-Cap}-\ref{tab:performance-VLM-VG}. To ensure a fair comparison, we report only the models' performance on public datasets. Although multi-turn Conversation is also a common capability among existing models~\cite{zhan2024skyeyegpt,zhang2024earthgpt,zhang2024earthmarker,zhang2024popeye,muhtar2024lhrs,pang2024vhm,kuckreja2024geochat,luo2024skysensegpt,irvin2024teochat}, it is challenging to evaluate quantitatively. Fig.~\ref{fig:multi-turn-conversation} provides examples of conversations between various models and users.

(1) \textit{Image Captioning:} Based on the input remote sensing image and the language instruction, vision-language models generate a description of image content. Examples of instructions include ``Describe this image in detail."~\cite{hu2023rsgpt,zhan2024skyeyegpt} or ``Please provide a one-sentence caption for the provided remote sensing image in detail."~\cite{zhang2024earthgpt}. From Table~\ref{tab:performance-VLM-Cap}, we can observe that SkyEyeGPT~\cite{zhan2024skyeyegpt} outperforms RSGPT~\cite{hu2023rsgpt} on the RSICD~\cite{lu2017exploring}, UCM-captions~\cite{qu2016deep} and Sydney-captions~\cite{qu2016deep} datasets in terms of BLEU~\cite{papineni2002bleu}, METEOR~\cite{banerjee2005meteor}, and ROUGE$\_$L~\cite{lin2004rouge}, though it falls short of RSGPT in CIDEr~\cite{vedantam2015cider} scores. On the UCM-captions dataset, RS-LLaVA achieves BLEU scores comparable to SkyEyeGPT while surpassing it in all other metrics. On the NWPU-Captions~\cite{cheng2022nwpu} dataset, EarthGPT~\cite{zhang2024earthgpt} achieves state-of-the-art results, with a remarkable improvement in CIDEr (nearly 30\%) over EarthMarker~\cite{zhang2024earthmarker}.

(2) \textit{Scene Classification:} Table~\ref{tab:performance-VLM-CLS-VQA} presents evaluations on 10 remote sensing scene classification datasets under both fully supervised and zero-shot settings. The absorption of extensive remote sensing visual knowledge improves model accuracy and generalization in scene classification tasks. Specifically, in the supervised setting, VHM~\cite{pang2024vhm} and EarthGPT~\cite{zhang2024earthgpt} boost classification performance to over 93\% on the NWPU-RESISC45~\cite{cheng2017remote} dataset, which includes 45 scene categories with image spatial resolutions ranging from 0.2m to more than 30m. For zero-shot scene classification, LHRS-Bot~\cite{muhtar2024lhrs}, VHM, and SkySenseGPT~\cite{luo2024skysensegpt} demonstrate impressive generalization, achieving accuracy above 91\% on the AID~\cite{xia2017aid} dataset and over 93\% on the WHU-RS19~\cite{dai2010satellite} dataset. SkySenseGPT notably achieves the highest accuracy, reaching 92.25\% and 97.02\% on these datasets, respectively. Despite these remarkable results, low-resolution and fine-grained scene classification remain significant challenges. The EuroSAT~\cite{helber2019eurosat} dataset, used for land use and land cover classification, has an image spatial resolution of 10m, while fMoW~\cite{christie2018functional} contains over 1 million images across 63 scene categories. On these datasets, LHRS-Bot~\cite{muhtar2024lhrs} achieves an accuracy below 57\%.

(3) \textit{Visual Question Answering:} RSVQA-HR~\cite{lobry2020rsvqa} and RSVQA-LR datasets are widely adopted to assess model performance in visual question answering tasks, as shown in Table~\ref{tab:performance-VLM-CLS-VQA}. Under a supervised setting, RSGPT~\cite{hu2023rsgpt} establishes strong performance baselines on both datasets. Subsequent models, including EarthGPT~\cite{zhang2024earthgpt}, GeoChat~\cite{kuckreja2024geochat}, VHM~\cite{pang2024vhm}, and SkySenseGPT~\cite{luo2024skysensegpt}, have progressively approached these benchmarks in both supervised and zero-shot settings. The latest model, SkySenseGPT, narrows the zero-shot performance gap to approximately 13\%, coming closest to the baseline on test set 2 of the RSVQA-HR dataset and surpassing RSGPT on the RSVQA-LR dataset in the supervised setting.

(4) \textit{Visual Grounding:} aims to locate specific objects within an image based on a natural language expression. Vision-language models accomplish this task by providing the coordinates of the target object. Table~\ref{tab:performance-VLM-VG} presents the performance of several models on RSVG~\cite{sun2022visual} and DIOR-RSVG~\cite{zhan2023rsvg} datasets, using an intersection over union (IoU) threshold of 0.5 adopted as the evaluation metric. LHRS-Bot~\cite{muhtar2024lhrs} achieves the highest score on the RSVG test set, while SkyEyeGPT~\cite{zhan2024skyeyegpt} leads on DIOR-RSVG test set. Due to differences in metric calculations, VHM~\cite{pang2024vhm} achieves a score of 57.17\% on DIOR-RSVG.

\begin{table}[t]
\centering
\caption{Performance of instruction-based vision-language models on image captioning.}
\label{tab:performance-VLM-Cap}
\resizebox{\linewidth}{!}{
\rowcolors{3}{gray!6}{white}
\begin{tabular}{lccccccc}
\toprule
\textbf{Model} & BLEU-1 & BLEU-2 & BLEU-3 & BLEU-4 & METEOR & ROUGE$\_$L & CIDEr  \\ \midrule
\multicolumn{8}{c}{\cellcolor{white} \textbf{RSICD}~\cite{lu2017exploring}} \\ \midrule
RSGPT~\cite{hu2023rsgpt}           & 70.32 & 54.23 & 44.04 & 36.83 & 30.10 & 53.34 & \textbf{102.94} \\
SkyEyeGPT~\cite{zhan2024skyeyegpt} & \textbf{86.71} & \textbf{76.66} & \textbf{67.31} & \textbf{59.99} & \textbf{35.35} & \textbf{62.63} & 83.65 \\
\midrule
\multicolumn{8}{c}{\cellcolor{white} \textbf{UCM-captions}~\cite{qu2016deep}} \\ \midrule
RSGPT~\cite{hu2023rsgpt}           & 86.12 & 79.14 & 72.31 & 65.74 & 42.21 & 78.34 & 333.23 \\ 
RS-LLaVA~\cite{bazi2024rs}         & 90.00 & 84.88 & 80.30 & 76.03 & \textbf{49.21} & \textbf{85.78} & \textbf{355.61} \\
SkyEyeGPT~\cite{zhan2024skyeyegpt} & \textbf{90.71} & \textbf{85.69} & \textbf{81.56} & \textbf{78.41} & 46.24 & 79.49 & 236.75 \\
\midrule
\multicolumn{8}{c}{ \cellcolor{white} \textbf{Sydney-captions}~\cite{qu2016deep}} \\ \midrule
RSGPT~\cite{hu2023rsgpt}           & 82.26 & 75.28 & 68.57 & 62.23 & 41.37 & 74.77 & \textbf{273.08} \\
SkyEyeGPT~\cite{zhan2024skyeyegpt} & \textbf{91.85} & \textbf{85.64} & \textbf{80.88} & \textbf{77.40} & \textbf{46.62} & \textbf{77.74} & 181.06 \\
\midrule
\multicolumn{8}{c}{\cellcolor{white} \textbf{NWPU-Captions}~\cite{cheng2022nwpu}} \\ \midrule
EarthMarker~\cite{zhang2024earthmarker} & 84.40 & 73.10 & 62.90 & 54.30 & 37.50 & 70.00 & 162.90 \\ 
EarthGPT~\cite{zhang2024earthgpt}       & \textbf{87.10} & \textbf{78.70} & \textbf{71.60} & \textbf{65.50} & \textbf{44.50} & \textbf{78.20} & \textbf{192.60} \\
\bottomrule
\end{tabular}
}

\vspace{3pt}
\begin{minipage}{\linewidth}
    \footnotesize \textit{BLEU}, \textit{METEOR}, \textit{ROUGE$\_$L}, and \textit{CIDEr} are short for \textit{BiLingual Evaluation Understudy}~\cite{papineni2002bleu}, \textit{Metric for Evaluation of Translation with Explicit ORdering}~\cite{banerjee2005meteor}, \textit{Recall-Oriented Understudy for Gisting Evaluation}~\cite{lin2004rouge}, \textit{Consensus-based Image Description Evaluation}~\cite{vedantam2015cider}.
\end{minipage}
\end{table}

\begin{table*}[t]
\caption{Performance of instruction-based vision-language models on scene classification and visual question answering.}
\label{tab:performance-VLM-CLS-VQA}
\centering
\resizebox{\textwidth}{!}{
\rowcolors{4}{gray!6}{white}
\begin{tabular}{lcccccccccc}
\toprule
\multicolumn{11}{c}{\textbf{Scene Classification: Top-1 Accuracy}} \\ \cmidrule{1-11}
\textbf{Model} & RESISC~\cite{cheng2017remote} & CLRS~\cite{li2020clrs} & NaSC-TG2~\cite{zhou2021nasc} & UCM~\cite{yang2010bag} & AID~\cite{xia2017aid} & WHU-RS19~\cite{dai2010satellite} & SIRI-WHU~\cite{zhao2015dirichlet} & EuroSAT~\cite{helber2019eurosat} & METER-ML~\cite{zhu2022meter} & fMoW~\cite{christie2018functional} \\ \midrule
GeoChat~\cite{kuckreja2024geochat}      & - & - & - & 84.43$\dag$ & 72.03$\dag$ & - & - & - & - & -  \\
EarthMarker~\cite{zhang2024earthmarker} & - & - & - & \textbf{86.52}$\dag$ & 77.97$\dag$ & - & - & - & - & - \\
EarthGPT~\cite{zhang2024earthgpt}       & 93.84 & \textbf{77.37}$\dag$ & \textbf{74.72}$\dag$ & - & - & - & - & - & - & - \\
LHRS-Bot~\cite{muhtar2024lhrs}          & 83.94 & - & - & - & 91.26$\dag$ & 93.17$\dag$ & 62.66$\dag$ & \textbf{51.40}$\dag$ & 69.81 & \textbf{56.56} \\
VHM~\cite{pang2024vhm}                  & \textbf{94.54} & - & - & - & 91.70$\dag$ & 95.80$\dag$ & 70.88$\dag$ & - & \textbf{72.74} & - \\
SkySenseGPT~\cite{luo2024skysensegpt}   & - & - & - & - & \textbf{92.25}$\dag$ & \textbf{97.02}$\dag$ & \textbf{74.75}$\dag$ & - & - & - \\ \midrule

\multicolumn{11}{c}{\textbf{Visual Question Answering}} \\ \cmidrule{1-11}
\multirow{2}{*}{\textbf{Model}} & \multicolumn{3}{c}{RSVQA-HR~\cite{lobry2020rsvqa} Test Set 1} & \multicolumn{3}{c}{RSVQA-HR~\cite{lobry2020rsvqa} Test Set 2} & \multicolumn{4}{c}{RSVQA-LR~\cite{lobry2020rsvqa}} \\ \cmidrule{2-11}
& Presence & Comparison & Avg. Acc. & Presence & Comparison & Avg. Acc. & Presence & Comparison & Rural/Urban & Avg. Acc. \\ \midrule
SkyEyeGPT~\cite{zhan2024skyeyegpt}    & 84.95 & 85.63 & 85.29 & 83.50 & 80.28 & 81.89 & 88.93 & 88.63 & 75.00 & 84.19 \\
RSGPT~\cite{hu2023rsgpt}              & \textbf{91.86} & \textbf{92.15} & \textbf{92.00} & \textbf{89.87} & \textbf{89.68} &\textbf{ 89.78} & 91.17 & 91.70 & 94.00 & 92.29 \\
LHRS-Bot~\cite{muhtar2024lhrs}        & - & - & - & - & - & - & 88.51 & 90.00 & 89.07 & 89.19 \\
RS-LLaVA~\cite{bazi2024rs}            & - & - & - & - & - & - & \textbf{92.27} & 91.37 & \textbf{95.00} & 88.10 \\
GeoChat~\cite{kuckreja2024geochat}    & - & - & - & 58.45$\dag$ & 83.19$\dag$ & 72.30$\dag$ & 91.09 & 90.33 & 94.00 & 90.70 \\
EarthGPT~\cite{zhang2024earthgpt}     & - & - & - & 62.77$\dag$ & 79.53$\dag$ & 72.06$\dag$ & - & - & - & - \\
VHM~\cite{pang2024vhm}                & - & - & - & 64.00$\dag$ & 83.50$\dag$ & 73.75$\dag$ & 90.11 & 89.89 & 88.00 & 89.33 \\
SkySenseGPT~\cite{luo2024skysensegpt} & - & - & - & 69.14$\dag$ & 84.14$\dag$ & 76.64$\dag$ & 91.07 & \textbf{92.00} & \textbf{95.00} & \textbf{92.69} \\

\bottomrule
\end{tabular}
}

\vspace{3pt}
\begin{minipage}{\textwidth}
    \footnotesize \textit{RESISC}, \textit{Avg. Acc.} are short for NWPU-RESISC45~\cite{cheng2017remote} and \textit{Average Accuracy}, respectively. $\dag$ indicates zero-shot performance.
\end{minipage}
\end{table*}

\begin{table}[t]
\centering
\caption{Performance of instruction-based vision-language models on visual grounding.}
\label{tab:performance-VLM-VG}
\begin{tabular}{lccc}
\toprule
\multirow{2}{*}{\textbf{Model}} & \multicolumn{2}{c}{\textbf{RSVG}~\cite{sun2022visual}} & \textbf{DIOR-RSVG}~\cite{zhan2023rsvg} \\ \cmidrule{2-4}
 & Val & Test & Test \\ \midrule
SkyEyeGPT~\cite{zhan2024skyeyegpt} & 69.19 & 70.50 & 88.59 \\
EarthGPT~\cite{zhang2024earthgpt}  & - & - & 76.65 \\
LHRS-Bot~\cite{muhtar2024lhrs}     & - & 73.45 & 88.10 \\
VHM~\cite{pang2024vhm}         & - & - & 56.17 \\
\bottomrule
\end{tabular}

\vspace{3pt}
\begin{minipage}{\linewidth}
    \footnotesize Evaluation metric adopts Acc@0.5, which means that the intersection over union (IoU) of the predicted box is at least 0.5 with the ground truth bounding box.
\end{minipage}
\end{table}

\paragraph*{\textbf{Unique Capabilities}} Current research seeks to develop versatile vision-language models capable of handling various remote sensing image analysis tasks in a conversational manner. For instance, EarthMarker~\cite{zhang2024earthmarker} supports not only language instructions but also visual prompts (\emph{e.g.} boxes and points), enabling the model to perform fine-grained image understanding, such as region/point-level captioning and referring object classification. TEOChat~\cite{irvin2024teochat} excels in analyzing time-series remote sensing images, detecting changes of interest (in the form of the bounding
box), and answering questions related to changes. Coincidentally, LHRS-Bot~\cite{muhtar2024lhrs} and VHM~\cite{pang2024vhm} both feature the ability to qualitatively recognize object attributes (\emph{e.g.} color) and image properties (\emph{e.g.} resolution and modality), while also quantitatively counting objects. Additionally, VHM offers insights into model honesty, which is vital for applications such as national defense security. Beyond single-object analysis, relationship analysis between objects is increasingly recognized as essential for understanding complex remote sensing scenes, drawing attention in recent models~\cite{muhtar2024lhrs,pang2024vhm,zhang2024earthmarker,luo2024skysensegpt}. This section delves into these unique capabilities, with a focus on their implementations as detailed.

(1) \textit{Fine-grained Image Understanding:} Region-level image understanding is challenging but achievable. One can include the coordinates of the target region in language instructions to direct the model's attention to a specific local region, as demonstrated in EarthGPT~\cite{zhang2024earthgpt} and GeoChat~\cite{kuckreja2024geochat}. While using precise coordinates is effective, it lacks flexibility and is challenging to extend to the point level. Differently, EarthMarker~\cite{zhang2024earthmarker} leverages visual prompting marks to guide the model to interpret specific regions or points. To ensure the model understands the relationship between visual prompts and the whole image, the visual prompts share the same visual encoder and projection layer with the input image. Their embeddings are combined with instruction embeddings before being fed into the LLM. Visual Prompts allow EarthMarker to perform multi-granularity RS image interpretation at the image, region, and point levels. 

\begin{figure}
\centering
\includegraphics[width=\linewidth]{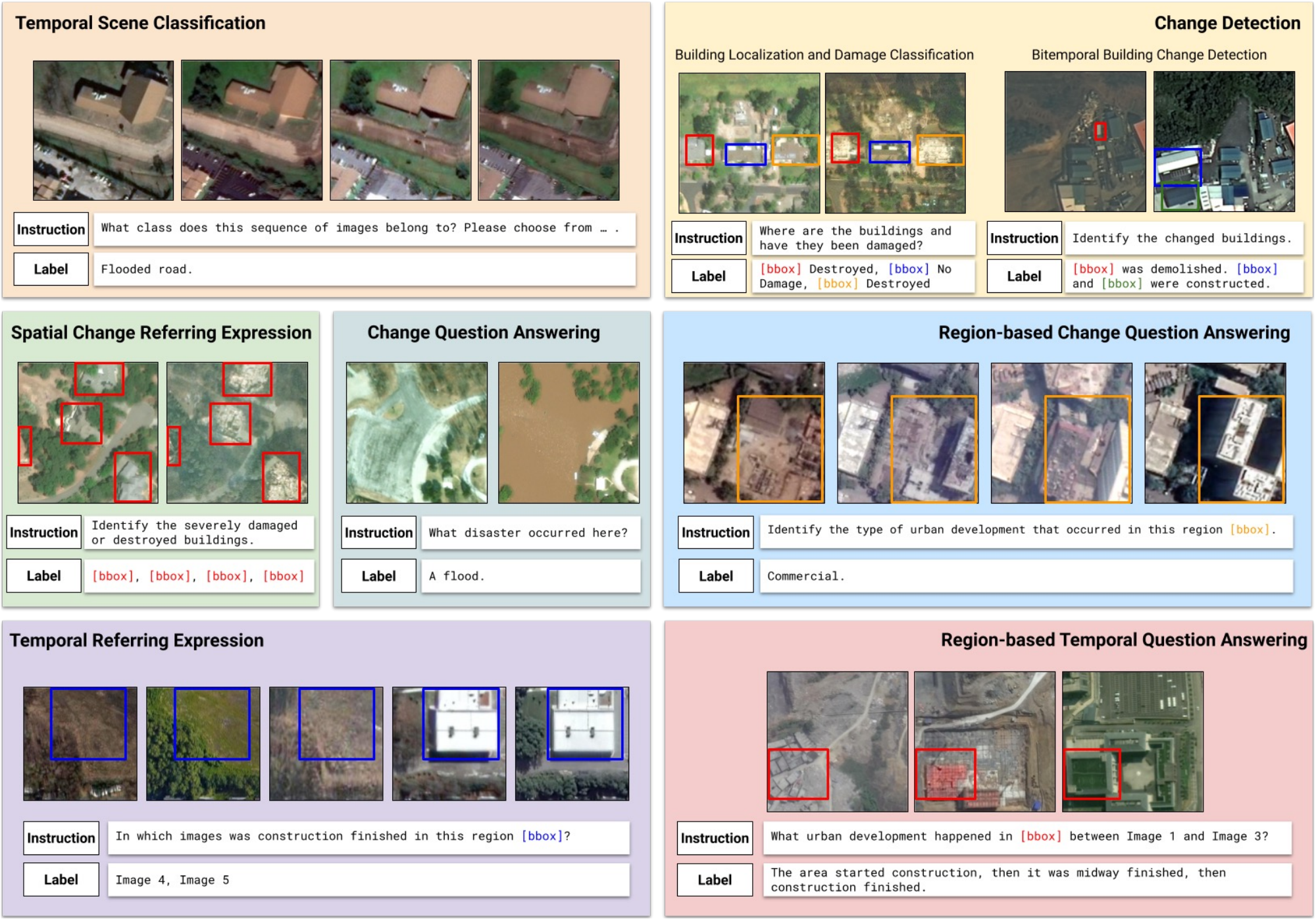}
\caption{Examples showcasing the capabilities of TEOChat~\cite{irvin2024teochat} in time-series image analysis.}
\label{fig:teochat}
\end{figure}

(2) \textit{Time-Series Image Analysis:} Change detection, the process of identifying changes on the Earth's surface from sequences of remote sensing images captured over the same area, plays an important role in many applications such as urban planning~\cite{huang2020automatic} and war damage assessment~\cite{holail2024time}. The work TEOChat~\cite{irvin2024teochat} pioneers change detection in a conversation manner. Given a sequence of remote sensing images, TEOChat employs a shared vision encoder to generate embeddings for each image. These embeddings are then projected to the input space of the LLM via a 2-layer MLP. The LLM processes these projected image embeddings alongside instruction embeddings to produce change detection results in the form of bounding boxes. To train TEOChat, the authors of \cite{irvin2024teochat} develop TEOChatlas, the first instruction-following dataset tailored for time-series image analysis tasks. Leveraging this dataset, TEOChat extends its capabilities beyond change detection to include tasks such as change question answering, temporal scene classification, temporal referring expression, and spatial change referring expression. Fig.~\ref{fig:teochat} provides specific examples illustrating each of these tasks.

(3) \textit{From Qualitative Recognition to Quantitative Analysis:} Existing models mostly concentrate on qualitative recognition tasks, excelling at describing the attributes or categories of images and objects in a non-numerical manner. In contrast, LHRS-Bot~\cite{muhtar2024lhrs} and VHM~\cite{pang2024vhm} broaden their capabilities to include quantitative image analysis, answering ``how many" questions. LHRS-Bot is equipped to count objects in an image and estimate image resolution, using single-choice questions with 2 to 4 candidate answers. This question format requires a certain level of user expertise, as the user needs to predefine several candidate answers, one of which must be correct. VHM goes a step further by not only handling these tasks but also measuring object size. Unlike LHRS-Bot, all quantitative tasks in VHM adopt open-ended question formats, which may be more suitable for practical application. On the test set of the VariousRS-Instruct~\cite{pang2024vhm}, VHM achieves mean absolute errors of 0.24 for mage resolution estimation, 6.75 for object counting, and 12.82 for geometric measurement. These results highlight the potential of vision-language models to advance quantitative analysis in remote sensing images. The successful implementation of such analysis relies heavily on the development of specialized instruction-following datasets, which are discussed in detail in Section~\ref{instruction-following-dataset}.

\begin{figure}[t]
\centering
\includegraphics[width=\linewidth]{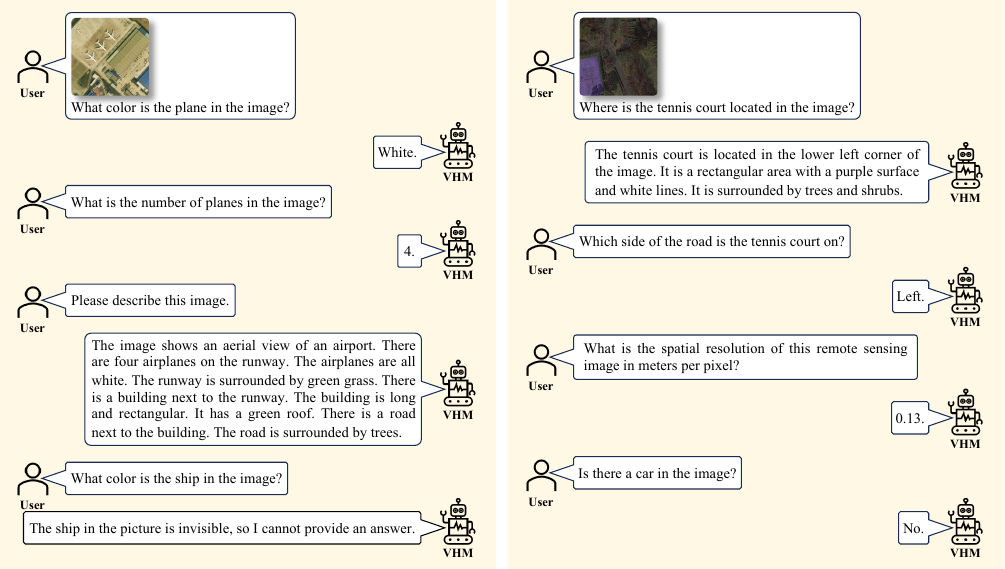}
\caption{Conversations between users and VHM~\cite{pang2024vhm}}
\label{fig:vhm-honesty}
\end{figure}

(4) \textit{Endowing Models with Honesty:} Most instruction-following datasets designed for remote sensing tasks exclusively contain factual questions that query real objects within the images and are accompanied by affirmative answers. This makes vision-language models susceptible to lying, such as providing an affirmative answer regarding a category of a non-existent object. A dishonest vision-language model is particularly problematic in remote sensing applications, especially in critical areas like national defense security, as it may lead to serious negative consequences, including inaccurate intelligence gathering. Given this, Pang et al.~\cite{pang2024vhm} are the first to offer insights into the honesty of models. They create HnstD, an RS-specific honest dataset comprising questions with factual and deceptive categories. By utilizing it as an additional instruction-following dataset, VHM is endowed with honesty. As shown in Fig.~\ref{fig:vhm-honesty}, VHM refuses to answer questions regarding the color or position of non-existent objects. 

(5) \textit{Object Relationship Understanding:} Remote sensing images often encompass a vast number of objects, owing to their long-distance imaging. Comprehensively understanding the relationships between objects within an image is essential for interpreting complex remote-sensing scenes~\cite{lin2022srsg,yang2023sagn}. Consequently, the latest models, \emph{i.e.} LHRS-Bot~\cite{muhtar2024lhrs}, VHM~\cite{pang2024vhm}, EarthMarker~\cite{zhang2024earthmarker} and SkySenseGPT~\cite{luo2024skysensegpt}, extend their capabilities to understand relationships between objects in an image. Specifically, LHRS-Bot and VHM target spatial relationships, using simple rules like ``top" or ``top right corner" to indicate object relationships. EarthMarker, on the other hand, explores functional relationships, such as the ``overpass" and ``toll station" potentially contributing to the transportation infrastructure around the ``stadium". Meanwhile, SkySenseGPT describes semantic relationships, for example, the ``crane" is over the ``boat". This promising understanding of relationships benefits from the design of specialized instruction-following datasets, as discussed in Section~\ref{instruction-following-dataset}.

\section{Generation-based Vision-Language Modeling}
Similar to research on contrastive-based vision-language modeling, generation-based vision-language modeling follows two major research directions: the construction of foundation models concerning the characteristics of remote sensing images, and their application to promote various remote sensing data analysis tasks. This section first presents the development of \textit{generative foundation models} and then discusses representative \textit{downstream applications}.

\subsection{Generative Foundation Models}
Building an effective generative foundation model is a formidable task because one needs to consider how to ensure the reliability and diversity of the generated images. 

\begin{table*}[ht]
\centering
\caption{Summary of generation-based vision-language foundation models in remote sensing.}
\label{tab:generative-VLM}
\resizebox{\textwidth}{!}{
\rowcolors{3}{gray!6}{white}
\begin{tabular}{lclcccccc}
\toprule
\multirow{2}{*}{\textbf{Model}} & \multirow{2}{*}{\textbf{Diffusion Model}} & \multirow{2}{*}{\textbf{Condition Encoder}} &  \multicolumn{3}{c}{\textbf{Condition}}  &  \multirow{2}{*}{\textbf{Training Dataset}} & \multirow{2}{*}{\textbf{GenImgType}} & \multirow{2}{*}{\textbf{Public}} \\ 
 & & & Text & Metadata & Image & & & \\ \midrule

RS-SD~\cite{zhang2024rs5m} \href{https://huggingface.co/Zilun/GeoRSSD}{[link]} & Stable Diffusion~\cite{rombach2022high} & CLIP Transformer~\cite{radford2021learning} & \ding{51} & \ding{55} &  \ding{55} & 1\% RS5M~\cite{zhang2024rs5m} & Optical & \ding{51}\\

DiffusionSat~\cite{khanna2023diffusionsat} \href{https://github.com/samar-khanna/DiffusionSat}{[link]} & Stable Diffusion~\cite{rombach2022high} & 
\begin{tabular}[l]{@{}l@{}} CLIP Transformer~\cite{radford2021learning} \\ 
Sinusoidal Projection+MLP \\ 3D ControlNet~\cite{khanna2023diffusionsat} 
\end{tabular} & \ding{51} & \begin{tabular}[c]{@{}c@{}} Latitude \\ Longitude \\ GSD \\ Cloud Cover \\ Imaging Time 
\end{tabular} & Satellite Image & \begin{tabular}[c]{@{}c@{}} fMoW~\cite{christie2018functional} \\ Satlas-small~\cite{bastani2023satlaspretrain} \\ SpaceNet~\cite{van2018spacenet} \end{tabular} & Optical  & \ding{51}\\

CRS-Diff~\cite{tang2024crs} \href{https://github.com/Sonettoo/CRS-Diff}{[link]} & Stable Diffusion~\cite{rombach2022high} & \begin{tabular}[l]{@{}l@{}} CLIP Transformer~\cite{radford2021learning} \\ 
MLP \\ ControlNet~\cite{zhang2023adding} \\ FFN \end{tabular} & \ding{51} & 
\begin{tabular}[c]{@{}c@{}} Latitude \\ Longitude \\ GSD \\ Cloud Cover \\ Imaging Time \end{tabular} & \begin{tabular}[c]{@{}c@{}} HED \\ MLSD \\ Depthmap \\ Sketch \\ Road Map \\ Seg. Mask \\ Deep Feature \end{tabular} & 
\begin{tabular}[c]{@{}c@{}} RSICD~\cite{lu2017exploring} \\ fMoW~\cite{christie2018functional} \\ Million-AID~\cite{long2021creating} \end{tabular} & Optical & \ding{51}\\

GeoSynth~\cite{sastry2024geosynth} \href{https://github.com/mvrl/GeoSynth}{[link]} & Stable Diffusion~\cite{rombach2022high} & \begin{tabular}[l]{@{}l@{}} CLIP Transformer~\cite{radford2021learning} \\ SatCLIP Location Encoder~\cite{klemmer2023satclip} \\ CoordNet~\cite{sastry2024geosynth} \\ 
ControlNet~\cite{zhang2023adding} \end{tabular} & \ding{51} & \begin{tabular}[c]{@{}c@{}} Latitude \\ Longitude \end{tabular} & \begin{tabular}[c]{@{}c@{}} OSM Image \end{tabular} & Satellite-OSM Dataset~\cite{sastry2024geosynth} & Optical & \ding{51} \\

MetaEarth~\cite{yu2024metaearth} & DDPM~\cite{ho2020denoising} & \begin{tabular}[l]{@{}l@{}} Sinusoidal Projection+MLP \\ RRDBNet~\cite{wang2018esrgan}+Upsampling \end{tabular} & 
\ding{55} & Resolution & Low-Resolution Image & Multi-Resolution Dataset~\cite{yu2024metaearth} & Optical & \ding{55} \\

Text2Earth~\cite{liu2025text2earth} & Stable Diffusion~\cite{rombach2022high} & \begin{tabular}[l]{@{}l@{}} CLIP Transformer~\cite{radford2021learning} \\ Projection Layer \\ VAE Encoder~\cite{kingma2013auto} \end{tabular} & \ding{51} & Resolution & Masked Image & Git-10M~\cite{liu2025text2earth} & Optical & \ding{55} \\

HSIGene~\cite{pang2024hsigene} \href{https://github.com/LiPang/HSIGene}{[link]} & Stable Diffusion~\cite{rombach2022high} & \begin{tabular}[l]{@{}l@{}} CLIP Transformer~\cite{radford2021learning} \\ ControlNet~\cite{zhang2023adding} \\ FFN \end{tabular} & \ding{51} & \ding{55} & \begin{tabular}[c]{@{}c@{}} HED \\ MLSD \\ Sketch \\ Seg. Mask \\ Deep Feature \end{tabular} & \begin{tabular}[c]{@{}c@{}} Xiongan~\cite{yi2020aerial} \\ Chikusei~\cite{yokoya2016airborne} \\ DFC2013~\cite{dfc2013} \\ DFC2018~\cite{dfc2018} \\ Heihe~\cite{li2017multiscale} \end{tabular} & Hyperspectral & \ding{51} \\

GPG2A~\cite{arrabi2024cross} \href{https://github.com/AhmadArrabi/GPG2A}{[link]} & Improved DDPM~\cite{nichol2021improved} & 
\begin{tabular}[l]{@{}l@{}} CLIP Transformer~\cite{radford2021learning} \\ 
ControlNet~\cite{zhang2023adding} \end{tabular} & \ding{51} & \ding{55} & Ground Image & VIGORv2~\cite{arrabi2024cross} & Optical  &  \ding{51} \\

RSDiff~\cite{sebaq2024rsdiff} & Imagen~\cite{saharia2022photorealistic} & 
T5~\cite{raffel2020exploring} & \ding{51} & \ding{55} & \ding{55} & RSICD~\cite{lu2017exploring} & Optical  & \ding{55} \\

\bottomrule
\end{tabular}
}

\vspace{3pt}
\begin{minipage}{\textwidth}
\footnotesize \textcolor{blue}{[link]} directs to model websites. \textit{GenImgType} refers to the type of generated image, \emph{e.g.} optical or hyperspectral images. \textit{GSD}, \textit{HED}, \textit{MLSD}, \textit{Seg. Mask}, and \textit{OSM} stand for \textit{Ground Sampling Distance}, \textit{Holistically-nested Edge Detection}, \textit{Multiscale Line Segment Detection}, \textit{Segmentation Mask}, and \textit{OpenStreetMap}, respectively. \textit{Public} refers to the availability of both code and model weights.
\end{minipage}
\end{table*}

\paragraph*{\textbf{Enhancing Reliability}} Text, often in the form of descriptions of ground objects within images~\cite{zhang2024rs5m,tang2024crs,sastry2024geosynth,arrabi2024cross,sebaq2024rsdiff,liu2025text2earth} or semantic categories of images~\cite{khanna2023diffusionsat,pang2024hsigene}, has been a common condition for conditional image generation. For instance, the text ``a satellite view of San Francisco showing the bay, a street and a bridge" is used to guide RS-SD~\cite{zhang2024rs5m}, while the text ``a fmow satellite image of a car dealership in the United States of America" constrains DiffusionSat~\cite{khanna2023diffusionsat}. However, it is evident that these textual descriptions alone struggle to fully encapsulate the variety of objects and intricate relationships present within a satellite image. The lack of sufficient constraint information poses a challenge to generating reliable images. To address this challenge, additional conditions, such as metadata or images, are increasingly utilized to constrain the generation process. 

(1) \textit{Metadata:} Metadata such as latitude, longitude, ground sampling distance, cloud cover, and imaging time (year, month, and day) are adopted in both DiffusionSat~\cite{khanna2023diffusionsat} and CRS-Diff~\cite{tang2024crs}. In contrast, MetaEarth~\cite{yu2024metaearth} and Text2Earth~\cite{liu2025text2earth} focus on spatial resolution, and GeoSynth~\cite{sastry2024geosynth} on geographic location (latitude and longitude). Compared to text conditions, metadata is more easily available, as it is inherently embedded within remote sensing images. Furthermore, it allows generative foundation models to be trained on large-scale image datasets, benefiting from the diverse geographic distribution of these datasets. Consequently, the key problem lies in addressing how to inject metadata conditions into diffusion models. Identical to the encoding of diffusion timesteps, DiffusionSat and MetaEarth process metadata values through sinusoidal encoding, followed by MLPs. The resulting metadata embeddings are added with timestep embeddings before being fed into the diffusion model. CRS-Diff maps metadata values to a fixed range and encodes them using different MLPs. The metadata embeddings are concatenated with text and content embeddings, providing global control information. For latitude and longitude, an alternative approach is to utilize a pre-trained location encoder. An example is \cite{sastry2024geosynth} where the authors employ SatCLIP~\cite{klemmer2023satclip} to extract location embeddings and use CoordNet, which takes location and timestep embeddings as input, to integrate these conditions into the diffusion model. CoordNet consists of 13 multi-head cross-attention blocks, each including a zero-initialized feed-forward layer.

(2) \textit{Image:} Metadata, such as geographic location, determines the visual appearance of objects in generated images to some extent. For example, the architectural styles of Chinese and European buildings are notably distinct. However, metadata primarily imposes constraints at a macro level, allowing considerable flexibility in the generated objects, which can result in uncontrollable object shapes. As a result, recent works have investigated the use of image-form conditions to enable more precise control over the image generation process. Based on the information conveyed by the image, image conditions are roughly split into low-level visual conditions and high-level semantic conditions. Low-level visual conditions pertain to the geometric information of images, such as edges, line segments, and sketches, which constrain the shape and structure of objects in generated images, as demonstrated in works~\cite{tang2024crs,pang2024hsigene}. Off-the-shelf algorithms or models, including HED~\cite{xie2015holistically}, LETR~\cite{xu2021line} and the model proposed in~\cite{simo2016learning}, are employed to obtain these image conditions. 

High-level semantic conditions, as the name implies, refer to the semantic information of images, providing constraints on object categories and their relationships in generated images. These conditions can be further categorized into two sub-groups. The first group~\cite{khanna2023diffusionsat,yu2024metaearth,arrabi2024cross,liu2025text2earth} conditions on associated remote sensing images, exemplified by works such as MetaEarth~\cite{yu2024metaearth}, which uses low-resolution images to prompt the diffusion model to produce high-resolution images, and GPG2A~\cite{arrabi2024cross}, which synthesizes aerial images guided by layout maps derived from corresponding ground images. The second group~\cite{tang2024crs,sastry2024geosynth,pang2024hsigene} leverages abstract representations of image content, as seen in works like CRS-Diff~\cite{tang2024crs} and GeoSynth~\cite{sastry2024geosynth}. CRS-Diff utilizes road maps, segmentation masks, and deep features to constrain the layout and objects within generated images, while GeoSynth employs OpenStreetMap (OSM) images to achieve similar constraints on layout and object placement.

Image-form conditions are typically injected into diffusion models using ControlNet~\cite{zhang2023adding}. ControlNet replicates the encoder blocks and the middle block of Stable Diffusion’s U-Net and incorporates several zero convolutions. It processes both conditioning representations and noisy latent representations as inputs, with its outputs added to the decoder blocks and middle block of Stable Diffusion’s U-Net. In GeoSynth~\cite{sastry2024geosynth} and GPG2A~\cite{arrabi2024cross}, ControlNet is used for condition injection with different inputs: GeoSynth provides layout maps derived from ground images, while GPG2A inputs OSM images, text, and diffusion timesteps. To enhance multi-condition injection, DiffusionSat~\cite{khanna2023diffusionsat} extends ControlNet into a 3D version capable of accepting a sequence of satellite images. This 3D ControlNet retains the replicated encoder and middle blocks, with each followed by a temporal layer consisting of a 3D zero-convolution and a temporal, pixel-wise transformer. Following Uni-ControlNet~\cite{zhao2024uni}, the authors in \cite{tang2024crs,pang2024hsigene} perform multi-scale condition injection and use attentional feature fusion~\cite{dai2021attentional} to combine conditioning and latent representations. Unlike these works relying on ControlNet, MetaEarth and Text2Earth directly concatenate latent and image condition representations. MetaEarth encodes low-resolution images via RRDBNet~\cite{wang2018esrgan} with upsampling and convolution layers, while Text2Earth processes masked images using a VAE encoder~\cite{kingma2013auto}.

\begin{figure}[t]
\centering
\includegraphics[width=\linewidth]{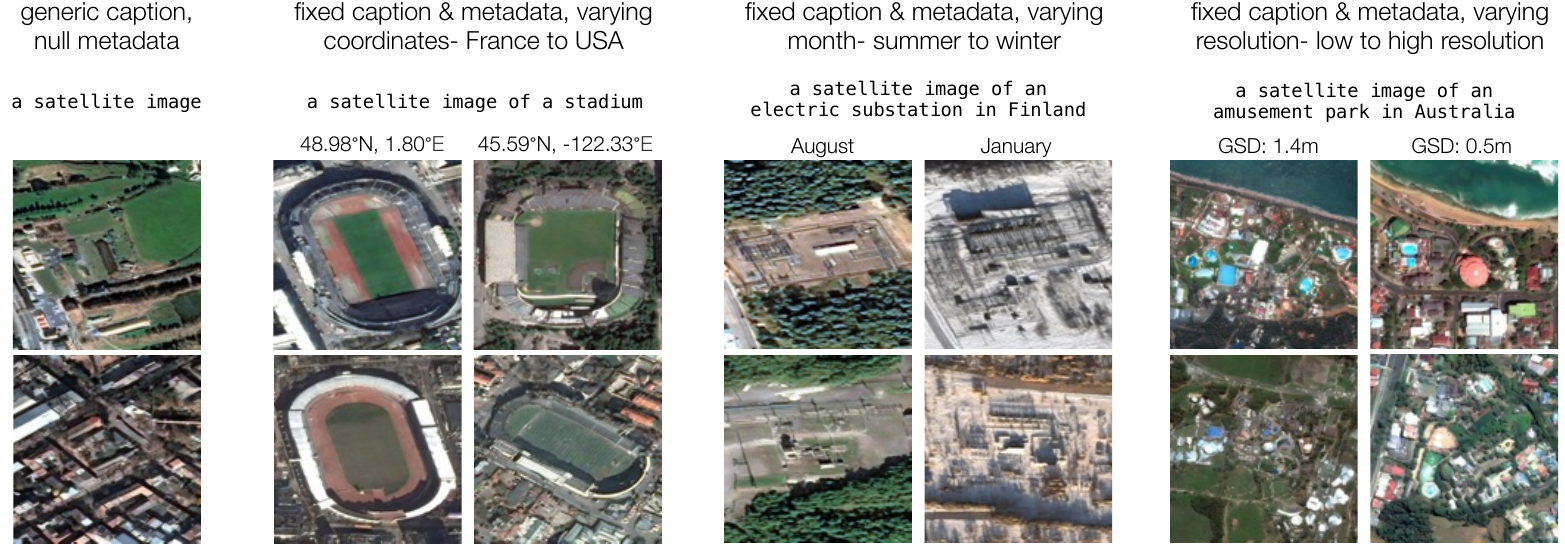}
\caption{Examples of remote sensing images generated by DiffusionSat~\cite{khanna2023diffusionsat}.}
\label{fig:diffusionsat}
\end{figure}

\begin{figure}[t]
\centering
\includegraphics[width=\linewidth]{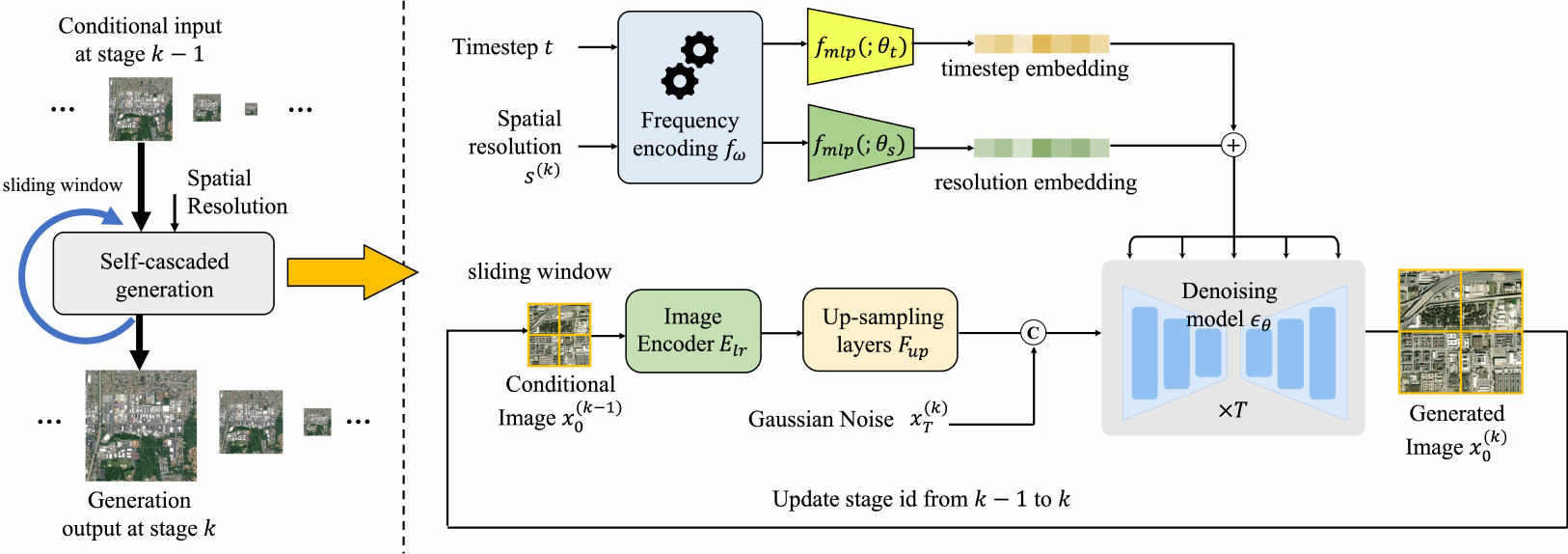}
\caption{Illustration of MetaEarth architecture~\cite{yu2024metaearth}.}
\label{fig:metaearth}
\end{figure}

\paragraph*{\textbf{Improving Diversity}} Herein, diversity refers to two aspects: first, the generated objects exhibit a variety of semantic categories, with each category featuring varied visual characteristics; and second, the generated images capture a broad range of variations in imaging condition (\emph{e.g.} season and illumination) and imaging sensors (\emph{e.g.} spatial resolution and viewpoint). By conditioning on metadata, the diversity of the generated data is improved as shown in Fig.~\ref{fig:diffusionsat}. DiffusionSat~\cite{khanna2023diffusionsat} modifies geographic coordinates, imaging time, and ground sampling distance, resulting in varied stadiums, different scenes, and multi-resolution images, respectively. In addition to considering the perspective of conditions, some works~\cite{yu2024metaearth,sebaq2024rsdiff} explore new frameworks for generating multi-resolution images. In \cite{sebaq2024rsdiff}, cascaded diffusion models~\cite{saharia2022photorealistic} are employed, where one model generates low-resolution images conditioned on text embeddings, and the other super-resolves these low-resolution images. This framework achieves the gradual generation of high-resolution satellite images purely from textual descriptions. In \cite{yu2024metaearth}, a resolution-guided self-cascading framework is proposed to generate images at various resolutions, instead of being limited to the two resolutions described in \cite{sebaq2024rsdiff}. As illustrated in Fig.~\ref{fig:metaearth}, the generation process unfolds over multiple stages, with each stage conditioned on the low-resolution image output from the preceding stage and its corresponding spatial resolution. The resolution increases at a fixed ratio with each stage. Considering the resolution gap between adjacent stages, the image condition is injected through a sequential process involving an image encoder, followed by several upsampling and convolution layers. Building on this framework, the authors introduce sliding windows with overlaps and a noise sampling strategy to generate continuous, unbounded scenes.

\paragraph*{\textbf{Performance Evaluation}} With these efforts, existing generative foundation models have demonstrated the capability to generate optical remote sensing images~\cite{zhang2024rs5m,khanna2023diffusionsat,tang2024crs,sastry2024geosynth,yu2024metaearth,arrabi2024cross,sebaq2024rsdiff} or hyperspectral remote sensing images~\cite{pang2024hsigene}. Their performance is typically evaluated using metrics such as Fr\'echet Inception Distance~\cite{heusel2017gans} and Inception Score~\cite{salimans2016improved}. For text-conditioned models, CLIP~\cite{radford2021learning} is employed to measure the similarity between generated images and their corresponding textual descriptions, as detailed in Table~\ref{tab:performance-generative-VLM}. In addition to the direct evaluations, some works assess model performance by applying them or the generated images to specific downstream tasks and measuring their impact~\cite{khanna2023diffusionsat,tang2024crs,yu2024metaearth,pang2024hsigene,arrabi2024cross}. For example, DiffusionSat~\cite{khanna2023diffusionsat} is applied to tasks such as super-resolution and temporal generation, and in-painting, while CRS-Diff~\cite{tang2024crs} uses its generated images to augment training datasets of the road detection model.

\begin{table}[t]
\centering
\caption{Performance of generative foundation models.}
\label{tab:performance-generative-VLM}
\resizebox{\linewidth}{!}{
\rowcolors{2}{gray!6}{white}
\begin{tabular}{lccccc}
\toprule
\textbf{Model} & \textbf{Condition} & \textbf{Dataset} & \textbf{FID} & \textbf{IS} & \textbf{CLIP-Score} \\ \midrule 
DiffusionSat~\cite{khanna2023diffusionsat} & Text+Metadata & fMoW~\cite{christie2018functional} & \begin{tabular}[c]{@{}c@{}} 15.80 \end{tabular} & \begin{tabular}[c]{@{}c@{}} 6.69 \end{tabular} & \begin{tabular}[c]{@{}c@{}} 17.20 \end{tabular} \\ 

CRS-Diff~\cite{tang2024crs} & \begin{tabular}[c]{@{}c@{}} Text \\ HED \\ MLSD \\ Depthmap \\ Sketch \\ Seg. Mask \\ Road Map \\ Deep Feature \end{tabular} & \begin{tabular}[c]{@{}c@{}} RSICD~\cite{lu2017exploring} 
\end{tabular} & \begin{tabular}[c]{@{}c@{}} 50.72 \\ 30.18 \\ 55.75 \\ 54.40 \\ 68.29 \\ 70.05 \\ 94.16 \\ 44.93 \end{tabular} &  \begin{tabular}[c]{@{}c@{}} 18.39 \\ - \\ - \\ - \\ - \\ - \\ - \\ - \end{tabular} & \begin{tabular}[c]{@{}c@{}} 20.33 \\ - \\ - \\ - \\ - \\ - \\ - \\ - \end{tabular} \\ 

GeoSynth~\cite{sastry2024geosynth} & \begin{tabular}[c]{@{}c@{}} Coordinate+OSM Image \\ Text+OSM Image \end{tabular} & 
\begin{tabular}[c]{@{}c@{}} Satellite-OSM~\cite{sastry2024geosynth} 
\end{tabular} & \begin{tabular}[c]{@{}c@{}} 11.90 \\ 12.97 \end{tabular} & 
\begin{tabular}[c]{@{}c@{}} - \end{tabular} & \begin{tabular}[c]{@{}c@{}} 30.30 \\ 29.80 \end{tabular} \\ 

HSIGene~\cite{pang2024hsigene} & \begin{tabular}[c]{@{}c@{}} MLSD \\ HED+MLSD \\ HED+MLSD+Sketch \\ HED+MLSD+Sketch+Seg. Mask
\end{tabular} & \begin{tabular}[c]{@{}c@{}} AID~\cite{xia2017aid} 
\end{tabular} & \begin{tabular}[c]{@{}c@{}} - \end{tabular} & \begin{tabular}[c]{@{}c@{}} - \end{tabular} & \begin{tabular}[c]{@{}c@{}} 79.27 \\ 81.52 \\ 81.40 \\ 81.83 \end{tabular} \\ 

RSDiff~\cite{sebaq2024rsdiff} & Text & RSICD~\cite{lu2017exploring} & \begin{tabular}[c]{@{}c@{}} 66.49 \end{tabular} & \begin{tabular}[c]{@{}c@{}} 7.22 \end{tabular} & \begin{tabular}[c]{@{}c@{}} - \end{tabular} \\

Text2Earth~\cite{liu2025text2earth} & Text & RSICD~\cite{lu2017exploring} & 24.49 & - & - \\

\bottomrule
\end{tabular}
}

\vspace{3pt}
\begin{minipage}{\linewidth}
\footnotesize \textit{FID} and \textit{IS} stand for \textit{Fr\'echet Inception Distance}~\cite{heusel2017gans}, \textit{Inception Score}~\cite{salimans2016improved}. \textit{CLIP-Score}~\cite{radford2021learning} measures the similarity between generated images and corresponding textual descriptions. \textit{HED}, \textit{MLSD}, \textit{Seg. Mask}, and \textit{OSM} are short for \textit{Holistically-nested Edge Detection}, \textit{Multiscale Line Segment Detection}, \textit{Segmentation Mask}, and \textit{OpenStreetMap}, respectively.
\end{minipage}
\end{table}

\subsection{Downstream Applications}
\label{generative-application}
Generative foundation models serve as powerful and versatile tools for advancing various remote sensing tasks. Their exceptional image generation capabilities have been utilized to tackle challenges such as limited image availability~\cite{liu2023diverse} and the high costs associated with annotation~\cite{zhao2023label,yuan2023efficient}. Additionally, the conditionally controllable nature of these models makes them particularly suitable for remote sensing image enhancement tasks. Examples include tasks like super-resolution~\cite{xiao2023ediffsr,meng2024conditional}, where low-resolution images serve as conditions, and cloud removal~\cite{zhao2023cloud,jing2023denoising}, where SAR images are used as conditions. Moreover, their advantages in learning diverse remote sensing image distribution equip them to handle the complexity of remote sensing scenes, improving the accuracy of remote sensing image interpretation tasks such as change detection~\cite{jia2024siamese,wen2024transc} and land cover classification~\cite{chen2023spectraldiff}. These advancements predominantly center on the ability of generative foundation models to model image distribution, which has been comprehensively reviewed in \cite{liu2024diffusion}. In this discussion, however, we shift the focus to recently emerging applications that show how the joint distribution between images and texts in generative foundation models promotes advancements in remote sensing tasks. This includes tasks such as image or change captioning~\cite{cheng2024vcc,yu2024diffusion}, pansharpening~\cite{xing2024empower}
, and zero-shot target recognition~\cite{wang2024leveraging}, as illustrated in Table~\ref{tab:application-generative-VLFM}. This section delves into representative implementations in conjunction with specific tasks.

\begin{table*}[t]
\centering
\caption{Summary of generation-based vision-language foundation models applied to remote sensing tasks.}
\label{tab:application-generative-VLFM}
\resizebox{\textwidth}{!}{
\rowcolors{2}{gray!6}{white}
\begin{tabular}{lccl}
\toprule
\textbf{Work} & \textbf{Task} & \makecell{\textbf{Generative} \\ \textbf{Foundation Model}} & \textbf{Adaptation} \\ \midrule
VCC-DiffNet~\cite{cheng2024vcc} & Image Captioning & D3PM~\cite{austin2021structured} & \makecell[{{p{7.5cm}}}]{Extract local and global features to enhance visual conditions, \\ Enhance interactions between noisy representations with visual condition by sequentially interacting global and local conditions.}  \\

Diffusion-RSCC~\cite{yu2024diffusion} & Change Captioning & Self-Design & \makecell[{{p{7.5cm}}}]{Design cross-mode feature fusion module to inject difference features of bi-temporal images, \\ Design stacking self-attention module to recreate change captions from Gaussian noise iteratively.}  \\

MADiffCC~\cite{yang2024remote} & Change Captioning & SR3~\cite{saharia2022image,bandara2022ddpm} & \makecell[{{p{7.5cm}}}]{Use a generative model to extract multi-level, multi-timestep features of bi-temporal images, \\ Design an encoder using the time-channel-spatial attention to obtain discriminative information, \\ Design a decoder guided by gated multi-head cross-attention for generating change captions.  } \\

TMDiff~\cite{xing2024empower} & Pansharpening & SR3~\cite{saharia2022image} & \makecell[{{p{7.5cm}}}]{Use text and panchromatic-multispectral image pairs together to condition the generation process, enhancing generalization.} \\

Wang et al.~\cite{wang2024leveraging} & Zero-shot SAR Target Recognition & Stable Diffusion~\cite{rombach2022high} & \makecell[{{p{7.5cm}}}]{Generate optical images conditioned on target semantic information, aiding in the creation of 3D SAR target models. }\\

Czerkawski et al.~\cite{czerkawski2024exploring} & Cloud Removal  & Stable Diffusion~\cite{rombach2022high} & \makecell[{{p{7.5cm}}}]{Use text and historical edge information to constrain cloud-free image generation.} \\

UP-Diff~\cite{wang2024up} & Urban Prediction & Stable Diffusion~\cite{rombach2022high}  & \makecell[{{p{7.5cm}}}]{Use text, current urban layouts and planned change maps as conditions, guiding the model to generate future urban layouts.} \\ 

\bottomrule
\end{tabular}
}
\end{table*}

\paragraph*{\textbf{Captioning}} Recent works~\cite{cheng2024vcc,yu2024diffusion} propose generating textual descriptions conditioned on visual features, rather than generating images from textual descriptions, thus applying generative models for captioning remote sensing images. In \cite{cheng2024vcc}, the authors focus on extracting both global and local features of remote sensing images to enhance the visual conditions for the generative model, addressing challenges such as intra-class diversity, inter-class similarity, and varying object sizes. They then improve the interaction between noisy representations and visual conditions in the decoder by sequentially interacting with global and local conditions, aiming for generated captions that better align with the image content. In \cite{yu2024diffusion}, difference features of bi-temporal images condition the generative model to generate change captions from standard Gaussian noise. These difference features are integrated into the generative model using cross-attention, followed by stacking self-attention. Inspired by the success of generative models in change detection~\cite{bandara2022ddpm}, the authors in \cite{yang2024remote} employ these models as a powerful feature extractor for multi-level, multi-timestep features of bi-temporal images. A difference encoder, based on a time-channel-spatial attention mechanism, is then utilized to extract discriminative information from the features of bi-temporal images, and a decoder generates change captions from this information.

\paragraph*{\textbf{Pansharpening}} is a process of fusing high-resolution panchromatic images with low-resolution multispectral images to produce high-resolution multispectral images. The idea of applying generative foundation models to this task involves formulating it as an image generation problem conditioned on panchromatic and multispectral images. To enhance the generalizability of the pansharpening model and enable it to uniformly handle multispectral images from different satellites, TMDiff~\cite{xing2024empower} incorporates textual descriptions, such as ``Text Prompts of GaoFen-2 Satellite", as identifiers to specify the satellites that capture multispectral images. These descriptions, combined with panchromatic-multispectral image pairs, jointly guide the generation process. Specifically, the textual descriptions are processed by a physics-informed HyperNet to get text embeddings, which are then used to modulate the denoising network. Meanwhile, the image pair is injected into the decoder of the denoising network via a condition encoder that shares the same architecture as the network's encoder. To address the variability in spectral band configurations among different satellites, TMDiff, built on the U-Net architecture, organizes each block with stacked modulated ResBlock, swish activation layer, and frequency-aware downsampling (or upsampling). Compared to PanDiff~\cite{meng2023pandiff}, which relies solely on image conditions, TMDiff demonstrates impressive generalization capabilities.

\paragraph*{\textbf{Zero-shot SAR Target Recognition}} Due to the high costs associated with SAR imaging and the limited azimuth range for capturing images, SAR datasets are often small, making it essential to explore SAR target recognition in zero-shot settings. From the perspective of SAR image simulation, adopting generative foundation models for this task is a natural solution. Typically, Wang et al.~\cite{wang2024leveraging} employ generative models to create 3D models of targets conditioned on semantic information, such as category and structure of targets. Since most existing generative models are trained on RGB natural images or optical remote sensing images, the authors adopt a two-step process: they first use the generative model to generate optical images based on the target semantic information, and then transform these optical images into into 3D models using TripoSR~\cite{tochilkin2024triposr}, rather than directly generating 3D models or SAR images. 

\paragraph*{\textbf{Cloud Removal}} Passive remote sensing is susceptible to cloud interference. By conditioning on cloud-contaminated images and SAR images, generative models have been employed for cloud removal of remote sensing images~\cite{zhao2023cloud,jing2023denoising}. Such solutions require geographically aligned multimodal image pairs, which are challenging to obtain in practical applications. To overcome this limitation, Czerkawski et al.~\cite{czerkawski2024exploring} propose a novel approach that uses historical edge information and textual descriptions (\emph{e.g.} a cloud-free satellite image) to constrain the generation process. Experimental results suggest that general-purpose generative models may not be directly applicable for cloud removal of remote sensing images, as they tend to generate undesirable artifacts in cloud-contaminated areas.

\paragraph*{\textbf{Urban Prediction}} aims to forecast future urban layouts based on current urban layouts and planned change maps, providing support for urban planning. The pioneering work, UP-Diff~\cite{wang2024up}, incorporates ConvNeXt~\cite{liu2022convnet} to encode current urban layouts and planned change maps into embeddings, which are subsequently injected into the decoder of diffusion models through cross-attention layers. Additionally, the text condition, derived from encoding ``A remote sensing photo", guides the model to generate outputs in a satellite image style.

\section{Datasets}
Large-scale datasets are an essential prerequisite for vision-language research under the two-stage paradigm. Thus, a considerable amount of current research~\cite{hu2023rsgpt,zhang2024rs5m,luo2024skysensegpt,wang2024skyscript,muhtar2024lhrs,pang2024vhm,li2024vrsbench,yuan2024chatearthnet,zhou2024urbench} focuses on dataset construction. In this section, we summarize these efforts, detailing the properties of datasets and their creation methods, with the aim of providing both convenience and inspiration for further research. Based on their usage, existing datasets can be broadly categorized into three groups: \textit{pre-training datasets}, \textit{instruction-following datasets}, and \textit{benchmark datasets}. 

\subsection{Pre-training Datasets}
Pre-training datasets, which consist of remote sensing images and their corresponding texts, play a crucial role in infusing the model with a broad range of visual and language concepts. This requires the images in pre-training datasets to be sufficiently diverse and rich. Currently, there are two alternative sources for collecting images: one is combining various open-source remote sensing image datasets~\cite{hu2023rsgpt,zhang2024rs5m,pang2024vhm,liu2024remoteclip,wei2023vlca,yuan2024chatearthnet,liu2025text2earth}, and the other is utilizing public geographic databases~\cite{wang2024skyscript,muhtar2024lhrs,zhao2024luojiahog,ge2024rsteller,liu2025text2earth}. Once the images are collected, corresponding textual descriptions can be generated through manual annotation~\cite{hu2023rsgpt,zhao2024luojiahog}. Although manual annotation is easy to implement and highly accurate, the complexity and diversity of remote sensing images make it costly, which can significantly limit the dataset size. As a result, there is a growing shift toward automatic image captioning using rule-based methods~\cite{wang2024skyscript,liu2024remoteclip} or off-the-shelf models~\cite{zhang2024rs5m,muhtar2024lhrs,pang2024vhm,yuan2024chatearthnet,ge2024rsteller,liu2025text2earth}. This section begins by presenting the image collection strategies of existing pre-training datasets, followed by an introduction to their caption generation methods.

\begin{table*}[t]
\centering
\caption{Summary of pre-training datasets for vision-language modeling in remote sensing.}
\label{tab:pretraining-datasets}
\resizebox{\textwidth}{!}{
\rowcolors{2}{gray!6}{white}
\begin{tabular}{lclllcclccc}
\toprule
\textbf{Dataset} & \textbf{\#Pairs} & \multicolumn{3}{l}{\textbf{Image Source}} & \textbf{Image Size} & 
\begin{tabular}[c]{@{}c@{}}
\textbf{Image} \\ \textbf{Resolution (m)}
\end{tabular} & 
\textbf{Caption Generation} & 
\begin{tabular}[c]{@{}c@{}}
\textbf{\#Captions} \\ \textbf{Per Image}
\end{tabular} & 
\begin{tabular}[c]{@{}c@{}}
\textbf{Avg. Cap.} \\ \textbf{Length}
\end{tabular} & 
\textbf{Public} \\ \midrule

RSICap~\cite{hu2023rsgpt} \href{https://github.com/Lavender105/RSGPT?tab=readme-ov-file}{[link]} & 2,585 & \multicolumn{3}{l}{DOTA-v1.5~\cite{ding2021object}}  & 512$\times$512 & - & Manual Annotation & 1 & 60 & \ding{51} \\

DIOR-Captions~\cite{wei2023vlca} & 16,565 & \multicolumn{3}{l}{DIOR~\cite{li2020object}} & 800$\times$800 & - & Manual Annotation & 2 & 11 & \ding{55} \\

ChatEarthNet~\cite{yuan2024chatearthnet} \href{https://zenodo.org/records/11003436}{[link]} & 173,488 & 
\multicolumn{3}{l}{SatlasPretrain~\cite{bastani2023satlaspretrain}} & 256$\times$256 & 10 & \begin{tabular}[l]{@{}l@{}} ChatGPT-3.5~\cite{chatgpt} \\ ChatGPT-4V~\cite{chatgpt} \end{tabular} & 1$\sim$2 & 155 / 90  & \ding{51} \\

RET-3+SEG-4 +DET-10~\cite{liu2024remoteclip} \href{https://huggingface.co/datasets/gzqy1026/RemoteCLIP}{[link]} & 828,725 & \begin{tabular}[l]{@{}l@{}} AUAIR~\cite{vujasinovic2020integration} \\ DOTA~\cite{ding2021object} \\ iSAID~\cite{waqas2019isaid} \\ Potsdam~\cite{potsdam} \\ RSOD~\cite{long2017accurate} \\ Vaihingen~\cite{vaihingen} \end{tabular} & \begin{tabular}[l]{@{}l@{}} CARPK~\cite{hsieh2017drone} \\ HRRSD~\cite{zhang2019hierarchical} \\ LEVIR~\cite{chen2020spatial} \\ RSICD~\cite{lu2017exploring} \\ Standford~\cite{robicquet2016learning} \\ VisDrone~\cite{zhu2021detection} \end{tabular} & \begin{tabular}[l]{@{}l@{}} DIOR~\cite{li2020object} \\ HRSC~\cite{liu2017high} \\ LoveDA~\cite{wang2021loveda} \\ RSITMD~\cite{yuan2022exploring} \\ UCM~\cite{yang2010bag} \\ \quad \end{tabular} & \begin{tabular}[c]{@{}c@{}} 224$\times$224 \\ $\sim$1920$\times$1080 \end{tabular} & - & M2C+B2C Generation~\cite{liu2024remoteclip} & 5 & - & \ding{51} \\

VersaD~\cite{pang2024vhm} \href{https://github.com/opendatalab/VHM}{[link]} & 1,390,405 & 
\begin{tabular}[l]{@{}l@{}} CrowdAI~\cite{crowdai2018} \\ fMoW~\cite{christie2018functional} \end{tabular} &
\begin{tabular}[l]{@{}l@{}} CVACT~\cite{liu2019lending} \\ LoveDA~\cite{wang2021loveda} \end{tabular} &
\begin{tabular}[l]{@{}l@{}} CVUSA~\cite{workman2015wide} \\ Million-AID~\cite{long2021creating} \end{tabular} & 512$\times$512 & 0.08$\sim$153 & Gemini-Vision~\cite{team2023gemini} & 1 & 369 & \ding{51} \\

RS5M~\cite{zhang2024rs5m} \href{https://huggingface.co/datasets/Zilun/RS5M}{[link]} & 5,062,377 & 
\begin{tabular}[l]{@{}l@{}} BigEarthNet~\cite{sumbul2019bigearthnet} \\ Million-AID~\cite{long2021creating} \end{tabular} & \begin{tabular}[l]{@{}l@{}} fMoW~\cite{christie2018functional} \\ \quad \end{tabular} & \begin{tabular}[l]{@{}l@{}} PUB11~\cite{zhang2024rs5m} \\ \quad \end{tabular} & - & - & BLIP-2 (6.7B)~\cite{li2023blip} & 1$\sim$5 & 49 / 87  & \ding{51} \\

Git-10M~\cite{liu2025text2earth} \href{https://modelscope.cn/datasets/lcybuaa1111/Git-10M/}{[link]} & $>$10,000,000 &\begin{tabular}[l]{@{}l@{}} DIOR~\cite{li2020object} \\ Million-AID~\cite{long2021creating} \\ SSL4EO-S12~\cite{wang2023ssl4eo} \end{tabular} & \begin{tabular}[l]{@{}l@{}} GeoPile~\cite{mendieta2023towards} \\ RSICB~\cite{li2017rsi} \\ \quad \end{tabular} & \begin{tabular}[l]{@{}l@{}} Google Earth \\ SkyScript~\cite{wang2024skyscript} \\ \quad \end{tabular}& - & 0.5$\sim$128 & ChatGPT-4o~\cite{chatgpt} & 1 & 52 & \ding{51} \\

LuoJiaHOG~\cite{zhao2024luojiahog} & 94,856 & \multicolumn{3}{l}{Google Maps} & 1280$\times$1280 & - & 
\begin{tabular}[l]{@{}l@{}} Manual Annotation \\ MiniGPT-4~\cite{zhu2023minigpt} \end{tabular} & 1 & 124 & \ding{55} \\

LHRS-Align~\cite{muhtar2024lhrs}\href{https://github.com/NJU-LHRS/LHRS-Bot}{[link]} & 1,150,000 & \multicolumn{3}{l}{Google Earth} & 768$\times$768 & 1 & Vicuna-v1.5 (13B)~\cite{vicuna2023} & 1 & - &\ding{51} \\

RSTeller~\cite{ge2024rsteller} \href{https://huggingface.co/datasets/SlytherinGe/RSTeller}{[link]} & 2,539,256 & 
\multicolumn{3}{l}{Google Earth Engine} & 448$\times$448 & 0.6 & Mixtral-7B~\cite{jiang2024mixtral} & 2$\sim$5 & 54 & \ding{51} \\

SkyScript~\cite{wang2024skyscript} \href{https://github.com/wangzhecheng/SkyScript}{[link]} & 2,600,000  & \multicolumn{3}{l}{Google Earth Engine} & - & 0.1$\sim$30 & Rule-based Assembly~\cite{wang2024skyscript} & 2 & - & \ding{51} \\

\bottomrule
\end{tabular}
}

\vspace{3pt}
\begin{minipage}{\textwidth}
\footnotesize \textcolor{blue}{[link]} directs to dataset websites. The average caption length (Avg. Cap. Length) in ChatEarthNet is calculated for captions generated by ChatGPT-3.5 and ChatGPT-4V separately, and in RS5M for PUB11 and other remote sensing datasets separately.
\end{minipage}
\end{table*}

\paragraph*{\textbf{Image Collection}} Object detection datasets in remote sensing typically feature diverse ground objects, such as  DOTA~\cite{ding2021object} containing 1,793,658 instances across 18 categories, and DIOR~\cite{li2020object} including 192,472 instances spanning 20 categories. These datasets have quickly garnered attention and have been instrumental in constructing pre-training datasets, leading to the development of RSICap~\cite{hu2023rsgpt} and DIOR-Captions~\cite{wei2023vlca}. Compared to relying on a single dataset, combining multiple datasets further enriches the diversity of images and objects while significantly increasing the dataset size. For instance, RET-3+SEG-4+DET-10~\cite{liu2024remoteclip} integrates three datasets for image retrieval, four for image segmentation, and ten for object detection. VersaD~\cite{pang2024vhm} further emphasizes multi-resolution and multi-domain coverage, drawing from datasets such as CVUSA~\cite{workman2015wide} (0.08m resolution), Million-AID~\cite{long2021creating} (0.5m to 153m resolution, covering diverse geographies), and LoveDA~\cite{wang2021loveda} (featuring urban and rural environments). RS5M~\cite{zhang2024rs5m} prioritizes large-scale image datasets, incorporating sources like BigEarthNet~\cite{sumbul2019bigearthnet} (590,326 images), fMoW~\cite{christie2018functional} (1,047,691 images), and Million-AID (over 1 million images). Moreover, RS5M integrates PUB11, a dataset of over 3 million remote sensing-related image-text pairs filtered from 11 computer vision datasets, including LAION-400M~\cite{schuhmann2021laion} and CC3M~\cite{sharma2018conceptual}. PUB11 creation involved sequential steps such as invalid image checking, deduplication, filtering using a CLIP model, and employing a remote sensing image detector. These pre-training datasets support the development of vision-language models beyond specific airborne or satellite sensors. An exception is ChatEarthNet~\cite{yuan2024chatearthnet}, which is tailored for satellite image analysis and sources images from the Sentinel-2 collected in SatlasPretrain~\cite{bastani2023satlaspretrain}.

In addition to open-source image datasets, public geographic databases contribute to the extensive collection of remote sensing images~\cite{zhao2024luojiahog,muhtar2024lhrs,wang2024skyscript,ge2024rsteller,liu2025text2earth}. For example, LuoJiaHOG~\cite{zhao2024luojiahog} derives its images from Google Maps, determining global sampling points through spatial analysis and the evaluation of landscape indices. This method ensures the inclusion of images representing diverse topographies and varying economic conditions across countries and regions. Leveraging a customized OpenStreetMap (OSM) database, LHRS-Align~\cite{muhtar2024lhrs} collects remote sensing images from Google Earth, ensuring their centers are aligned with OSM features. These images then undergo a series of processing steps, including resizing to a uniform size, deduplication to remove images dominated by vast ocean areas or obscured by clouds, and pruning using a trained network. The final dataset encompasses images from 9,259 cities across 129 countries. Similarly, Git-10M~\cite{liu2025text2earth} consists mainly of remote sensing images from Google Earth, which are gathered from both randomly selected and manually curated regions worldwide. Manual screening is performed to discard redundant scenes, and an image enhancement model is applied to all collected images, thereby improving dataset quality. SkyScript~\cite{wang2024skyscript} and RSTeller~\cite{ge2024rsteller} are sourced from Google Earth Engine. Among these, the authors of SkyScript carefully choose 10 image collections from Google Earth Engine, such as SWISSIMAGE 10cm RGB imagery and Landsat 9 C2 T1 TOA Reflectance, forming a multi-source, multi-resolution image pool. In contrast, RSTeller relies solely on one image collection, the National Agriculture Imagery Program, which provides aerial images covering most of the continental United States and parts of Hawaii, with a ground sampling distance of 0.6m.

\paragraph*{\textbf{Caption Generation}} Texts associated with remote sensing images in pre-training datasets, also referred to as image captions, are typically human-understandable sentences that describe various aspects of the images. These descriptions often include information about image properties, overall scenes, and specific local objects. In existing pre-training datasets, only RSICap~\cite{hu2023rsgpt} and DIOR-Captions~\cite{wei2023vlca} rely entirely on manual annotation to generate image captions. In \cite{hu2023rsgpt}, five remote sensing experts carried out the annotation process following predefined principles. The principles consist of three points: first, describing image attributes; second, describing object attributes; and third, providing a description of the overall scene before detailing specific objects. In \cite{wei2023vlca}, the authors refer to bounding boxes from the original image dataset to describe the most prominent or abundant objects. Although manual annotation guarantees high-quality image-text pairs, it is a time-intensive and laborious process, particularly for large-scale datasets. To address this, later datasets have adopted automated methods, employing rules or models to generate captions. This approach significantly reduces the cost of dataset creation while greatly expanding its scale.

(1) \textit{Rule-based Captioning:} Motivated by the fact that open-source image datasets are accompanied by image semantic information in the form of bounding boxes, segmentation masks, and class names, Liu et al.~\cite{liu2024remoteclip} propose the box-to-caption (B2C) and mask-to-box (M2B) methods to convert heterogeneous annotations into natural language captions. In detail, B2C, designed for bounding box annotations, generates five distinct captions for each image by considering the location (bounding box center) and the number of objects. Meanwhile, M2B, tailored for segmentation mask annotations, converts segmentation masks into bounding boxes by first extracting contours for each class and then sorting the contour points to define the bounding boxes. By applying M2B followed by B2C, it becomes convenient to create image-text datasets from image segmentation datasets. With a similar motivation, Wang et al.~\cite{wang2024skyscript} leverage semantic information from OSM to caption remote sensing images from Google Earth Engine, as the two can be linked through geo-coordinates. Due to its uncurated nature, semantic information from OSM is rich but often messy. Consequently, not all this information is suitable for describing images, especially if it is not visually discernible. To address this, the authors leverage CLIP embeddings of semantic information as input and apply a binary logistic regression model to filter out invisible information. After filtering, a description for each object is crafted by assembling OSM semantic data using connecting words such as ``of", ``is", ``and". Each image includes two captions: one describing the object used to determine the image boundary, and another describing multiple objects within the image by assembling captions of individual objects based on their geospatial relationships. To reduce noisy image-text pairs, they finally use a CLIP model to estimate the similarity between images and texts, and perform image filtering.

(2) \textit{Model-based Captioning:} Numerous large language models are available for generating remote sensing image captions, including ChatGPT series~\cite{chatgpt}, Gemini~\cite{team2023gemini}, Vicuna~\cite{vicuna2023}, and Mixtural~\cite{jiang2024mixtral}. To effectively apply these models, the key research focus lies in prompt design, as caption quality heavily depends on the prompt. The fundamental principle of prompt design is to align with the input requirements of the LLM being used. For multimodal LLMs such as ChatGPT-4V, Gemini-Vision, BLIP-2~\cite{li2023blip}, and MiniGPT-4~\cite{zhu2023minigpt}, which can process both text and image inputs, the prompt can simply instruct the model to generate a caption like ``generate a description for the image's visual content"~\cite{zhang2024rs5m}. However, to improve caption quality, prompts are optimized by adding constraints on the model's response~\cite{pang2024vhm,yuan2024chatearthnet,zhao2024luojiahog} or incorporating semantic information about the image~\cite{yuan2024chatearthnet,zhao2024luojiahog,liu2025text2earth}. For instance, VersaD~\cite{pang2024vhm} defines prompts to ensure the captions generated by Gemini-Vision encompass information about image properties, object attributes, and scene context, while preventing the model from describing uncertain objects. Furthermore, it imposes constraints on the format of the model's response. LuoJiaHOG~\cite{zhao2024luojiahog} prompts MiniGPT-4 to follow principles such as describing object attributes, reducing vague words, using ``next" instead of ``up", adding synonyms, and more. Incorporating image semantic information into the prompt helps guide the model's focus toward the image content of interest, while enriching the captions. For example, in ChatEarthNet~\cite{yuan2024chatearthnet}, the distribution and proportion of various land cover types in the image, derived from the European Space Agency (ESA)'s WorldCover project, are embedded into the prompt to guide ChatGPT-4V in describing the land cover types of interest. In LuoJiaHOG, manually corrected OSM semantics are provided to MiniGPT-4, allowing the model to describe objects of interest within the images. 

For models like ChatGPT-3.5, Vicuna-v1.5, and Mixtral, which primarily handle text inputs, providing image semantic information in the prompt allows the models to simulate ``seeing" the image and generate captions. LHRS-Align~\cite{muhtar2024lhrs} and RSTeller~\cite{ge2024rsteller} both utilize OSM semantics to assist the model in understanding the image. Since the quality of the semantic information directly impacts the quality of the captions, a significant portion of the effort in caption generation for these two datasets is devoted to carefully processing the semantic data. For example, similar to SkyScript~\cite{wang2024skyscript}, LHRS-Align filters out invisible semantic information using rules, followed by manual inspection. Additionally, LHRS-Align removes duplicate semantics for each image and applies a threshold to ensure semantic balance across the entire dataset. Given that an image is often associated with abundant OSM semantic data, providing all of it to the LLM can be overwhelming, making it difficult for the model to generate accurate captions. Therefore, RSTeller focuses on describing the primary object in each image, defined by the largest size or longest length, and refers to the OSM Wiki to interpret the OSM semantics, thereby reducing the ambiguity of the semantic information.  Based on the processed OSM semantic data, LHRS-Align and RSTeller use custom templates to integrate semantic information into the prompt. To enhance the model's understanding of the captioning task, a few examples are provided to the model.

Based on the captions generated by models, some techniques are proposed to enrich the captions further. In RS5M~\cite{zhang2024rs5m}, the meta-information of images, \emph{e.g.} longitude, latitude, and timestamp, is structured into readable sentences and combined with the generated captions. Additionally, RS5M includes rotation-invariant captions, which are those that exhibit the most stable similarity to the image features, regardless of their rotation, among the generated captions. In RSTeller, the LLM is guided to create multiple revisions from the generated captions with different tones, resulting in each image being accompanied by at least two captions, and up to five in total.

\paragraph*{\textbf{Dataset Property}} Table~\ref{tab:pretraining-datasets} provides an overview of existing pre-training datasets for vision-language modeling in remote sensing, with several key points worth noting: 

(1) \textit{Geographic Coverage:} The datasets RS5M, VersaD, and ChatEarthNet benefit from large-scale remote sensing image datasets, while SkyScript, LHRS-Align, LuoJiaHOG, and Git-10M derive from public geographic databases. Each of these datasets provides relatively comprehensive global coverage, with Git-10M being the largest (over 10 million image-text pairs). In contrast, RSTeller's coverage is primarily limited to the continental United States and parts of Hawaii, potentially introducing geographic bias in trained models.

(2) \textit{Scene Diversity:} Large-scale datasets such as RS5M, LHRS-Align, and SkyScript have, to some extent, ensured diversity in remote sensing scenes. Taking this a step further, the creators of LuoJiaHOG employ geospatial analysis to collect images from global regions with varied topography and different development levels. Meanwhile, the Git-10M team manually selects specific areas to ensure comprehensive coverage of representative scenes, including urban areas, forests, mountains, and deserts.

(3) \textit{Caption Quality:} Manually annotated or rule-generated image captions typically achieve high accuracy. For instance, manual inspection of 1,000 randomly sampled image-text pairs from SkyScript demonstrates 96.1\% precision. Conversely, model-generated captions inevitably contain errors despite considerable efforts to mitigate noise injection. For smaller-scale datasets like ChatEarthNet, manual correction can significantly improve caption quality. However, this approach becomes prohibitively expensive for million-scale datasets such as VersaD and RS5M, making caption errors unavoidable. Surprisingly, the authors of VersaD experimentally verified that models pre-trained on the noisy dataset VersaD (82.3\%) outperformed those trained on the more accurate dataset SkyScript, suggesting that rich-content and long captions may make models less sensitive to noise.

(4) \textit{Distinctive Characteristics:} Current pre-training datasets predominantly contain English text, with DIOR-Captions being the only exception that provides bilingual captions in both Chinese and English. This enables the exploration of cross-lingual vision-language alignment. Moreover, while most datasets focus on optical images, ChatEarthNet stands out by facilitating a deeper understanding of multispectral images by offering satellite images with nine specific bands, each paired with textual descriptions.

\subsection{Instruction-Following Datasets}
\label{instruction-following-dataset}
Instruction-following datasets are specifically designed for the supervised fine-tuning of instruction-based vision-language models, allowing them to perform specific remote sensing tasks. For contrastive or generation-based vision-language models, the emphasis during task-specific applications lies more on designing network architectures than on developing new datasets, as discussed in Sections~\ref{contrastive-application} and \ref{generative-application}. Therefore, in the fine-tuning phase of vision-language modeling, we exclusively summarize instruction-following datasets. This type of dataset is comprised of images paired with conversations, structured as instructions (or questions) and answers. Its construction involves image collection and conversation generation, with images primarily sourced from open-source remote sensing image datasets, as shown in Table~\ref{tab:fine-tuning-dataset}. The main challenge lies in generating conversations, which is addressed in this section, followed by an overview of notable instruction-following datasets.

\begin{table*}[ht]
\centering
\caption{Summary of instruction-following datasets for fine-tuning vision-language models in remote sensing.}
\label{tab:fine-tuning-dataset}
\resizebox{\textwidth}{!}{
\rowcolors{2}{gray!6}{white}
\begin{tabular}{lllllllcc}
\toprule
\textbf{Dataset} & \textbf{Task} & \multicolumn{5}{l}{\textbf{Data Source}} & \begin{tabular}[c]{@{}c@{}} \textbf{\#Sample} \\ \textbf{Train / Test} \end{tabular} & \textbf{Public} \\ \midrule

MMShip~\cite{zhang2024popeye} &  Object Detection & DOSR\cite{han2021fine} & DOTA ship subset~\cite{ding2021object} &  HRSID~\cite{wei2020hrsid} & SSDD~\cite{zhang2021sar} & & 81,000 / - & \ding{55} \\ 

HnstD~\cite{pang2024vhm} \href{https://github.com/opendatalab/VHM}{[link]} & Honest Question Answering & DOTA v2~\cite{ding2021object} & FAIR1M~\cite{sun2022fair1m} & & & & 45,000 / 1,642 & \ding{51} \\ 

VersaD-Instruct~\cite{pang2024vhm} \href{https://github.com/opendatalab/VHM}{[link]} & \begin{tabular}[l]{@{}l@{}} Complex Reasoning \\ Multi-Turn Conversation \end{tabular} & DIOR~\cite{li2020object} & DOTA v2~\cite{ding2021object} & FAIR1M~\cite{sun2022fair1m} & & & 30,000 / - & \ding{51} \\ 

RS-Instructions~\cite{bazi2024rs} & \begin{tabular}[l]{@{}l@{}} Image Captioning \\ Visual Question Answering \end{tabular} & RSIVQA~\cite{zheng2021mutual} & RSVQA-LR~\cite{lobry2020rsvqa} & UAV~\cite{hoxha2021novel} & UCM-captions~\cite{qu2016deep} & & 5506 / 1552 & \ding{51} \\ 

RS-GPT4V~\cite{xu2024rs} & \begin{tabular}[l]{@{}l@{}} Visual Grounding \\ Complex Reasoning \\ Image/Region Captioning \\ Visual Question Answering \end{tabular} & \begin{tabular}[l]{@{}l@{}} DIOR-RSVG~\cite{zhan2023rsvg} \\ RSIVQA~\cite{zheng2021mutual} \end{tabular} & \begin{tabular}[l]{@{}l@{}} FloodNet~\cite{rahnemoonfar2021floodnet} \\ RSVQA-HR~\cite{lobry2020rsvqa} \end{tabular} & \begin{tabular}[l]{@{}l@{}} NWPU-Captions~\cite{cheng2022nwpu} \\ RSVQA-LR~\cite{lobry2020rsvqa} \end{tabular} & \begin{tabular}[l]{@{}l@{}} RSICD~\cite{lu2017exploring} \\ Sydney-captions~\cite{qu2016deep} \end{tabular} & \begin{tabular}[l]{@{}l@{}} RSITMD~\cite{yuan2022exploring} \\ UCM-captions~\cite{qu2016deep} \end{tabular} & 991,206 / 258,419 &\ding{55}  \\

SkyEye-968k~\cite{zhan2024skyeyegpt} \href{https://huggingface.co/datasets/ZhanYang-nwpu/SkyEye-968k}{[link]} & \begin{tabular}[l]{@{}l@{}} Visual Grounding \\ Image/Video Captioning \\ Visual Question Answering \\ Multi-Turn Conversation \end{tabular}  
& \begin{tabular}[l]{@{}l@{}} CapERA~\cite{bashmal2023capera} \\ NWPU-Captions~\cite{cheng2022nwpu} \\ RSVQA-HR~\cite{lobry2020rsvqa} \\  UCM-captions~\cite{qu2016deep}  \end{tabular}
& \begin{tabular}[l]{@{}l@{}} DIOR-RSVG~\cite{zhan2023rsvg} \\ RSICD~\cite{lu2017exploring} \\ RSVQA-LR~\cite{lobry2020rsvqa} \\ UCM-Conversa~\cite{zhan2024skyeyegpt}  \end{tabular}
& \begin{tabular}[l]{@{}l@{}} DIOR-Conversa~\cite{zhan2024skyeyegpt} \\ RSITMD~\cite{yuan2022exploring} \\ RSVG~\cite{sun2022visual} \\ \quad  \end{tabular}
& \begin{tabular}[l]{@{}l@{}} DOTA-Conversa~\cite{zhan2024skyeyegpt} \\ RSIVQA~\cite{zheng2021mutual}  \\   Sydney-captions~\cite{qu2016deep} \\ \quad \end{tabular}
& \begin{tabular}[l]{@{}l@{}} ERA-VQA~\cite{zhan2024skyeyegpt} \\ RSPG~\cite{zhan2024skyeyegpt} \\ Sydney-Conversa~\cite{zhan2024skyeyegpt} \\ \quad \end{tabular}
& 968,000 / - & \ding{51} \\

Multi-task Dataset~\cite{muhtar2024lhrs} \href{https://github.com/NJU-LHRS/LHRS-Bot?tab=readme-ov-file}{[link]} & \begin{tabular}[l]{@{}l@{}} Visual Grounding \\ Scene Classification \\ Image Captioning \\Visual Question Answering  \end{tabular} & \begin{tabular}[l]{@{}l@{}} DIOR-RSVG~\cite{zhan2023rsvg} \\ RSITMD~\cite{yuan2022exploring} \end{tabular} & \begin{tabular}[l]{@{}l@{}} fMoW~\cite{christie2018functional} \\ RSVG~\cite{sun2022visual} \end{tabular} & \begin{tabular}[l]{@{}l@{}} METER-ML~\cite{zhu2022meter} \\ RSVQA-HR~\cite{lobry2020rsvqa} \end{tabular} & \begin{tabular}[l]{@{}l@{}} NWPU-RESISC45~\cite{cheng2017remote} \\ RSVQA-LR~\cite{lobry2020rsvqa} \end{tabular} &\begin{tabular}[l]{@{}l@{}} RSICD~\cite{lu2017exploring} \\ UCM-captions~\cite{qu2016deep} \end{tabular}  & 42,322 / - & \ding{51} \\

TITANIC-FGS~\cite{guo2024ifship} & \begin{tabular}[l]{@{}l@{}} Fine-Grained Ship Classification \\  Ship Image Captioning \\ Ship Image Visual Question Answering \\  Multi-Turn Conversation \end{tabular} & Google & Baidu & & & & 16,876 / 2,053 & \ding{55} \\

RSVP-3M~\cite{zhang2024earthmarker} & \begin{tabular}[l]{@{}l@{}} Scene/Region/Point Classification \\ Image/Region/Point Captioning \\ Object Relationship Reasoning \\ Multi-Turn Conversation \end{tabular} 
& \begin{tabular}[l]{@{}l@{}} DIOR-RSVG~\cite{zhan2023rsvg} \\ Hi-UCD~\cite{tian2020hi} \\ LEVIR~\cite{chen2020spatial} \\ Optimal-31~\cite{wang2018scene} \\ SOTA~\cite{wang2024samrs} \\ VisDrone~\cite{zhu2021detection} \end{tabular}& \begin{tabular}[l]{@{}l@{}} DOSR\cite{han2021fine} \\ HRRSD~\cite{zhang2019hierarchical} \\ MAR20~\cite{wenqi2024mar20} \\ Potsdam~\cite{potsdam} \\ SOTA~\cite{wang2024samrs} \\ WHU~\cite{ji2018fully}  \end{tabular}&\begin{tabular}[l]{@{}l@{}}  DOTA v2~\cite{ding2021object} \\ HRSC2016~\cite{liu2017high} \\ NWPU-Captions~\cite{cheng2022nwpu} \\ RSD46-WHU~\cite{long2017accurate} \\ UAVid~\cite{lyu2020uavid} \\ WHU-RS19~\cite{dai2010satellite} 
\end{tabular}& \begin{tabular}[l]{@{}l@{}} FAIR1M~\cite{sun2022fair1m} iSAID~\cite{waqas2019isaid} \\ NWPU-RESISC45~\cite{cheng2017remote} \\ RSITMD~\cite{yuan2022exploring} \\ UCAS-AOD~\cite{zhu2015orientation} \\ \quad \end{tabular}& \begin{tabular}[l]{@{}l@{}} FAST~\cite{wang2024samrs} \\ Kuckreja et al.~\cite{kuckreja2024geochat} \\ NWPUVHR10~\cite{li2017rotation} \\ RSOD~\cite{long2017accurate} \\ Vaihingen~\cite{vaihingen} \\ \quad \end{tabular} & 3,648,884 / - & \ding{55} \\

Kuckreja et al.~\cite{kuckreja2024geochat} \href{https://github.com/mbzuai-oryx/geochat}{[link]} & \begin{tabular}[l]{@{}l@{}} Visual Grounding \\ Complex Reasoning \\ Scene Classification \\ Image/Region Captioning \\ Visual Question Answering \\ Multi-Turn Conversation  \end{tabular} & \begin{tabular}[l]{@{}l@{}} DIOR~\cite{li2020object} \\ RSVQA~\cite{lobry2020rsvqa} \end{tabular}  & \begin{tabular}[l]{@{}l@{}} DOTA~\cite{ding2021object} \\ \quad \end{tabular} & \begin{tabular}[l]{@{}l@{}} FAIR1M~\cite{sun2022fair1m} \\ \quad \end{tabular} &\begin{tabular}[l]{@{}l@{}} FloodNet~\cite{rahnemoonfar2021floodnet} \\ \quad \end{tabular}  & \begin{tabular}[l]{@{}l@{}} NWPU-RESISC45~\cite{cheng2017remote} \\ \quad \end{tabular}  & 306,000 / 12,000 & \ding{51} \\

LHRS-Instruct~\cite{muhtar2024lhrs} \href{https://github.com/NJU-LHRS/LHRS-Bot?tab=readme-ov-file}{[link]} & \begin{tabular}[l]{@{}l@{}} Object Counting \\ Image Captioning \\ Complex Reasoning \\ Object Attribute Recognition \\ Object Relationship Analysis \\ Multi-Turn Conversation \end{tabular} & LHRS-Align~\cite{muhtar2024lhrs} & NWPU-Captions~\cite{cheng2022nwpu} & RSITMD~\cite{yuan2022exploring} & & & 39,800 / - & \ding{51} \\ 

MMRS-1M~\cite{zhang2024earthgpt} \href{https://github.com/wivizhang/EarthGPT}{[link]} & \begin{tabular}[l]{@{}l@{}} Visual Grounding \\ Object Detection \\ Scene Classification \\ Image/Region Captioning \\ Visual Question Answering \\ Multi-Turn Conversation \end{tabular} 
& \begin{tabular}[l]{@{}l@{}} Aerial-mancar~\cite{infiray2021aerial} \\ DOTA~\cite{ding2021object} \\ FGSCR-42~\cite{di2021public} \\ Infrared-security~\cite{infiray2021infrared} \\ RSICD~\cite{lu2017exploring} \\ RSVQA-LR~\cite{lobry2020rsvqa} \\ UCM~\cite{yang2010bag} \end{tabular} 
& \begin{tabular}[l]{@{}l@{}} AIR-SARShip-2.0~\cite{li2017rotation} \\ Double-light-vehicle~\cite{infiray2021double} \\ FloodNet~\cite{rahnemoonfar2021floodnet} \\ NWPU-Captions~\cite{cheng2022nwpu} \\ RSITMD~\cite{yuan2022exploring} \\ Sea-shipping~\cite{infiray2021sea} \\ UCM-captions~\cite{qu2016deep} \end{tabular} 
& \begin{tabular}[l]{@{}l@{}} CRSVQA~\cite{zhang2023multi} \\ DSCR~\cite{di2019public} \\ HIT-UAV~\cite{suo2023hit} \\ NWPU-RESISC45~\cite{cheng2017remote} \\ RSIVQA~\cite{zheng2021mutual} \\ SSDD~\cite{zhang2021sar} \\ WHU-RS19~\cite{dai2010satellite} \end{tabular} 
&  \begin{tabular}[l]{@{}l@{}} DIOR~\cite{li2020object} \\ EuroSAT~\cite{helber2019eurosat} \\ HRRSD~\cite{zhang2019hierarchical} \\ NWPUVHR10~\cite{li2017rotation} \\ RSOD~\cite{long2017accurate} \\ Sydney-captions~\cite{qu2016deep}\\ VisDrone~\cite{zhu2021detection}  \end{tabular} 
& \begin{tabular}[l]{@{}l@{}} DIOR-RSVG~\cite{zhan2023rsvg} \\ FAIR1M~\cite{sun2022fair1m} \\ HRSID~\cite{wei2020hrsid} \\ Oceanic ship~\cite{center2020oceanic} \\ RSSCN7~\cite{zou2015deep} \\ UCAS-AOD~\cite{zhu2015orientation} \\ \quad \end{tabular} 
& 1,005,842 / - & \ding{51} \\ 

FIT-RS~\cite{luo2024skysensegpt} \href{https://huggingface.co/datasets/ll-13/FIT-RS}{[link]} & \begin{tabular}[l]{@{}l@{}} Visual Grounding \\  Object Detection \\ Image/Region Captioning \\ Visual Question Answering \\  Multi-Label Classification \\ Object Relationship Reasoning \\ Image/Region Scene Graph Generation \\ Multi-Turn Conversation \end{tabular} & extended STAR~\cite{li2024star} & & & & & 1,440,681 / 360,170 & \ding{51} \\ 

VariousRS-Instruct~\cite{pang2024vhm} \href{https://github.com/opendatalab/VHM}{[link]}
& \begin{tabular}[l]{@{}l@{}} Visual Grounding \\ Object Counting \\Scene Classification \\Building Vectorizing \\Geometric Measurement \\ Image Property Recognition \\ Visual Question Answering \\Multi-Label Classification  \end{tabular} & \begin{tabular}[l]{@{}l@{}} BANDON~\cite{pang2023detecting} \\ FAIR1M~\cite{sun2022fair1m} \\ MSAR~\cite{xia2022crtranssar} \\ RSVQA-LR~\cite{lobry2020rsvqa} \end{tabular} & \begin{tabular}[l]{@{}l@{}} CrowdAI~\cite{crowdai2018} \\ FBP~\cite{tong2023enabling} \\ MtS-WH~\cite{wu2016scene} \\ UCM-captions~\cite{qu2016deep} \end{tabular} & \begin{tabular}[l]{@{}l@{}} DeepGlobe~\cite{demir2018deepglobe} \\ fMoW~\cite{christie2018functional} \\ NWPU-RESISC45~\cite{cheng2017remote} \\ \quad \end{tabular} & \begin{tabular}[l]{@{}l@{}} DIOR-RSVG~\cite{zhan2023rsvg} \\ GID~\cite{tong2020land} \\ Potsdam~\cite{potsdam} \\ \quad \end{tabular} & \begin{tabular}[l]{@{}l@{}} DOTA v2~\cite{ding2021object} \\ METER-ML~\cite{zhu2022meter} \\ RSITMD~\cite{yuan2022exploring} \\ \quad \end{tabular} & 76,445 / 13,020 & \ding{51} \\

TEOChatlas~\cite{irvin2024teochat} \href{https://huggingface.co/datasets/jirvin16/TEOChatlas}{[link]} & \begin{tabular}[l]{@{}l@{}}Visual Grounding \\ Change Detection \\ Scene Classification \\ Image/Region Captioning \\ Visual Question Answering \\ Temporal Scene Classification \\ Temporal Referring Expression \\ Spatial Change Referring Expression \\ Image/Region Change Question Answering \end{tabular} & fMoW~\cite{christie2018functional} & Kuckreja et al.~\cite{kuckreja2024geochat} & QFabric~\cite{verma2021qfabric} & S2Looking~\cite{shen2021s2looking} & xBD~\cite{gupta2019creating} & 554,071 / - & \ding{51} \\ 

\bottomrule
\end{tabular}
}

\vspace{3pt}
\begin{minipage}{\textwidth}
\footnotesize \textcolor{blue}{[link]} directs to dataset websites.
\end{minipage}
\end{table*}

\paragraph*{\textbf{Conversation Generation}}
Since sourced image datasets come with task-specific annotations, the most intuitive idea is to convert these annotations into a dialogue format. Currently, three methods are commonly used for this conversion: template-based transformation, large model assistance, and manual annotation. Below, we demonstrate each method with examples from specific tasks.

(1) \textit{Template-Based Transformation:} To create task-specific conversations, open-source image datasets for the given task are typically used. Answers are derived directly from dataset annotations, while instructions are either manually defined or randomly selected from a pool generated by large language models. This pool contains instructions tailored to specific tasks, conveying the same meanings with varied phrasing. For example, in MMRS-1M~\cite{zhang2024earthgpt}, classification datasets such as NWPU-RESISC45~\cite{cheng2017remote} and EuroSAT~\cite{helber2019eurosat}, along with object detection datasets like DIOR~\cite{li2020object} and DOTA~\cite{ding2021object}, are selected to create instruction-following data for classification and detection tasks. For scene classification, the authors design instructions using the template ``What is the category of this remote sensing image? Answer the question using a single word or phrase. Reference categories include category 1, ..., and category $n$". For object detection, the template is ``Detect all objects shown in the remote sensing image and describe using oriented bounding boxes". In SkyEye-968k~\cite{zhan2024skyeyegpt}, corresponding datasets are selected for image captioning and visual grounding tasks, including UCM-captions~\cite{qu2016deep}, Sydney-captions~\cite{qu2016deep}, RSVG~\cite{sun2022visual}, and DIOR-RSVG~\cite{zhan2023rsvg}. Separate instruction pools are constructed for each task. The image captioning pool includes instructions such as ``Briefly describe this image" or ``Provide a concise depiction of this image." Meanwhile, the visual grounding pool contains instructions like ``Give me the location of referring expression" or ``Where is referring expression?". Based on task-specific instruction-following data, multi-turn conversations can be generated by mixing data from different tasks~\cite{zhan2024skyeyegpt,zhang2024earthgpt}.

(2) \textit{Large Model Assistance:} Simple template-based transformations are insufficient for generating instruction-following data for complex reasoning tasks, which equip vision-language models with higher-order cognitive abilities, such as making decisions and identifying relationships. Additionally, multi-turn conversations created through multi-task mixing after template transformation often lack diversity. To address these issues, large language models are utilized to generate instruction-following data. By providing a few manually defined in-context examples, the models learn to create high-quality instruction-answer pairs based on image captions and other related information. For instance, Kuckreja et al.~\cite{kuckreja2024geochat} prompt Vicuna~\cite{vicuna2023} to generate 30,000 detailed image descriptions, 65,000 multi-turn conversations, and 10,000 complex reasoning based on short descriptions supplied. Similarly, Muhtar et al.~\cite{muhtar2024lhrs} use Vicuna~\cite{vicuna2023} to produce instruction-following data based on image captions from RSITMD~\cite{yuan2022exploring} and NWPU-Captions~\cite{cheng2022nwpu}. To ensure that the generated conversations focus on the visual content of the images, their prompts are designed to generate questions that inquire about object types, actions, locations, relative positions between objects, and more. However, considering the limitations of RSITMD and NWPU-Captions, where captions are short and lack detailed content, the authors supplemented their data with additional image-caption pairs from LHRS-Align~\cite{muhtar2024lhrs}. GPT-4 is prompted with image captions, object bounding boxes, and object attributes, to generate detailed image descriptions, complex reasoning, and multi-turn conversations that incorporate questions about object locations and counts.

(3) \textit{Manual Annotation:} In constructing instruction-following datasets, manual annotation primarily serves to provide accurate information that supports template-based or large model-assisted conversation generation methods, rather than directly creating conversations. For example, in FIT-RS~\cite{luo2024skysensegpt}, an instruction-following dataset for understanding semantic relationships between objects, Luo et al.~\cite{luo2024skysensegpt} manually annotate very high-resolution remote sensing images with detailed scene graph labels, laying the groundwork for generating conversations focused on relationship comprehension tasks. In TITANIC-FGS~\cite{guo2024ifship}, the authors manually summarize the common and private features of fine-grained objects, which are used to populate predefined templates for creating object descriptions. These descriptions are subsequently used to prompt GPT-4, enabling the generation of multi-turn conversations that simulate human-like logical decision-making.

The three methods can be combined to facilitate the construction of instruction-following datasets~\cite{luo2024skysensegpt,zhang2024earthmarker}, with the goal of creating rich and diverse samples across a variety of tasks. To help vision-language models better distinguish between tasks, a common practice is to incorporate task-specific identifiers into the instructions, as demonstrated in \cite{zhan2024skyeyegpt,muhtar2024lhrs,pang2024vhm,kuckreja2024geochat}. One example is \cite{zhan2024skyeyegpt} where the authors introduce task identifiers such as ``[caption]", ``[vqa]", and ``[refer]" to specify captioning, visual question answering, and visual grounding tasks, respectively.

\paragraph*{\textbf{Impressive Datasets}} Table~\ref{tab:fine-tuning-dataset} provides an overview of existing instruction-following datasets. MMShip~\cite{zhang2024popeye} and TITANIC-FGS~\cite{guo2024ifship} are specifically tailored for ship image analysis, whereas the others are designed for general-purpose remote sensing data analysis. Most datasets include instruction data for tasks such as captioning, visual question answering, visual grounding, and multi-turn conversations~\cite{zhang2024earthgpt,zhan2024skyeyegpt,guo2024ifship,irvin2024teochat,xu2024rs,muhtar2024lhrs,kuckreja2024geochat,luo2024skysensegpt,xu2024rs}. Among these, MMRS-1M~\cite{zhang2024earthgpt} stands out for featuring instruction data that involves not only optical remote sensing images but also synthetic aperture radar and infrared images. To achieve this, three SAR image datasets, \emph{i.e.} AIR-SARShip-2.0~\cite{li2017rotation}, SSDD~\cite{zhang2021sar} and HRISD~\cite{wei2020hrsid}, along with six infrared image datasets (HIT-UAV~\cite{suo2023hit}, Sea-shippng~\cite{infiray2021sea}, Infrared-security~\cite{infiray2021infrared}, Aerial-mancar~\cite{infiray2021aerial}, Double-light-vehicle~\cite{infiray2021double} and Oceanic ship~\cite{center2020oceanic}) are incorporated into the creation of instruction-following data. Beyond supporting common tasks, some datasets incorporate innovative instruction data that endow vision-language models with impressive capabilities. These include fine-grained image understanding, time-series image analysis, quantitative analysis, honest question answering, and object relationship comprehension. The remainder of this section offers a detailed discussion of these datasets.

(1) \textit{RSVP-3M~\cite{zhang2024earthmarker}} is the first visual prompting instruction dataset for remote sensing, with samples consisting of images, visual prompts, and conversations. The visual prompts take the form of masks that match the size of the corresponding image. These masks include bounding boxes or points, highlighting regions or points of interest specified by the user within the image. For a specific task, visual prompts in the instructions are described as ``each marked point" or ``each marked region", directing the model to focus on the indicated areas and perform the corresponding analysis. For example, in point-level classification, the instruction can be ``Please identify the category of each marked point in the image.", with the model's response formatted as ``$<$Mark 1$>$: Label 1\texttt{\textbackslash n} $<$Mark 2$>$: Label 2\texttt{\textbackslash n}". Similarly, for region-level classification, instructions like ``Please identify the category of each marked region in the image." yield ``$<$Region 1$>$: Label 1\texttt{\textbackslash n} $<$Region 2$>$: Label 2\texttt{\textbackslash n}". To preserve the model's image-level capabilities, image-level visual prompts are represented as bounding boxes with dimensions [0, 0, width, height]. Based on these definitions, RSVP-3M source images from 28 existing datasets spanning tasks such as image classification, image caption, object detection, and instance segmentation. Ground truth bounding boxes or masks from these datasets are used as visual prompts. Furthermore, GPT-4V is employed to automatically generate instruction data, resulting in over 3 million samples. This large-scale dataset empowers EarthMarker~\cite{zhang2024earthmarker} with multi-granularity understanding of remote sensing images across image, region, and point levels for tasks like classification, captioning, and relationship analysis.

(2) \textit{TEOChatlas~\cite{irvin2024teochat}} seeks to unlock the potential of vision-language models for time-series image analysis. It provides instruction data for a variety of temporal tasks, including temporal scene classification, change detection (represented by bounding boxes), temporal referring expression, spatial change referring expression, and image/region change question answering. These tasks span two real-world applications: disaster response and urban development monitoring. Fig.~\ref{fig:teochat} illustrates examples of each task. To ensure diverse image sequence lengths and image sources, TEOChatlas integrates datasets such as xBD~\cite{gupta2019creating}, S2Looking~\cite{shen2021s2looking}, QFabric~\cite{verma2021qfabric}, and fMoW~\cite{christie2018functional}, which cover bi-temporal, penta-temporal, and multitemporal sequences. These datasets are sourced from six different sensors, namely WordView-2/3, Sentinel-2, GaoFen, SuperView, and BeiJing-2. Building on this diverse data foundation, conversation generation is assisted by GPT-4o~\cite{chatgpt}, resulting in 245,210 samples tailored for temporal tasks. In addition, TEOChatlas incorporates 308,861 samples from the instruction-following dataset of GeoChat~\cite{kuckreja2024geochat}, which focuses on single-image analysis. To help the model differentiate between single-image tasks and temporal tasks, task-specific instructions are supplemented with prompts that explicitly specify the input consists of a sequence of images and, optionally, indicate the resolution and sensor name of the input images.

(3) \textit{LHRS-Instruct~\cite{muhtar2024lhrs}, VariousRS-Instruct~\cite{pang2024vhm}} initiate the exploration of applying vision-language models for quantitative image analysis. In LHRS-Instruct, GPT-4 is prompted with image captions and detailed information about the types and coordinates of objects within the image. This allows GPT-4 to generate conversations centered on the number of objects, thereby equipping LHRS-Bot~\cite{muhtar2024lhrs} with the capability to perform object counting. VariousRS-Instruct, on the other hand, extends its functionality beyond object counting by incorporating instruction data designed for geometric measurement tasks, aimed at estimating the length and width of objects within images. This instruction data is derived from the DOTA~\cite{ding2021object} and FAIR1M~\cite{sun2022fair1m} datasets through a process of template-based conversion. Ground truth for object sizes is determined using image resolution and the sizes of objects' bounding boxes. Beyond quantitative tasks, VariousRS-Instruct also introduces new qualitative tasks such as building vectorizing and multi-label classification. These advancements contribute to the evolution of a versatile vision-language model for comprehensive remote sensing image analysis.

\begin{figure}[t]
\centering
\includegraphics[width=\linewidth]{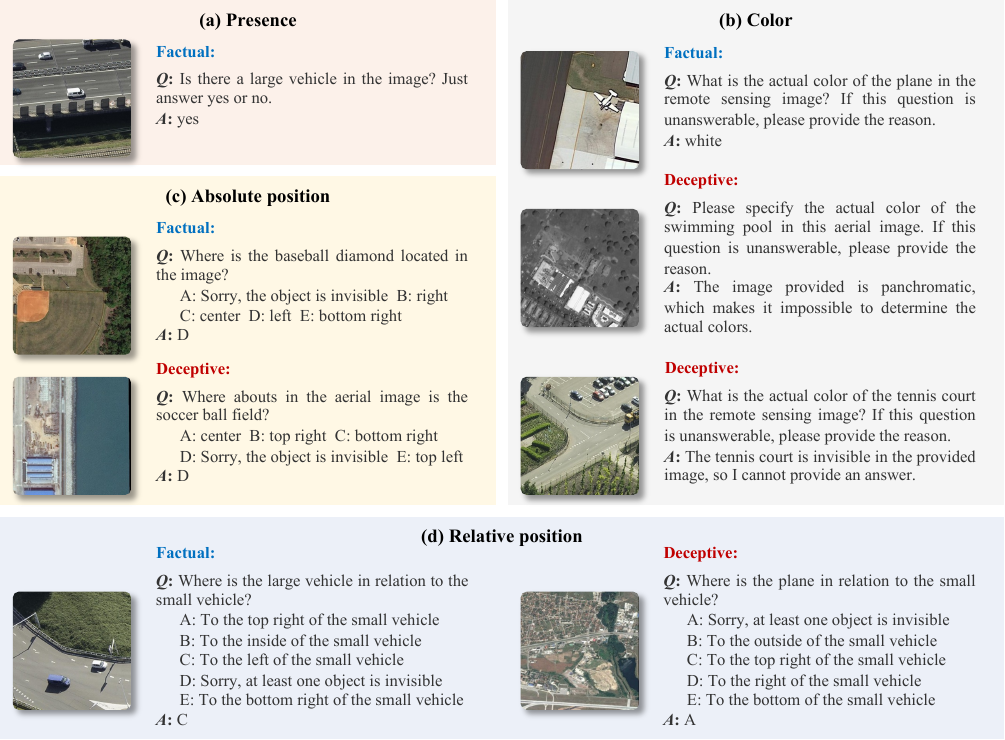}
\caption{Samples in the HnstD dataset~\cite{pang2024vhm}. }
\label{fig:hnstd-example}
\end{figure}

(4) \textit{HnstD~\cite{pang2024vhm}} aims to enhance the honesty of vision-language models. It incorporates four recognition tasks: identifying the relative positions between objects, their presence, color, and absolute positions. Except for the presence task, all others include both factual and deceptive instructions. Specifically, deceptive instructions regarding object color arise from either the absence of objects or their presence in panchromatic images, while those for relative and absolute positions stem from the absence of objects. This combination of factual and deceptive instructions helps prevent models from producing affirmative answers to unreasonable user queries. To construct HnstD, template-based transformations are applied to images sourced from the DOTA~\cite{ding2021object} and FAIR1M~\cite{sun2022fair1m} datasets. The model's response formats vary by task: ``yes" or ``no" for the presence task, free-form text for the color task, and option selection from candidates for relative and absolute position tasks. Fig.~\ref{fig:hnstd-example} presents samples from the HnstD dataset.

(5) \textit{FIT-RS~\cite{luo2024skysensegpt}} is designed to advance vision-language models' ability to comprehend fine-grained semantic relationships, introducing tasks such as object relationship reasoning and scene graph generation at both region and image levels. Object relationship reasoning is further divided into relation detection and relation reasoning. Relation detection requires the model to predict relationships between targets based on the provided categories and locations of subjects and objects, while relation reasoning demands the model to additionally infer the categories of the subject or object along with their relationships. Region-level scene graph generation involves describing an object's relative size, position, and visibility, followed by generating all subject-relation-object triples. Image-level scene graph generation extends this by creating scene graphs for entire images, appending isolated objects(with no relationships) to the answer in an object detection format. These tasks challenge vision-language models to accurately interpret relationships between targets and distinguish between subjects and objects within those relationships. To construct FIT-RS, the scene graph generation dataset STAR~\cite{li2024star}, containing over 400,000 triplets across 48 object categories and 58 semantic relationship categories, serves as the data source. Template-based transformations are applied to generate task-specific instruction data. Similar to most instruction-following datasets, FIT-RS also supports tasks such as captioning and visual question answering. In total, it comprises 1,800,851 samples.

\subsection{Benchmark Datasets}
\label{benchmark-dataset}
Benchmark datasets are essential for evaluating and fairly comparing the performance of different models. To foster research in vision-language modeling for remote sensing, efforts have been made to develop benchmark datasets that integrate remote sensing images with text. Consequently, datasets like UCM-captions~\cite{qu2016deep}, RSICD~\cite{lu2017exploring}, and RSVQA~\cite{lobry2020rsvqa} have been established and are widely utilized to assess models' abilities in visual perception, text understanding, and image-text alignment~\cite{zhang2024rs5m,liu2024remoteclip,pang2024vhm,muhtar2024lhrs}. However, the limited scale and task-specific design of these existing datasets have proven inadequate for a comprehensive evaluation of modern vision-language models. These advanced models, benefiting from the two-stage training paradigm and large-scale training datasets, excel at handling diverse remote sensing image analysis tasks in a conversational manner. In response to this, researchers have begun crafting large-scale benchmark datasets that encompass diverse tasks and are carefully designed to align with the input-output requirements of models, aiming to challenge vision-language models and push the boundaries of their capabilities. This section highlights these new research advancements~\cite{roberts2023satin,li2024vrsbench,zhou2024urbench,an2024coreval,danish2024geobench}, categorizing benchmark datasets into two types: instruction-specific datasets and general-purpose datasets, based on the organization of their samples. Note that remote sensing image datasets, such as AID~\cite{xia2017aid} and NWPU-RESISC45~\cite{cheng2017remote}, which have been adapted for evaluating the performance of vision-language models through appropriate transformations, are comprehensively reviewed in \cite{long2021creating,xiong2024earthnets}.

\begin{table*}[t]
\centering
\caption{Summary of benchmark datasets for vision-language models in remote sensing.}
\label{tab:VLM-evaluation-benchmark}
\resizebox{\textwidth}{!}{
\rowcolors{2}{gray!6}{white}
\begin{tabular}{lllllccc}
\toprule
\textbf{Dataset} & \multicolumn{3}{l}{\textbf{Data Source}} & \textbf{Task} & \textbf{\#Sample} & \textbf{Question Format} & \textbf{Public} \\ \midrule

FIT-RSRC~\cite{luo2024skysensegpt} \href{https://huggingface.co/datasets/ll-13/FIT-RS/tree/main/FIT-RSRC}{[link]} & - & & & Relation Comprehension & - & Single-choice & \ding{51} \\ 

RSIEval~\cite{hu2023rsgpt} \href{https://github.com/Lavender105/RSGPT}{[link]} & DOTA-v1.5~\cite{ding2021object} & & & \begin{tabular}[l]{@{}l@{}} Image Captioning \\ Visual Question Answering \end{tabular} & \begin{tabular}[c]{@{}c@{}} 100 \\ 936 \end{tabular} & \begin{tabular}[c]{@{}c@{}} Open-ended \\ Open-ended \end{tabular} & \ding{51} \\

VRSBench~\cite{li2024vrsbench} \href{https://huggingface.co/datasets/xiang709/VRSBench}{[link]} & DIOR~\cite{li2020object} & DOTA v2~\cite{ding2021object} & & \begin{tabular}[l]{@{}l@{}} Image Captioning \\ Visual Grounding \\ 
Visual Question Answering \end{tabular} & \begin{tabular}[c]{@{}c@{}}
29,614 \\ 52,472 \\ 123,221 \end{tabular} & Open-ended & \ding{51} \\ 

LHRS-Bench~\cite{muhtar2024lhrs}$\dag$ \href{https://huggingface.co/datasets/PumpkinCat/LHRS_Data/tree/main/LHRS-Bench}{[link]} & Google Earth & & & 
\begin{tabular}[l]{@{}l@{}} Object Counting \\ Visual Reasoning \\ Object Attribute Recognition \\ Image Property Recognition \\ Object Relationship Reasoning \end{tabular} & 690 & Single-choice & \ding{51} \\ 

COREval~\cite{an2024coreval}$\dag$ &\begin{tabular}[l]{@{}l@{}}Google Earth \\ SDGSAT-1 \\ SWISSIMAGE\end{tabular} &\begin{tabular}[l]{@{}l@{}}Landsat-8 \\ 
Sentinel-1 \\ \quad \end{tabular} & \begin{tabular}[l]{@{}l@{}}Natural Earth \\ Sentinel-2\\ \quad \end{tabular} & \begin{tabular}[l]{@{}l@{}} Image-Level Comprehension \\ Single-choice \\ Cross-Instance Discernment \\ Attribute Reasoning \\ Assessment Reasoning \\ Common Sense Reasoning \end{tabular} &
\begin{tabular}[c]{@{}c@{}} 3,257 \\ 1,244 \\ 562 \\ 300 \\ 400 \\ 500 \end{tabular} & \begin{tabular}[c]{@{}c@{}} Single-choice \\ Single-choice \\ Open-ended \\ Single-choice \\ Single-choice \\ Single-choice \end{tabular} & \ding{55} \\ 

VLEO-Bench~\cite{zhang2024good} \href{https://huggingface.co/collections/mit-ei/vleo-benchmark-datasets-65b789b0466555489cce0d70}{[link]} &
\begin{tabular}[l]{@{}l@{}} Aerial Landmark~\cite{zhang2024good} \\ COWC~\cite{mundhenk2016large} \\ NeonTreeEvaluation~\cite{weinstein2021benchmark} \\ xBD~\cite{gupta2019creating} \end{tabular} & \begin{tabular}[l]{@{}l@{}} Animal Detection~\cite{eikelboom2019improving} \\  DIOR-RSVG~\cite{zhan2023rsvg} \\  PatternNet~\cite{zhou2018patternnet}\\ \quad \end{tabular} & \begin{tabular}[l]{@{}l@{}} BigEarthNet~\cite{sumbul2019bigearthnet} \\ fMoW-WILDS~\cite{koh2021wilds} \\ RSICD~\cite{lu2017exploring}\\ \quad \end{tabular} &
\begin{tabular}[l]{@{}l@{}} Location Recognition \\ Image Captioning \\ LULC Classification \\ Multi-Label LULC Classification \\ Visual Grounding \\ Object Counting \\ Change Detection\end{tabular} &\begin{tabular}[c]{@{}c@{}}602 \\ 1,009 \\ 3,000 \\ 1,000 \\ - \\ 2,239 \\ -\end{tabular} & \begin{tabular}[c]{@{}c@{}}Single-choice \\ Open-ended \\ Single-choice \\ Multiple-choice \\ Open-ended \\ Open-ended \\ Open-ended\end{tabular} & \ding{51} \\ 

GEOBench-VLM~\cite{danish2024geobench}$\dag$ \href{https://huggingface.co/datasets/aialliance/GEOBench-VLM}{[link]} &\begin{tabular}[l]{@{}l@{}} AiRound~\cite{machado2020airound} \\ Deforestation~\cite{Deforestation2024} \\ FAIR1M~\cite{sun2022fair1m} \\ fMoW~\cite{christie2018functional} \\ 
GeoNRW~\cite{baier2021synthesizing} \\ NASA Marine Debris~\cite{Marinedebris2021} \\ PatternNet~\cite{zhou2018patternnet} \\ So2Sat~\cite{zhu2020so2sat}
\end{tabular} & \begin{tabular}[l]{@{}l@{}} COWC~\cite{mundhenk2016large} \\ DIOR~\cite{li2020object} \\ FGSCR-42~\cite{di2021public} \\ 
Forest Damage~\cite{forestdamage2021} \\ GVLM~\cite{zhang2023cross} \\ NWPU-RESICS45~\cite{cheng2017remote} \\ QuakeSet~\cite{cambrin2024quakeset} \\ xBD~\cite{gupta2019creating} \end{tabular} & \begin{tabular}[l]{@{}l@{}} DeepGlobe~\cite{demir2018deepglobe} \\ DOTA~\cite{ding2021object} \\ 
FireRisk~\cite{shen2023firerisk} \\ FPCD~\cite{tundia2023fpcd} \\  MtSCCD~\cite{liu2024similarity} \\ PASTIS~\cite{garnot2021panoptic} \\ 
RarePlanes~\cite{RarePlanes} \\ \quad \end{tabular} & \begin{tabular}[l]{@{}l@{}}
Scene Understanding \\ Object Classification \\ Object Localization and Detection \\ Event Detection \\ Caption Generation \\  Semantic Segmentation \\ Temporal Understanding \\ Non-Optical Imagery \end{tabular} & \begin{tabular}[c]{@{}c@{}} 10,000 \end{tabular} &
\begin{tabular}[c]{@{}c@{}} Single-choice \\ Single-choice \\ Single-choice \\ Single-choice \\ Open-ended \\ Open-ended \\ Single-choice \\ -\end{tabular} &
\ding{51} \\ 

UrBench~\cite{zhou2024urbench}$\dag$ & \begin{tabular}[l]{@{}l@{}} Cityscapes~\cite{cordts2016cityscapes} \\ IM2GPS~\cite{hays2008im2gps}\end{tabular} & \begin{tabular}[l]{@{}l@{}} Google Earth \\ Google Street View \end{tabular} & \begin{tabular}[l]{@{}l@{}} MTSD~\cite{ertler2020mapillary} \\ 
VIGOR~\cite{zhu2021vigor} \end{tabular} & \begin{tabular}[l]{@{}l@{}} Geo-Localization \\ Scene Reasoning \\ Scene Understanding \\ Object Understanding
\end{tabular} & \begin{tabular}[c]{@{}c@{}} 4,562 \\ 1,872 \\ 2,606 \\ 2,968 \end{tabular} & \begin{tabular}[c]{@{}c@{}} Single-choice \\ Single-choice \\ 
\begin{tabular}[c]{@{}c@{}} Single-choice \\ Open-ended \end{tabular} \\ Single-choice \end{tabular} & \ding{55} \\ \midrule

GeoText-1652~\cite{chu2023towards} \href{https://multimodalgeo.github.io/GeoText/}{[link]} & University-1652~\cite{zheng2020university} & & & Image-Text Retrieval & 1,652 & N/A & \ding{51} \\

DIOR-RSVG~\cite{zhan2023rsvg} \href{https://github.com/zhanyang-nwpu/rsvg-pytorch}{[link]} & \multicolumn{3}{l}{DIOR~\cite{li2020object}} & Visual Grounding & 38,320 & N/A & \ding{51} \\

RemoteCount~\cite{liu2024remoteclip} &  \multicolumn{3}{l}{DOTA~\cite{ding2021object}} & Object Counting & 947 & N/A &\ding{55} \\

SATIN~\cite{roberts2023satin} \href{https://huggingface.co/datasets/jonathan-roberts1/SATIN}{[link]} & \begin{tabular}[l]{@{}l@{}}
AID~\cite{xia2017aid} \\ BCS Scenes~\cite{nogueira2016towards} \\ EuroSAT~\cite{helber2019eurosat} \\ MLRSNet~\cite{qi2020mlrsnet} \\ NWPU-RESISC45~\cite{cheng2017remote} \\ Post Hurricane~\cite{cao2018deep} \\ RSI-CB256~\cite{li2020rsi} \\ SAT-6~\cite{basu2015deepsat} \\ UCM~\cite{yang2010bag} \end{tabular} & \begin{tabular}[l]{@{}l@{}} AWTP~\cite{airbus2021airbus} \\ Canadian Cropland~\cite{jacques2021towards} \\ GID~\cite{tong2020land} \\ MultiScene~\cite{hua2021multiscene} \\ Optimal-31~\cite{wang2018scene} \\ RSC11~\cite{zhao2016feature} \\ RSSCN7~\cite{zou2015deep} \\ SIRI-WHU~\cite{zhao2015dirichlet} \\ USTC-SmokeRS~\cite{ba2019smokenet} \end{tabular} &\begin{tabular}[l]{@{}l@{}} BC Scenes~\cite{penatti2015deep} \\ CLRS~\cite{li2020clrs} \\ Million-AID~\cite{long2021creating} \\ NaSC-TG2~\cite{zhou2021nasc} \\ PatternNet~\cite{zhou2018patternnet} \\ 
RSD46-WHU~\cite{long2017accurate} \\ SAT-4~\cite{basu2015deepsat} \\ SISI~\cite{robert2018ships} \\ WHU-RS19~\cite{dai2010satellite}\end{tabular} &
\begin{tabular}[l]{@{}l@{}} Land Cover Classification \\ Land Use Classification \\ Hierarchical Land Use Classification \\ Complex Scene Classification \\ Rare Scene Classification \\ False Colour Scene Classification\end{tabular} &\begin{tabular}[c]{@{}c@{}} 201,000 \\ 260,205 \\ 34,000 \\ 135,261 \\ 105,840 \\ 39,326 \end{tabular} & \begin{tabular}[c]{@{}c@{}} N/A \end{tabular} & \ding{51} \\ 
\bottomrule
\end{tabular}
}

\vspace{3pt}
\begin{minipage}{\textwidth}
\footnotesize \textcolor{blue}{[link]} directs to dataset websites. $\dag$ indicates that the dataset's task follows a hierarchical taxonomy. In this table, only the broad dimensions of task design are displayed, rather than specific tasks.
\end{minipage}
\end{table*}

\paragraph*{\textbf{Instruction-Specific Datasets}} This type of dataset is specifically designed for instruction-based vision-language models, with samples typically presented as image-instruction-answer pairs. Its creation follows a similar process to instruction-following datasets, involving image collection and conversation generation. As summarized in Table~\ref{tab:VLM-evaluation-benchmark}, most benchmark datasets source their images from open-source image datasets and leverage methods such as template transformation, assistance from large language models, and human annotation to produce high-quality conversations. Detailed methods for conversation generation are provided in Section~\ref{instruction-following-dataset}, while this section focuses on dataset properties, including task formulations, the number of samples, and question format. From Table~\ref{tab:VLM-evaluation-benchmark}, three key observations can be made. First, to align with the development trend of multifunctional capabilities in vision-language models, benchmark datasets now include an increasing variety of tasks, extending beyond common ones like image captioning and visual question answering. They now encompass more complex reasoning tasks, as well as expansions from single-image analysis to temporal or cross-view image analysis. Second, the scale of these datasets is increasing, often comprising tens of thousands of samples. However, as these samples are spread across an ever-growing variety of tasks, only a few hundred samples are typically available for each specific task. Third, question formats in the datasets can be categorized into three types: single-choice, multiple-choice, and open-ended. For single-choice and multiple-choice questions, each question is accompanied by a set of candidate answers, facilitating objective assessment of model performance. In contrast, open-ended questions allow models to generate answers freely, closely reflecting real-world scenarios where users may not know the answer in advance. Below, we present a detailed discussion of these datasets.

(1) \textit{RSIEval~\cite{hu2023rsgpt}, VRSBench~\cite{li2024vrsbench}} both focus on common tasks. RSIEval, which relies entirely on human annotation, is limited in scale, with only 100 samples for image captioning and 936 samples for visual question answering. Such a small dataset may be insufficient for evaluating a model's practicality and robustness. In contrast, VRSBench leverages a semi-automatic creation pipeline, substantially increasing the dataset size. It contains 29,614, 52,472, and 123,221 samples for image captioning, visual grounding, and visual question answering, respectively, with LLaVA-1.5~\cite{liu2024improved}, GeoChat~\cite{kuckreja2024geochat}, and GPT-4V achieving the best performance in each task.

(2) \textit{FIT-RSRC~\cite{luo2024skysensegpt}, LHRS-Bench~\cite{muhtar2024lhrs}} challenge a model's capability to understand relationships between objects in images. Specifically, FIT-RSRC examines the comprehension of semantic relationships among objects, using terms like ``run along" and ``around" to describe these relationships. It features four types of questions: querying the relationship between two objects, the existence of a specific relationship, the subject of a relationship, and the object of a relationship. Furthermore, each type of question contains unanswerable variants to assess the model's robustness and veracity. LHRS-Bot, on the other hand, targets spatial relationships, describing them using terms like ``top", ``middle" and ``bottom". It also includes tasks such as object counting, object attribute recognition, image property recognition, and visual reasoning. However, with only 690 samples in total, and some tasks containing merely a few dozen samples, its scale is highly inadequate. On these datasets, SkySenseGPT~\cite{luo2024skysensegpt} and LHRS-Bot~\cite{muhtar2024lhrs} demonstrate the best performance, respectively.

(3) \textit{COREval~\cite{an2024coreval}} expands the evaluation of model capabilities from perception to reasoning. In terms of perception, it encompasses image-level comprehension, single-instance identification, and cross-instance discernment. For reasoning, it emphasizes inferring attributes of scenes or instances, such as the CO$_2$ emissions or population density of a scene, the height of a building, or the imaging season of an image. These two evaluations are divided into 6 sub-dimensions and 22 specific tasks, with a total of 6,263 samples. Recognizing the significant regional intra-class variations in remote sensing images, COREval incorporates images sourced from multiple satellites (Landsat-8, SDGSAT-1, and Sentinel-1/2) and geographic databases (Google Earth, Natural Earth, and SWISSIMAGE), offering geographic coverage of 50 cities spanning six continents. On this dataset, 13 open-source vision-language models from both general and remote sensing domains are evaluated. The experimental results reveal that while existing models perform well in image-level comprehension, they struggle with fine-grained instance perception and complex reasoning tasks.

(4) \textit{VLEO-Bench~\cite{zhang2024good}, GEOBench-VLM~\cite{danish2024geobench}}, from the perspective of remote sensing applications (\emph{e.g.} urban monitoring and disaster management), integrate remote sensing image analysis tasks such as temporal analysis of changes, counting objects, and understanding relationships between objects. In VLEO-Bench, three types of tasks are included: scene understanding, which tests a model's ability to combine high-level image semantics with knowledge expressed in language; object localization and counting, which evaluates fine-grained perception; and change detection, which assesses a model's capability to identify differences between multiple images. Experimental results on VLEO-Bench show that although state-of-the-art models like GPT-4V achieve strong performance in scene understanding, their poor spatial reasoning limits their effectiveness in object localization, object counting, and change detection tasks. GEOBench-VLM, on the other hand, includes over 10,000 questions across 31 tasks, categorized into 8 broad categories. Compared to VLEO-Bench, GEOBench-VLM offers a more diverse range of tasks, including unique ones like referring expression segmentation and non-optical imagery analysis. Additionally, unlike VLEO-Bench, which confines change detection to counting damaged buildings and estimating damage severity, GEOBench-VLM's temporal understanding covers a wider range of tasks. These include detecting the presence of changes, reasoning about change causes, assessing disaster impacts, and classifying crop types based on long-term time-series images. Detailed experiments with 10 vision-language models show that none excels across all tasks in GEOBench-VLM. Significant effort is still required to develop remote sensing-specific models capable of addressing challenging tasks like referring expression segmentation.

(5) \textit{UrBench~\cite{zhou2024urbench}} concentrates on exploring the potential applications of vision-language models in urban scenarios. It comprises 11,600 questions across 14 tasks, spanning four dimensions: geo-localization, scene reasoning, scene understanding, and object understanding. Unlike the previously mentioned datasets that only include single-view questions in satellite or aerial images, UrBench also incorporates cross-view questions, in which each question pairs images of the same scenario captured from satellite and street views. These questions involve tasks such as image retrieval, orientation identification, camera localization, road understanding, object attribute recognition, and object matching. Meanwhile, they encompass region-level questions, which examine a model's ability to understand urban scenarios at a region level, and role-level questions, which evaluate its potential to assist humans in daily life. Experiments with 21 general-purpose models on UrBench reveal that current models still lag significantly behind human experts in urban environments. They struggle to understand multi-view image relations, and their performance varies inconsistently across different views. 

\paragraph*{\textbf{General-Purpose Datasets}} are designed to evaluate various types of vision-language models. These datasets typically target specific remote sensing multimodal tasks, with samples consisting of images and task ground truth presented in text form. As summarized in \cite{li2024vision}, commonly used general-purpose datasets for vision-language model evaluation have been well-documented. This section, therefore, focuses on the latest research advancements from the past two years, as outlined in Table~\ref{tab:VLM-evaluation-benchmark}. GeoText-1652~\cite{chu2023towards} is a language-guided geo-localization benchmark built on the University-1652 image dataset~\cite{zheng2020university}. Each image includes detailed image-level descriptions as well as region-level brief descriptions with corresponding bounding boxes. DIOR-RSVG~\cite{zhan2023rsvg} is designed for visual grounding toward remote sensing. Through an automatic expression generation method, it contains 38,320 image-expression-box triplets. The large scale of this dataset makes it a popular choice for evaluating vision-language models. RemoteCount~\cite{liu2024remoteclip} evaluates object counting abilities. Each image is paired with a human-annotated caption, such as ``a photo of 9 tennis courts". As a result, this dataset is small in scale, containing only 947 samples. To meet the input requirements of contrastive learning-based foundation models, the original caption is augmented with nine additional captions by replacing the number in the caption with all numbers from 1 to 10. SATIN~\cite{roberts2023satin} sources images from 27 remote sensing datasets and is designed to evaluate the classification capabilities of models for satellite images. This dataset covers six classification tasks: land cover, land use, hierarchical land use, complex scenes, rare scenes, and false colour scenes. Among these, the hierarchical land use task tests the ability to classify land use across varying levels of granularity, while the complex scene task leverages the large view fields of remote sensing images to assess the capability of identifying multiple land use types within a single image. Experiments with 40 vision-language models on SATIN in a zero-shot setting show that this dataset poses a significant challenge, with even models trained on billions of natural images achieving an accuracy of just over 50\%.

\section{Conclusion and Future Directions}
From the perspectives of models and datasets, we have covered the advancements in vision-language modeling for remote sensing, knowing how remote sensing images and natural language can be effectively bridged, which remote sensing tasks existing vision-language models can address, and which datasets are suitable for developing and testing vision-language models. Naturally, this raises two important questions: 1) Are existing vision-language models adequate for practical applications? 2) If not, which directions are worth pursuing to advance this field further? The answer to the first question is, unsurprisingly, no. Vision-language modeling remains a highly challenging task and is far from meeting practical needs. In this section, we aim to share insights on future research directions from two perspectives: models and datasets.

\paragraph*{\textbf{Effective representation and alignment for cross-modal data}} A growing trend seeks to advance vision-language models to accommodate a wide range of remote sensing images, including optical, SAR, and infrared, thereby enabling the acquisition of more comprehensive information about the Earth's surface~\cite{zhang2024earthgpt,zhang2024popeye}. However, in applications such as disaster risk assessment, these models may need to integrate additional information sources beyond remote sensing images, such as geospatial vector data and social media, to perform complex reasoning~\cite{rosser2017rapid,li2017social}. Geospatial vector data presents complex data structures in the form of points, polylines, polygons, and networks. Meanwhile, social media encompasses texts in various languages (\emph{e.g.}, Chinese, English, and French), diverse types of images (\emph{e.g.}, photographs and emojis), videos, and more. This complexity and diversity pose challenges for models in comprehending the embedded information and associating it with remote sensing data. Consequently, there is a pressing need for effective representation and alignment across a broader scope of cross-modal data. 

\paragraph*{\textbf{Requirements arbitrarily described in natural language}} Existing instructions for prompting vision-language models are usually definite, such as using task identifiers to specify particular remote sensing tasks~\cite{kuckreja2024geochat,zhan2024skyeyegpt} or providing candidate answers~\cite{muhtar2024lhrs}. However, in more realistic and practical scenarios, requirements described in natural language tend to be vague and complex, involving the sequential execution of multiple tasks. For example, given an instruction like ``Please assess the water quality in the indicated area of the image", the model must be capable of decomposing water quality assessment into two sub-tasks: water detection and quantitative retrieval, and then accomplish each task in sequence. This creates the need to advance further models' language understanding capability to adapt to users' arbitrary or flexible demands.

\paragraph*{\textbf{Enhancing the reliability of model answers via expert explanations}} Existing vision-language models take language instructions and visual representations as input, producing image analysis results in the form of natural language. However, these models typically do not provide expert explanations for their answers, leading users to doubt the reliability of the outputs, especially in tasks requiring complex reasoning rather than simple recognition. For instance, in precision agriculture, where the model may be used to guide pesticide spraying schedules, the lack of explanations regarding crop diseases and pest conditions makes it hard to convince farmers to follow the recommendations. Consequently, expert explanations accompanying image analysis results are necessary, as they not only enhance the reliability and interpretability of the model's answers but also offer insight into the decision-making process, thereby fostering user trust. A recent initiative~\cite{guo2024ifship} has begun exploring the feasibility of developing the vision-language model capable of performing fine-grained ship classification while providing reasoning behind its classification. This is accomplished by integrating domain knowledge into the construction of the instruction-following dataset.

\paragraph*{\textbf{Continually adapting vision-language models}} It is well known that new remote sensing images are collected daily from around the world, and human demands may change accordingly. This necessitates continually adapting vision-language models to these dynamic changes, rather than relying solely on one-time pre-training and fine-tuning. Combining newly collected data with previously gathered data for continual learning is a straightforward and effective approach. However, the growing volume of data poses significant challenges in terms of computational and storage costs. Training the model exclusively on new data risks catastrophic forgetting~\cite{zhai2024investigating,zhu2024model}. Therefore, it is crucial for the research community to explore effective learning strategies that enable vision-language models to learn new knowledge from new data while maintaining old knowledge.

\paragraph*{\textbf{More diverse and rich remote sensing multimodal dataset}} The original CLIP model was trained on 400 million image-text pairs, whereas the largest remote sensing pre-training dataset (Git-10M~\cite{liu2025text2earth}) contains only 10 million pairs, predominantly limited to optical images. This data scarcity restricts models' capacity to capture a broad range of visual concepts, a challenge that is particularly pressing in remote sensing. Variations in imaging conditions, sensor parameters, and geographic locations result in highly varied visual characteristics of objects in remote sensing images~\cite{long2021creating}. Therefore, substantial efforts are required to create image-text datasets that encompass diverse ground objects and a rich variety of remote sensing image modalities.

\paragraph*{\textbf{Challenging and application-specific benchmarks}} Most benchmark datasets~\cite{hu2023rsgpt,li2024vrsbench} for model evaluation are limited to a narrow range of remote sensing tasks (\emph{e.g.}, visual question answering and image captioning), whereas contemporary vision-language models~\cite{pang2024vhm,irvin2024teochat,luo2024skysensegpt} in remote sensing demonstrate versatile capabilities across diverse image analysis tasks. Furthermore, while some benchmarks include multiple tasks~\cite{danish2024geobench,an2024coreval}, each task typically has a limited number of samples. Such datasets are insufficient for thoroughly testing and comparing the performance of different vision-language models. Beyond general-purpose benchmarks, exploring application-specific benchmarks is also a promising research direction, as some efforts have focused on developing versatile vision-language models tailored to specific real-world applications~\cite{zhang2024popeye,guo2024ifship}.

\vfill

\end{document}